\documentclass[10pt,journal,compsoc]{IEEEtran}

\usepackage{amsfonts}
\usepackage{balance}
\usepackage{booktabs}
\usepackage{amsmath}
\usepackage{tabularx}
\usepackage{subfigure}
\usepackage{graphicx}
\usepackage{multirow}
\usepackage{rotating}
\usepackage{algorithmic}
\usepackage{footnote}
\usepackage{threeparttable}
\usepackage{color}
\usepackage{textcomp}

\usepackage{hyperref} 
\usepackage{url}

\ifCLASSOPTIONcompsoc
  \usepackage[nocompress]{cite}
\else
  \usepackage{cite}
\fi

\ifCLASSINFOpdf
\else
\fi

\begin{document}

\title{Fast Multi-view Clustering via Ensembles:\\ Towards Scalability, Superiority, and Simplicity}

\author{Dong Huang,~\IEEEmembership{Member,~IEEE, }
        Chang-Dong Wang,~\IEEEmembership{Member,~IEEE, }\\
        and~Jian-Huang Lai,~\IEEEmembership{Senior Member,~IEEE, }
\IEEEcompsocitemizethanks{\IEEEcompsocthanksitem D. Huang is with the College of Mathematics and Informatics, South China Agricultural University, Guangzhou, China, and also with Key Laboratory of Smart Agricultural Technology in Tropical South China, Ministry of Agriculture and Rural Affairs, China. \protect\\
E-mail: huangdonghere@gmail.com.
\IEEEcompsocthanksitem C.-D. Wang and J.-H. Lai are with the School of Computer Science and Engineering,
Sun Yat-sen University, Guangzhou, China, and also with Guangdong Key Laboratory of Information Security Technology, Guangzhou, China, and also with Key Laboratory of Machine Intelligence and Advanced Computing, Ministry of Education, China.\protect\\
E-mail: changdongwang@hotmail.com, stsljh@mail.sysu.edu.cn.}
}

\markboth{}%
{Shell \MakeLowercase{\textit{et al.}}: Bare Demo of IEEEtran.cls for Computer Society Journals}

\IEEEtitleabstractindextext{
\begin{abstract}
Despite significant progress, there remain three limitations to the previous multi-view clustering algorithms. First, they often suffer from high computational complexity, restricting their feasibility for large-scale datasets. Second, they typically fuse multi-view information via one-stage fusion, neglecting the possibilities in multi-stage fusions. Third,  dataset-specific hyperparameter-tuning is frequently required, further undermining their practicability. In light of this, we propose a \textbf{fast} \textbf{m}ulti-v\textbf{i}ew \textbf{c}lustering via \textbf{e}nsembles (FastMICE) approach. Particularly, the concept of random view groups is presented to capture the versatile view-wise relationships, through which the hybrid early-late fusion strategy is designed to enable efficient multi-stage fusions. With \emph{multiple} views extended to \emph{many} view groups, three levels of diversity (w.r.t. features, anchors, and neighbors, respectively) are jointly leveraged for constructing the view-sharing bipartite graphs in the early-stage fusion. Then, a set of diversified base clusterings for different view groups are obtained via fast graph partitioning, which are further formulated into a unified bipartite graph for final clustering in the late-stage fusion. Notably, FastMICE has almost linear time and space complexity, and is free of dataset-specific tuning. Experiments on 22 multi-view datasets demonstrate its advantages in scalability (for extremely large datasets), superiority (in clustering performance), and simplicity (to be applied) over the state-of-the-art. Code available: \url{https://github.com/huangdonghere/FastMICE}.
\end{abstract}

\begin{IEEEkeywords}
Data clustering, Multi-view clustering, Ensemble clustering, Large-scale clustering, Hybrid early-late fusion, Linear time.
\end{IEEEkeywords}}

\maketitle

\IEEEdisplaynontitleabstractindextext

\IEEEpeerreviewmaketitle

\IEEEraisesectionheading{\section{Introduction}\label{sec:introduction}}

\IEEEPARstart{C}{lustering} analysis has been a fundamental yet challenging research topic in knowledge discovery and data mining \cite{jain10_survey}. It aims to partition a set of data samples into a certain number of homogeneous groups, each of which is referred to as a cluster. Among the various sub-topics in clustering analysis,  multi-view clustering (MVC) has recently gained a considerable amount of attention due to its advantage in fusing common and complementary information from multiple views (or data sources) to enhance the clustering performance \cite{chao21_tai}. Despite the  proposals of many MVC algorithms, there are still three crucial questions (with regard to scalability, information fusion, and hyperparameter tuning, respectively) that remain to be addressed.

\emph{First, how to enable MVC for very large-scale datasets?} Though a large quantity of MVC algorithms have been developed in recent years, the high computational complexity is still a major hurdle for many of them to be applied in large-scale scenarios. In previous MVC works, there are several often-encountered complexity bottlenecks, such as affinity graph construction, graph partitioning, and some other expensive matrix computations. The affinity graph construction is a basic step in many MVC algorithms, which formulates the sample-wise relationship by computing an $N\times N$ affinity matrix and generally takes $O(N^2d)$ time and $O(N^2)$ space, where $N$ is the number of samples and $d$ is the dimension. The graph partitioning (typically by spectral clustering)  is another computationally expensive step in many MVC algorithms \cite{xia14_aaai,nei16_amgl,Liang2022,kang20_nn,tao20_tnnls,liang20_tkde}, which often requires singular value decomposition (SVD) and takes $O(N^3)$ time and $O(N^2)$ space. Especially, the graph partitioning via spectral clustering is adopted as an important step in many MVC algorithms, such as multi-view spectral clustering \cite{xia14_aaai,liang20_tkde}, multi-view subspace clustering \cite{kang20_nn,zhang15_iccv}, and multi-view graph learning \cite{nei16_amgl,Liang2022}, which, together with some other expensive matrix computations, contributes to the $O(N^3)$ complexity bottleneck in these MVC algorithms \cite{xia14_aaai,nei16_amgl,Liang2022,kang20_nn,tao20_tnnls,liang20_tkde}.

\begin{figure}[!t]
	\begin{center}
		{\subfigure[] 
			{\includegraphics[width=0.428\linewidth]{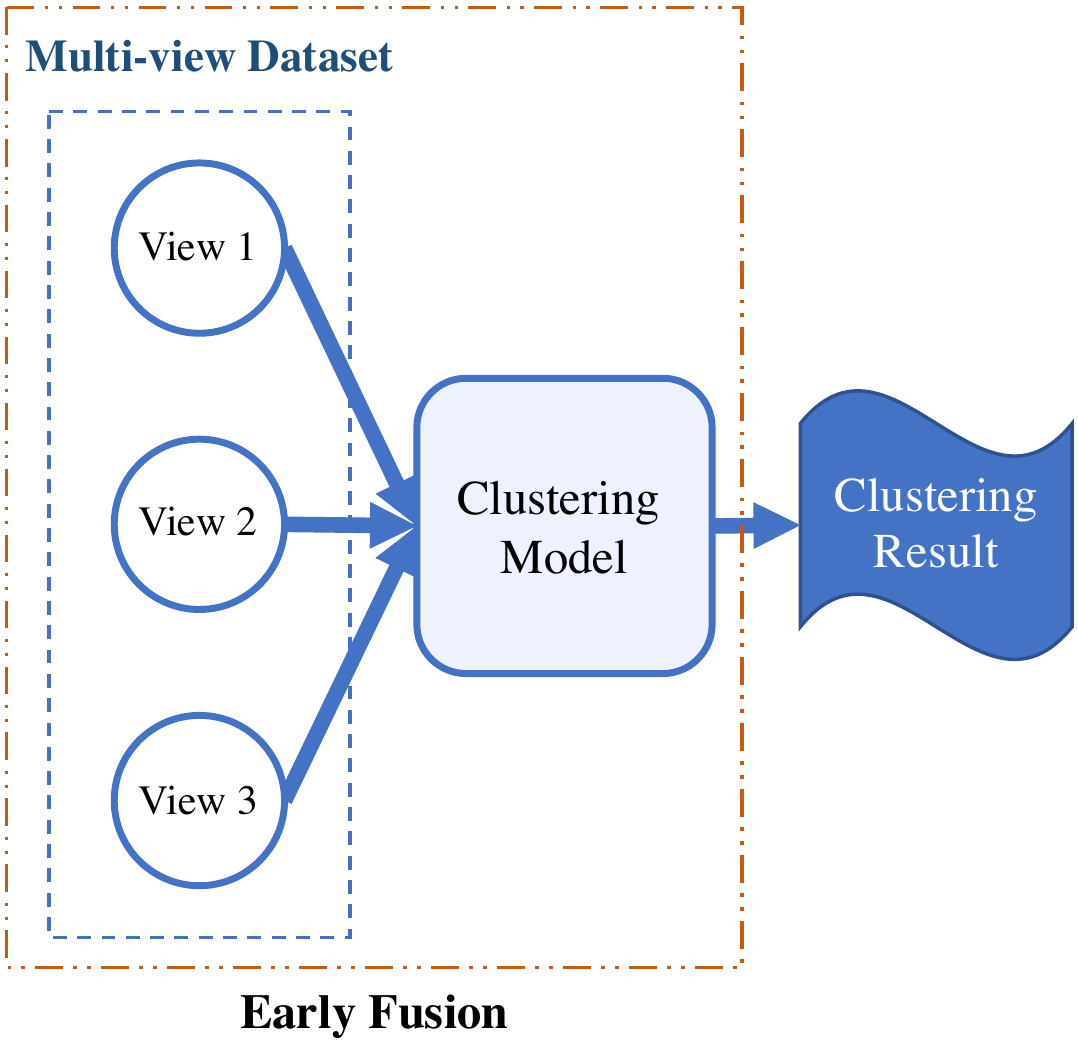}\label{fig:compFusionEarly}}}\hspace{0.001mm}
		{\subfigure[] 
			{\includegraphics[width=0.5571\linewidth]{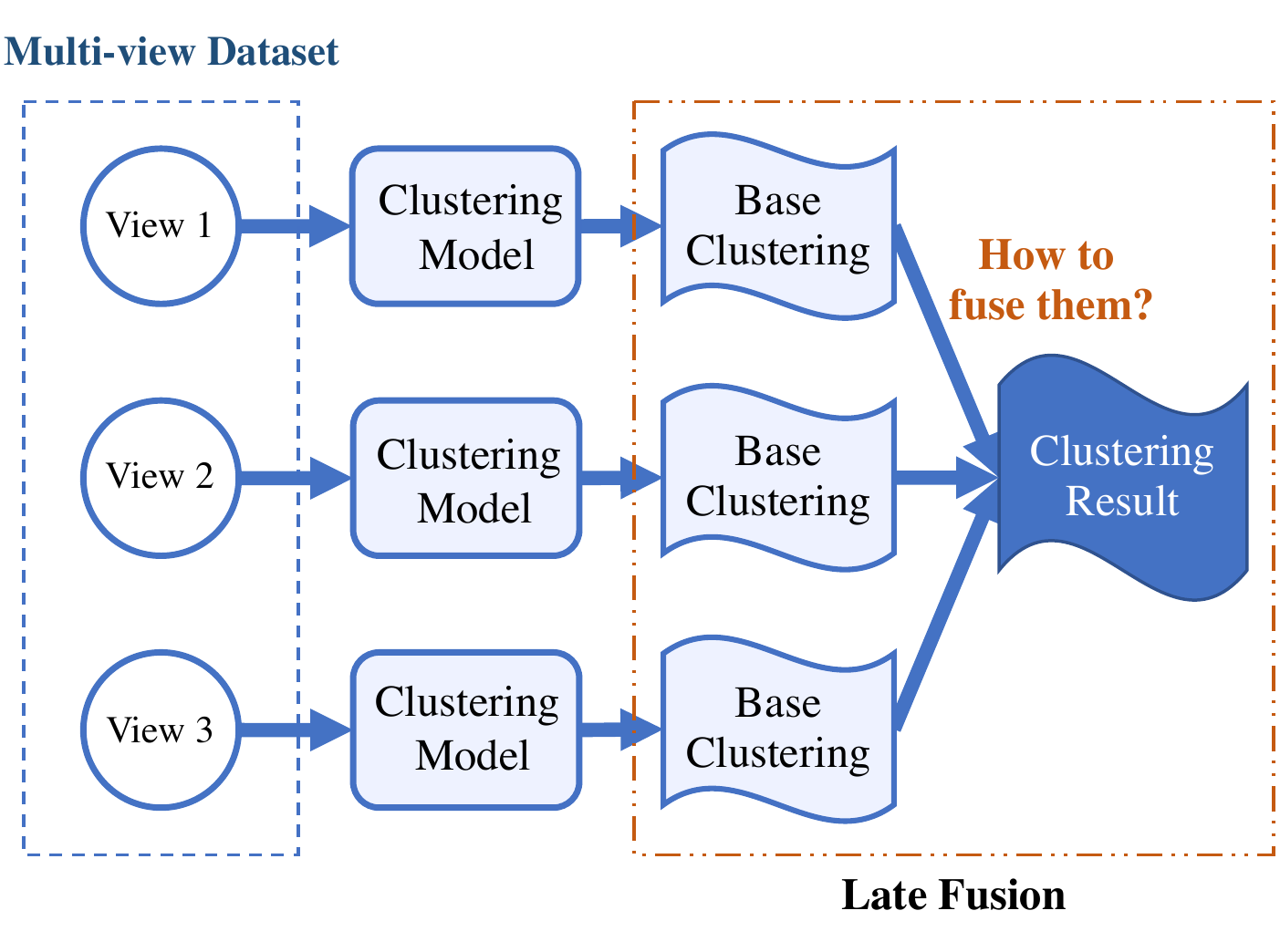}\label{fig:compFusionLate}}}
		\caption{In which stage can the multi-view information be fused? Most existing MVC works adopt either (a) early fusion or (b) late fusion (as illustrated on an example of three views).}
		\label{fig:compFusion1}
	\end{center}
\end{figure}

\emph{Second, in which stage should multi-view information be fused?} An essential task of MVC is to fuse the information from multiple views for robust clustering. The difference in various MVC algorithms is typically reflected by \emph{in which stage} and \emph{how} they conduct the multi-view fusion. A naive strategy is to directly concatenate the features from multiple views and then perform some single-view clustering algorithm on the concatenated features, which in practice is rarely adopted as it neglects the rich and complementary information across multiple views. In the MVC literature, the most widely-adopted strategy may be the \emph{early fusion} \cite{xia14_aaai, zhang15_iccv, liang19_icdm,liang20_tkde}, which typically fuse the multi-view information in a unified clustering model via optimization or some heuristics (as illustrated in Figs.~\ref{fig:compFusionEarly}).
Besides the early fusion,
another popular strategy is the \emph{late fusion} \cite{liu19_pami,wang19_ijcai,kang20_nn,liu21_icml}, which first obtains multiple base clusterings by performing the clustering process on each view separately and then fuses these base clusterings into a more robust clustering result in the final stage (as illustrated in Figs.~\ref{fig:compFusionLate}).  While most of the existing MVC algorithms adopt either early fusion or late fusion, it is surprising that few of them have gone beyond the single-stage fusion to explore the rich possibilities and potential benefits hidden in the multi-stage fusion formulation.

\emph{Third, is the dataset-specific fine-tuning necessary?} The regularization hyperparameters or some other types of hyperparameters are often involved in previous MVC algorithms to adjust the influences of different terms (or components) \cite{chao21_tai}, where dataset-specific fine-tuning is frequently required to seek the proper values of these hyperparameters in a probably extensive trial-and-error manner. However, unlike supervised or semi-supervised learning \cite{li19TPAMI,wang22TCSVT}, in the unsupervised situations it may be arguable whether the ground-truth labels can be used for guiding the fine-tuning process. Without the fine-tuning guided by partial or even all ground-truth labels, the practicality of these MVC algorithms may be significantly weakened.  Moreover, when the number of the tuning-intensive hyperparameters goes to three or even more,  their tuning costs (typically via grid-search) might become very expensive on large datasets (as reported in Table~\ref{table:compare_time}), which give rise to the critical question whether the dataset-specific tuning can be eliminated while maintaining robust clustering performance.

More recently several efforts have been made to deal with some of the above issues. In single-view clustering, it has proved to be an effective strategy to represent large-scale data samples via a set of anchors (also known as landmarks or representatives) \cite{cai15_LSC,huang19_tkde}, which can substantially facilitate the computation of the graph construction and partitioning for large-scale datasets. When it goes from single-view to multi-view, the anchor-based formulation still shows its promising ability \cite{li15_mvsc,kang2020_lmvsc,sun21_smvsc,wang22_tip}, but also faces a series of new challenges, ranging from multi-view anchor selection to multi-view information fusion. Typically, Li et al. \cite{li15_mvsc} selected a set of anchors by performing $k$-means on the concatenated multi-view features. With this unified anchor set, a bipartite graph is built between data samples and anchors on each view, and then multiple bipartite graphs are fused into a consensus graph for the final clustering \cite{li15_mvsc}. Wang et al. \cite{wang22_tip} exploited the self-expressive loss to learn a set of unified anchors and a bipartite graph for all views. However, considering the diverse characteristics of multiple views, a single set of unified anchors may not sufficiently capture the rich and complementary information of all views. Different from the pursuit of a single anchor set for \emph{all} views \cite{li15_mvsc,wang22_tip}, Kang et al. \cite{kang2020_lmvsc} learned a set of anchors on \emph{each} view, and then built a bipartite graph on each view separately. Then the multiple bipartite graphs are combined into a unified graph for final clustering. However, since each anchor set on a view is learned independently of other views, the cross-view information is inherently neglected in the construction of each single-view graph, which may lead to a degraded capacity of multi-view expressiveness. Despite the progress, these methods \cite{li15_mvsc,sun21_smvsc,wang22_tip,kang2020_lmvsc} either jointly construct a \emph{single} anchor set \emph{for all} views \cite{li15_mvsc,sun21_smvsc,wang22_tip} or separately construct a \emph{single} anchor set \emph{for each} view \cite{kang2020_lmvsc},  which, however, neglect the possibilities hidden between \emph{single} and \emph{all}, and dwell in the single-stage fusion strategy without the ability to capture the view-wise relationships in multiple stages.  Furthermore, the requirement of dataset-specific hyperparameter-tuning in many of them \cite{li15_mvsc,sun21_smvsc,kang2020_lmvsc,zhang19_bmvc} also poses a practical hurdle for their real-world applications.

\begin{figure}[!t]
\begin{center}
{
{\includegraphics[width=0.96\linewidth]{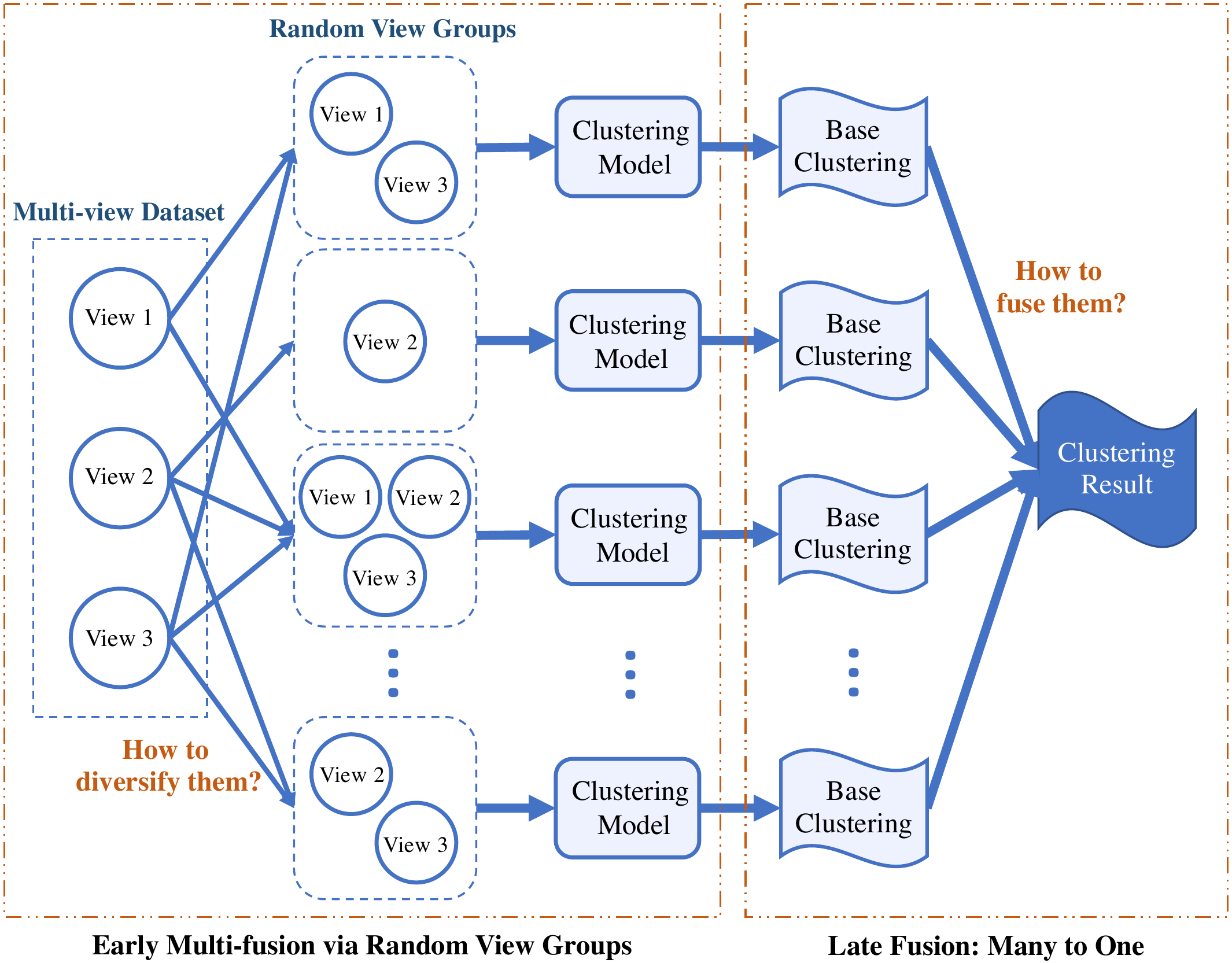}\label{fig:compFusionEarlyLate}}}
\caption{Illustration of the hybrid early-late fusion strategy on an example of three views. With the benefits brought in by the random view groups, the key questions arise as to how to diversify them, how to fuse them, and especially how to ensure the clustering robustness while maintaining the scalability for extremely large datasets.}
\label{fig:compFusion2}
\end{center}
\end{figure}

To jointly address the above-mentioned issues, in this paper, we propose a \textbf{fast} \textbf{m}ulti-v\textbf{i}ew \textbf{c}lustering via \textbf{e}nsembles (FastMICE) approach. Different from previous MVC approaches that tend to work at either single views or all views in each stage, this paper first presents the concept of random view groups, which serves as the basic form of our flexible view-organizations. Specifically, with each view group consisting of a random number of view members, the \emph{multiple} views can be extended to \emph{many} random view groups for investigating the view-wise relationships in a diversified manner. Based on random view groups, a hybrid early-late fusion strategy is devised to enable efficient and robust fusions at multiple stages (as shown in Fig.~\ref{fig:compFusion2}). In the early stage, multiple fusions are simultaneously  performed in multiple view groups, where three levels of diversity, namely, feature-level diversity, anchor-level diversity, and neighborhood-level diversity, are jointly leveraged to construct a set of view-sharing bipartite graphs. By efficient bipartite graph partitioning, a set of diversified base clusterings are obtained, which are further utilized to construct a unified bipartite graph for achieving the final clustering in the late-stage fusion. It is noteworthy that our FastMICE approach has almost linear time and space complexity, and is capable of producing high-quality clustering results without dataset-specific tuning. Extensive experiments are conducted on 22 real-world multi-view datasets, including 10 general-scale datasets and 12 large-scale datasets, which demonstrate the scalability (for extremely large datasets), the superiority (in clustering performance), and the simplicity (to be applied) of our approach.
For clarity, the contributions of this work are summarized below.

\begin{itemize}
\item This paper for the first time, to our knowledge, presents the concept of \textit{random view group} to capture the versatile view-wise relationships, which extends \emph{multiple} views to \emph{many} random view groups and may significantly benefit the clustering robustness while maintaining high efficiency.
\item A hybrid early-late fusion strategy is devised, which breaks through the conventional single-stage fusion paradigm and enables the multi-stage fusions to jointly explore different levels of the multi-view information.
\item A novel large-scale MVC approach termed FastMICE is proposed, whose advantages are three-fold: (i) it has almost \textit{linear time and space complexity} and is feasible for very large-scale datasets; (ii) it is able to achieve \textit{superior clustering performance} over the state-of-the-art approaches as confirmed by extensive experimental results; (iii) it is simple to be applied, where \textit{no dataset-specific hyperparameter-tuning} is required across various multi-view datasets.
\end{itemize}

The remainder of the paper is organized as follows. Section~\ref{sec:related_work} reviews the related works on multi-view clustering and ensemble clustering. Section~\ref{sec:framework} describes the overall framework of our FastMICE approach. Section~\ref{sec:experiments} reports the experimental results. Finally, Section~\ref{sec:conclusion} concludes this paper.

\section{Related Work} 
\label{sec:related_work}

In this paper, we propose a new large-scale MVC approach termed FastMICE, which involves both MVC and ensemble clustering (EC). In this section, the related works on MVC and EC will be reviewed in Sections~\ref{sec:related_work_mvc} and \ref{sec:related_work_ec}, respectively. 

\subsection{Multi-view Clustering}
\label{sec:related_work_mvc}

In recent years, many MVC methods have been developed from different technical perspectives \cite{chao21_tai}. In spite of the difference in their specific models, they typically share a common and essential task, that is, how to fuse the information from multiple views. A straightforward strategy is to concatenate the features from all views and then perform single-view clustering on the concatenated features, which, however, ignores the multi-view complementariness and is rarely adopted. Besides feature concatenation, according to their fusion stage, most of the existing MVC methods can be classified into two categories, i.e., the early fusion methods \cite{xia14_aaai, zhang15_iccv, liang19_icdm} and the late fusion methods \cite{liu19_pami,wang19_ijcai,kang20_nn}.

Early fusion is probably the most widely-adopted fusion strategy in MVC \cite{xia14_aaai, zhang15_iccv, liang19_icdm,liang20_tkde}, which typically formulates the information of all views in a unified optimization or heuristic model (as shown in Fig.~\ref{fig:compFusionEarly}). In early fusion, the information of each view can be given via different representations, such as the original features \cite{zhang15_iccv}, transition probability matrix \cite{xia14_aaai}, $K$-nearest-neighbor ($K$-NN) graph \cite{liang19_icdm}, and so forth. Xia et al. \cite{xia14_aaai} constructed a shared low-rank transition probability matrix by exploiting multiple transition probability matrices from multiple views, and then performed spectral clustering on the shared matrix for final clustering. Zhang et al. \cite{zhang15_iccv} conducted multi-view subspace clustering by minimizing a self-expressive loss on each view with the tensorized low-rank constraint. Xie et al. \cite{xie20_tcyb} extended the tensorized multi-view subspace clustering by further incorporating a local structure constraint. Liang et al.  \cite{liang19_icdm,Liang2022} performed graph fusion on multiple $K$-NN graphs from multiple views with cross-view consistency and inconsistency jointly modeled.

Late fusion is another popular fusion strategy in recent years \cite{liu19_pami,wang19_ijcai,kang20_nn,liu21_icml}, which first builds a base clustering on each view (often separately) and then fuses them at the partition-level to obtain a consensus clustering (as shown in Fig.~\ref{fig:compFusionLate}). Wang et al. \cite{wang19_ijcai} built the base clusterings by performing kernel k-means clustering on each view, and learned a consensus clustering by maximizing alignment between the consensus clustering and the base clusterings. Kang et al. \cite{kang20_nn} performed spectral clustering on the subspace representation of each view to obtain a corresponding base clustering, and learned a consensus clustering by minimizing the distance between a unified cluster indicator matrix and the multiple base clusterings \cite{kang20_nn}.

These MVC methods \cite{xia14_aaai, zhang15_iccv, liang19_icdm,liang20_tkde,liu19_pami,wang19_ijcai,kang20_nn,liu21_icml} seek to fuse the multi-view information in different stages and through different techniques. Yet surprisingly, most of them perform the fusion in a single-stage manner (either in the early stage or in the late stage), which lack the ability to go beyond the single-stage fusion to investigate more possibilities in multi-stage fusions. Besides the limitation in their fusion strategy, another limitation is that many of them still suffer from quadratic or cubic computational complexity, which makes them almost infeasible for large-scale datasets. Recently some large-scale MVC methods have been proposed, among which the anchor-based methods have been one of the representative categories \cite{li15_mvsc,sun21_smvsc,wang22_tip,kang2020_lmvsc}. However, in terms of anchor selection, these anchor-based methods either learn a unified anchor set for all views \cite{li15_mvsc,sun21_smvsc,wang22_tip} or learn a separate anchor set for each view \cite{kang2020_lmvsc}. In terms of fusion stage, they still dwell in the single-stage fusion strategy. Moreover, for most of the previous MVC methods,  including the general-scale methods \cite{xia14_aaai, zhang15_iccv, liang19_icdm,liang20_tkde,liu19_pami,wang19_ijcai,kang20_nn} and the large-scale  methods \cite{li15_mvsc,sun21_smvsc,kang2020_lmvsc,zhang19_bmvc}, their requirement of dataset-specific hyperparameter-tuning also poses a major hurdle for their real-world applications.

\subsection{Ensemble Clustering}
\label{sec:related_work_ec}

The purpose of EC is to combine multiple base clusterings into a better and more robust consensus clustering \cite{Fred05_EAC,huang17_tcyb,topchy05,huang15_ecfg,Huang16_TKDE,ren17_kais,huang18_tsmcs,huang21_tcyb}. In the final stage of our FastMICE approach, multiple base clusterings are fused into a unified clustering result, which can be viewed as an EC process. Therefore, in this section, we will also review the related works on EC.

Previous EC methods can mostly be classified into three categories, i.e., the pair-wise co-occurrence based methods \cite{Fred05_EAC,huang17_tcyb}, the median partition based methods \cite{topchy05,huang15_ecfg}, and the graph partitioning based methods \cite{strehl02,ren17_kais,huang18_tsmcs,huang19_tkde}. The pair-wise co-occurrence based methods typically construct a co-association matrix by considering the pair-wise co-occurrence relationship in  base clusterings, and then perform some clustering algorithm on the co-association matrix to obtain the consensus clustering. Fred and Jain \cite{Fred05_EAC} proposed the evidence accumulation clustering method which imposes hierarchical agglomerative clustering on the co-association matrix. Huang et al. \cite{huang17_tcyb} refined the co-association matrix by an entropy based local weighting strategy and presented the locally weighted evidence accumulation method. The median partition based methods treat the EC problem as an optimization problem, which aims to find a median clustering by maximizing the similarity between this clustering and the base clusterings. Topchy et al. \cite{topchy05} cast the EC problem as a maximum-likelihood problem and solved it via the EM algorithm. Huang et al. \cite{huang15_ecfg} formulated the EC problem as a binary linear programming problem and solved it via the factor graph model. The graph partitioning based methods represent the multiple base clusterings as a graph structure and obtain the consensus clustering by  partitioning this graph. Strehl and Ghosh \cite{strehl02} considered the concept of hyper-edge and presented three graph partitioning algorithms. Ren et al. \cite{ren17_kais} took into account the importance of the objects and devised three graph partitioning based consensus functions for weighted-object ensemble clustering.

Despite the progress of these EC works \cite{Fred05_EAC,huang17_tcyb,topchy05,huang15_ecfg,Huang16_TKDE,ren17_kais,huang18_tsmcs,huang21_tcyb}, most of them are devised for single-view datasets and lack the consideration of multi-view scenarios. Recently Tao et al.  
\cite{tao17_ijcai} proposed a multi-view ensemble clustering (MVEC) method, which learns a consensus clustering from the multiple co-association matrices built in multiple views with low-rank and sparse constraints. Tao et al. \cite{tao20_tnnls} further incorporated marginalized denoising autoencoder into MVEC, and presented a marginalized multi-view ensemble clustering (M$^2$VEC) method. However, in MVEC and M$^2$VEC, the base clusterings in different views are generated separately, without leveraging multi-view complementariness in their ensemble generation. Moreover, the high computational complexity  also restricts their applications in large-scale scenarios.

\section{Proposed Framework}
\label{sec:framework}

In this section, we describe the proposed FastMICE approach in detail. Specifically, the notations are introduced in Section~\ref{sec:notations}. The formation of random view groups is provided in Section~\ref{sec:view_group_formation}. The view-sharing bipartite graph is described in Section~\ref{sec:view_sharing_bipartite_graph}. The generation of the diversified base clusterings is introduced in Section~\ref{sec:ensemble_generation}. Then the base clusterings are fused into a unified clustering via a highly efficient consensus function in Section~\ref{sec:consensus_function}. The time and space complexity of FastMICE is analyzed in Section~\ref{sec:complexity}.

\begin{table}[!t]
	\caption{Notations used throughout the paper}\vskip -0.15 in
	\label{table:notations}
	\begin{center}
		\begin{tabular}{p{2.1cm}<{\centering}|p{5.9cm}}
			\toprule
			$\mathcal{X}$   &A multi-view dataset\\
			$N$         &\# of samples in the dataset\\
			$V$			&\# of views in the dataset\\
			$View_v$			&The $v$-th view in the dataset\\
			$\mathcal{X}_v\in\mathbb{R}^{N\times d_v}$         &Data matrix associated with $View_v$\\
			$\mathcal{VG}$ 	&The set of view groups\\
			$M$ 		&\# of view groups\\
			$\mathcal{VG}^{(m)}$ &The $m$-th view group\\
			$V^{(m)}$ 		&\# of view members in $\mathcal{VG}^{(m)}$\\
			$View^{(m)}_v$ 		&The $v$-th view members in $\mathcal{VG}^{(m)}$\\
			$\tau^{(m)}_v$ 		&Feature sampling ratio for $View^{(m)}_v$\\
			\mbox{$\tilde{\mathcal{X}}^{(m)}_{v}\in\mathbb{R}^{N\times \tilde{d}^{(m)}_v}$} 
			&Data matrix after feature sampling in $View^{(m)}_v$\\
			$\tilde{x}^{(m)}_{v,i}$   &The $i$-th sample in $\tilde{\mathcal{X}}^{(m)}_{v}$\\
			$p$ 	&\# of anchors for each view group\\
			$\bar{p}^{(m)}$ &\# of anchors for a view member\\
			$K$ 	&\# of nearest neighbors for each view group\\
			$\bar{K}^{(m)}$ &\# of nearest neighbors for a view member\\
			$\mathcal{A}^{(m)}_v$  	&The anchor set in $View^{(m)}_v$\\
			$a^{(m)}_{v,i}$ 		&The $i$-th anchor in $\mathcal{A}^{(m)}_v$\\
			$\mathcal{G}^{(m)}_v$    &The bipartite sub-graph for $View^{(m)}_v$\\
			$\mathcal{L}^{(m)}_v$    &The left node set of $\mathcal{G}^{(m)}_v$\\
			$\mathcal{R}^{(m)}_v$    &The right node set of $\mathcal{G}^{(m)}_v$\\
			\mbox{$B^{(m)}_v$}    &The cross-affinity matrix of $\mathcal{G}^{(m)}_v$\\
			$\mathcal{G}^{(m)}$ & The view-sharing bipartite graph for $\mathcal{VG}^{(m)}$\\
			$\mathcal{L}^{(m)}$    &The left node set of $\mathcal{G}^{(m)}$\\
			$\mathcal{R}^{(m)}$    &The right node set of $\mathcal{G}^{(m)}$\\
			\mbox{$B^{(m)}$}    &The cross-affinity matrix of $\mathcal{G}^{(m)}$\\
			$E^{(m)}$         &Full affinity matrix of graph $\mathcal{G}^{(m)}$\\
			$L^{(m)}$         &Graph Laplacian of graph $\mathcal{G}^{(m)}$\\
			$D^{(m)}$         &Degree matrix of graph $\mathcal{G}^{(m)}$\\
			$\lambda^{(m)}_i$  &The $i$-th eigenvalue of graph $\mathcal{G}^{(m)}$\\
			$y^{(m)}_i$  &The $i$-th eigenvector of graph $\mathcal{G}^{(m)}$\\
			$\mathcal{G}^{(m)}_s$ & A small graph with $\mathcal{R}^{(m)}$ as the node set\\
			$E^{(m)}_s$         &Affinity matrix of graph $\mathcal{G}^{(m)}_s$\\
			$L^{(m)}_s$         &Graph Laplacian of graph $\mathcal{G}^{(m)}_s$\\
			$D^{(m)}_s$         &Degree matrix of graph $\mathcal{G}^{(m)}_s$\\
			$\delta^{(m)}_i$  &The $i$-th eigenvalue of graph $\mathcal{G}^{(m)}_s$\\
			$u^{(m)}_i$       &The $i$-th eigenvector of graph $\mathcal{G}^{(m)}_s$\\
			$\Pi$       &The ensemble of $M$ base clusterings\\
			$\pi^m$       &The $m$-th base clustering in $\Pi$\\
			$\mathcal{C}$ &The set of all clusters in $\Pi$\\
			$C_i$       &The $i$-th cluster in $\mathcal{C}$\\
			$k_c$       &\# of clusters in $\mathcal{C}$\\
			$\mathcal{G}$ &A bipartite graph between $\mathcal{X}$ and $\mathcal{C}$\\
			$B$         &Cross-affinity matrix of graph $\mathcal{G}$.\\
			$\mathcal{G}_s$ & A small graph with $\mathcal{C}$ as the node set\\
			$E_s$         &Affinity matrix of graph $\mathcal{G}_s$\\
			\bottomrule
		\end{tabular}
	\end{center}\vskip -0.1 in
\end{table}

\subsection{Notations}
\label{sec:notations}
Let $\mathcal{X} = \{x_1,x_2,\cdots,x_N\}$ be a dataset with $N$ data samples, where $x_i$ is the $i$-th sample. For a multi-view dataset, each data sample can be represented by features from different views.  Thus, the multi-view dataset can be denoted as
$\mathcal{X} = \{\mathcal{X}_1,\mathcal{X}_2,\cdots,\mathcal{X}_V\}$, where $\mathcal{X}_v\in\mathbb{R}^{N\times d_v}$ is the data matrix of the $v$-th view, $V$ is the number of views, and $d_v$ is the dimension of the $v$-th view. 
For convenience of the later view group formation, we denote the set of $V$ views as $\mathcal{V}=\{View_1,View_2,\cdots,View_V\}$, where $View_v$ is the $v$-th view in the dataset. Note that $\mathcal{X}_v\in\mathbb{R}^{N\times d_v}$ is the data matrix associated with the $v$-th view (i.e., $View_v$). 
For clarity, the notations used throughout the paper are given in Table~\ref{table:notations}.

The purpose of MVC is to fuse the information of multiple views for enhanced clustering. In large-scale multi-view scenarios, where the data size $N$ can be very large, it becomes a critical challenge how to  robustly fuse the multi-view information while ensuring high efficiency and practicality, which will be the focus of our following sections.

\subsection{Early-Stage View Group Formation}
\label{sec:view_group_formation}

An essential task of MVC is to fuse the information of multiple views for robust clustering result. The previous MVC algorithms differ from each other mainly in their fusion stages and fusion techniques \cite{liang19_icdm,liang20_tkde,wang19_ijcai,kang20_nn,liu21_icml}, but they generally have two characteristics in common. First, they tend to perform the multi-view fusion in a single stage, either early or late. Second, they often implicitly comply with a single-or-all paradigm, where each stage of them involves \emph{either} a single view \emph{or} all views. For example, in the late fusion algorithms \cite{wang19_ijcai,kang20_nn,liu21_icml}, each base clustering is  constructed on a \emph{single} view independently of other views, while the final fusion process utilizes the base clusterings from \emph{all} views in a one-shot manner. However, the vast middle ground between single-views-independently and all-views-together has rarely be explored by previous works.

Different from the conventional single-or-all paradigm, in the section, we present the concept of random view groups, each of which encapsules a random number of views and serves as a basic unit for the view-wise diversification and the hybrid early-late fusion in our FastMICE framework. Formally, let $\mathcal{VG}^{(m)}$ be the $m$-th view group, with a randomly selected subset of views, and  $V^{(m)}$ be the number of selected views in $\mathcal{VG}^{(m)}$. The number of views in each view group can be randomly chosen in the range of $[V_{min},V_{max}]$, where $V_{min}$ and  $V_{max}$ are respectively the lower bound and the upper bound of the number of selected views such that $1 \leq V_{min}\leq V_{max}\leq V$.

Each selected view in the view group is called a \emph{\textbf{view member}}. To enhance the diversity of view groups, we set the lower bound $V_{min}=1$ and the upper bound $V_{max}=V$, through which each view group can have at least one view member and at most $V$ view members. Thus, the $m$-th view group $\mathcal{VG}^{(m)}$ can be formed by randomly selected $V^{(m)}$ views from the set of all views, denoted as
\begin{align}
	\mathcal{VG}^{(m)} = \{View^{(m)}_1,View^{(m)}_2,\cdots,View^{(m)}_{V^{(m)}}\},
\end{align}
where $View^{(m)}_v$ is the $v$-th view member in $\mathcal{VG}^{(m)}$. The data matrix associated with $View^{(m)}_v$ is denoted as $\mathcal{X}^{(m)}_v\in\mathbb{R}^{N\times d^{(m)}_v}$, where $d^{(m)}_v$ is the dimension of this view member. By performing the randomization process repeatedly, a set of random view groups can be obtained, that is
\begin{align}
	\mathcal{VG} = \{\mathcal{VG}^{(1)},\mathcal{VG}^{(2)},\cdots,\mathcal{VG}^{(M)}\},
\end{align}
where $M$ is the number of the generated random view groups. As each view group leads to the generation of a base clustering, the number of view groups is also the number of base clusterings, also known as the ensemble size.

It is worth mentioning that the early-fusion MVC methods (as shown in Fig.~\ref{fig:compFusionEarly}) can be viewed as a special instance of our view group based formulation with all views selected into a single view group by setting $V_{min}=V_{max}=V$ and $M=1$. Similarly, the late fusion MVC methods (as shown in Fig.~\ref{fig:compFusionLate}) can also be viewed as special instance of our view group based formulation with each view being a single view group by setting $V_{min}=V_{max}=1$ and $M=V$.

\subsection{View-Sharing Bipartite Graph Construction}
\label{sec:view_sharing_bipartite_graph}

With multiple random view groups obtained, our next goal is to perform the early-stage fusion in the view groups. Note that the early-stage fusion is not meant to achieve a single optimal solution in a one-shot manner. Instead, it performs fusions on multiple view groups, and builds multiple view-sharing bipartite graphs with multi-level diversities.
Besides the diversity, to enable the scalability for very large datasets, the efficiency is another of our key concerns during the early-stage multi-fusion process.

Particularly, in each view group, the bipartite graph structure is exploited to formulate the information of multiple view member. In recent years, the bipartite graph structure has shown its advantage in handling large-scale datasets \cite{cai15_LSC,huang19_tkde,sun21_smvsc,wang22_tip}. From the perspective of topology, the bipartite graph is built between the $N$ data samples and a set of $p$ anchors (or representatives), typically with $p\ll N$ for large-scale datasets. From the perspective of matrix notation, the bipartite graph can be represented by an $N\times p$ cross-affinity matrix with its $(i,j)$-th entry being the affinity between the $i$-th sample and the $j$-th anchor, which can also be regarded as encoding the data samples via this small set of anchors.

To enhance the diversity of the bipartite graph construction in multiple view groups, we simultaneously leverage three levels of diversification, corresponding to the \textit{feature}-level, the \textit{anchor}-level, and the \textit{neighborhood}-level, respectively.

In the feature-level diversification, we first perform random feature sampling on each view member in a view group. Let $\tau^{(m)}_v$ denote the feature sampling ratio for  the $v$-th view member in the $m$-th view group (i.e., $View^{(m)}_v$).
To inject the diversity, the sampling ratio for each view member is randomly chosen in the range of $[\tau_{min}, \tau_{max}]$, where $\tau_{min}$ and  $\tau_{max}$ are respectively the lower and the upper bounds of the sampling ratio such that $0<\tau_{min}\leq \tau_{max}\leq 1$. 
By performing feature sub-sampling with a randomized sampling ratio, we can obtain a subset of features for each view member. For $View^{(m)}_v$, its data matrix \textit{after} random feature sampling can be denoted as $\tilde{\mathcal{X}}^{(m)}_{v}\in\mathbb{R}^{N\times \tilde{d}^{(m)}_v}$, where $\tilde{d}^{(m)}_v=\lceil \tau^{(m)}_v \cdot d^{(m)}_v\rceil$ is the reduced dimension, and $\lceil\cdot\rceil$ obtains the ceiling of a real value.

After random feature sampling, we proceed to construct a bipartite sub-graph on each view member, and then combine these bipartite sub-graphs into a unified view-sharing bipartite graph, where a set of $p$ anchors are required and the $K$-NN sparsification is performed.
In prior works, the anchor set for multiple views can be obtained by performing $k$-means clustering on the concatenated features of all views \cite{li15_mvsc} or by optimizing some objective function to learn consensus anchors \cite{wang22_tip}. However, on the one hand, they typically aim to find a set of anchors suitable for all views, without sufficient consideration to view-specific characteristics. On the other hand, their anchor selection or learning process may also be computationally expensive when facing very large datasets. 

Instead of pursuing a set of consensus anchors, we \emph{distribute} the task of finding $p$ anchors to the multiple view members. Without prior knowledge, we expect the multiple view members in the same view group to contribute equally. Specifically, each view member is expected to make a contribution of $\bar{p}^{(m)}=\lceil p/V^{(m)}\rceil$ anchors. Similarly, in terms of the neighborhood, each view member is expected to contribute $\bar{K}^{(m)}=\lceil K/V^{(m)}\rceil$ nearest neighbors. Thereafter, on each view member in $\mathcal{VG}^{(m)}$,  we aim to build a bipartite sub-graph between $N$ samples and $\bar{p}^{(m)}$ anchors, with each sample connected to $\bar{K}^{(m)}$ nearest anchors.

For the $m$-th view group $\mathcal{VG}^{(m)}$,  the anchor selection on the $v$-th view member $View^{(m)}_v$  (associated with the data matrix $\tilde{\mathcal{X}}^{(m)}_{v}$) is performed via the hybrid representative selection strategy \cite{huang19_tkde}, which efficiently obtains a set of $\bar{p}^{(m)}$ anchors, denoted as
\begin{align}
	\mathcal{A}^{(m)}_v = \{a^{(m)}_{v,1}, a^{(m)}_{v,2}, \cdots, a^{(m)}_{v,\bar{p}^{(m)}}\},
\end{align}
where $a^{(m)}_{v,i}\in\mathbb{R}^{\tilde{d}^{(m)}_v}$ is the $i$-th anchor selected from $\tilde{\mathcal{X}}^{(m)}_{v}$. 

Then  we define the bipartite sub-graph for the view member $View^{(m)}_v$ as follows:
\begin{align}
	\mathcal{G}^{(m)}_v = \{\mathcal{L}^{(m)}_v, \mathcal{R}^{(m)}_v, B^{(m)}_v\},
\end{align}
where $\mathcal{L}^{(m)}_v = \mathcal{X}$ and $\mathcal{R}^{(m)}_v = \mathcal{A}^{(m)}_v $ are the left and right node sets of the bipartite sub-graph, respectively, and $B^{(m)}_v\in\mathbb{R}^{N\times \bar{p}^{(m)}}$ is the cross-affinity matrix. An edge between two nodes exists if and only if one node is a data sample, another node is an anchor, and this anchor is one of the sample's $\bar{K}^{(m)}$-nearest anchors. Formally, the $(i,j)$-th entry of the cross-affinity matrix $B^{(m)}_v$ can be defined as
\begin{align}
	b^{(m)}_{v,ij}=\begin{cases}
		Sim(\tilde{x}^{(m)}_{v,i},a^{(m)}_{v,j}), &\text{if~}a^{(m)}_{v,j}\in \mathcal{N}_{\bar{K}^{(m)}}(\tilde{x}^{(m)}_{v,i}),\\
		0, &\text{otherwise},
	\end{cases}
\end{align}
where $\tilde{x}^{(m)}_{v,i}$ denotes the $i$-th sample in this view member (with reduced dimension),  $Sim(\cdot)$ computes the similarity between two vectors, and  $\mathcal{N}_K(x)$ denotes the set of $K$-nearest anchors of sample $x$. Note that $Sim(\cdot)$ can be any similarity measure. Typically, we utilize the Gaussian kernel similarity, which maps the Euclidean distance to a similarity measure via a Gaussian kernel. Thus, with each sample linked to $\bar{K}^{(m)}$ nearest anchors, the cross-affinity matrix $B^{(m)}_v$ can be represented as a sparse matrix with only $N\cdot \bar{K}^{(m)}$ non-zeros entries, whose sparsity can significantly benefit the later matrix computations.

For the $v$-th view member in the $m$-th view group, a bipartite sub-graph $\mathcal{G}^{(m)}_v$ (with the cross-affinity matrix $B^{(m)}_v$) can be obtained. From the perspective of information encoding, 
the cross-affinity matrix $B^{(m)}_v$ can be regarded as a representation matrix to encode the $N$ data samples by the $\bar{p}^{(m)}$ anchors, with each row of $B^{(m)}_v$ being a low-dimensional feature vector for a data sample. As there are $V^{(m)}$ view members in $\mathcal{VG}^{(m)}$, each data sample can be represented (or encoded) by the totally $\bar{p}^{(m)}\cdot V^{(m)}$ anchors, which lead to the view-sharing bipartite graph for the entire view group, that is
\begin{align}
	\mathcal{G}^{(m)} = \{\mathcal{L}^{(m)}, \mathcal{R}^{(m)}, B^{(m)}\},
\end{align}
where $\mathcal{L}^{(m)} = \mathcal{X}$ is the left node set of the bipartite graph, and $\mathcal{R}^{(m)} = \mathcal{A}^{(m)}_1 \bigcup, \cdots, \bigcup \mathcal{A}^{(m)}_{V^{(m)}}$ is the right node set, which is the union of the anchor sets of the $V^{(m)}$ view members from $\mathcal{VG}^{(m)}$. The cross-affinity matrix for $\mathcal{G}^{(m)}$ is defined as $B^{(m)} = [\bar{B}^{(m)}_1,...,\bar{B}^{(m)}_{V^{(m)}}]$, where $\bar{B}^{(m)}_v$ is computed by normalizing each row of $B^{(m)}_v$ to unit norm, so as to adjust multiple bipartite sub-graphs into similar scales.

The time complexity of building the view-sharing bipartite graph $\mathcal{G}^{(m)}$ mainly comes from the anchor selection and the cross-affinity matrix construction. For the $v$-th view member in $\mathcal{VG}^{(m)}$, the anchor selection via hybrid representative selection takes $O((\bar{p}^{(m)})^2\tilde{d}^{(m)}_v t)$ time, where $t$ is the number of iterations. Then it takes $O(N(\bar{p}^{(m)})^{1/2})$ time to build the cross-affinity matrix $B^{(m)}_v$ via approximate $K$-NN computation \cite{huang19_tkde}. Thus the construction of a bipartite sub-graph (for a view member) takes $O(N(\bar{p}^{(m)})^{1/2}+(\bar{p}^{(m)})^2\tilde{d}^{(m)}_v t)$ time,
 which is dominated by  $O(N(\bar{p}^{(m)})^{1/2})$ with $\bar{p}^{(m)}\ll N$. The construction of a view-sharing bipartite graph for $\mathcal{VG}^{(m)}$ involves the construction of $V^{(m)}$ bipartite sub-graphs, which in total take $O(N(\bar{p}^{(m)})^{1/2}V^{(m)})$ time. With $\bar{p}^{(m)}=\lceil p/V^{(m)}\rceil$ and $V^{(m)}\leq V$, it is easy to know that $\bar{p}^{(m)}V^{(m)}\in [p,p+V^{(m)})$, leading to $(\bar{p}^{(m)})^{1/2}V^{(m)}=(\bar{p}^{(m)}(V^{(m)})^2)^{1/2}<((p+V)V)^{1/2}$.  With $V\ll p$, the time complexity of constructing a view-sharing bipartite graph can be obtained as $O(Np^{1/2}V^{1/2})$.

As $\bar{B}^{(m)}_v$ is a sparse matrix with $N\cdot \bar{K}^{(m)}$ non-zero entries and $B^{(m)}$ is a sparse matrix with $N\cdot \bar{K}^{(m)}\cdot V^{(m)}$ non-zero entries, the space complexity of constructing a view-sharing bipartite graph is $O(N\bar{K}^{(m)}V^{(m)})$. With $\bar{K}^{(m)}=\lceil K/V^{(m)}\rceil$ and $V^{(m)} \leq V$, we have $\bar{K}^{(m)}V^{(m)}\in [K, K+V)$. Therefore the space complexity can be written as $O(N(K+V))$.

\subsection{Ensemble Generation in View Groups}
\label{sec:ensemble_generation}

In this section, we describe the ensemble generation, i.e., the generation of a set of diversified base clusterings, based on the view-sharing bipartite graphs in multiple view groups. 

Specifically, with a view-sharing bipartite graph constructed for each view group, a set of $M$ view-sharing bipartite graphs, ranging from $\mathcal{G}^{(1)}$ to $\mathcal{G}^{(M)}$, can be built for the $M$ view groups. 
For the base clustering in each view group, the number of clusters, say, $k^{(m)}$, will be randomly selected in the range of $[k_{min},k_{max}]$, where $k_{min}$ and  $k_{max}$ are respectively the lower bound and the upper bound of the cluster number. In the following, we proceed to describe the fast partitioning of the $m$-th view-sharing bipartite graph into $k^{(m)}$ clusters.

For the $m$-th view group, there are totally $N+\bar{p}^{(m)}\cdot V^{(m)}$ nodes in the view-sharing bipartite graph $\mathcal{G}^{(m)}$, with $N$ data samples in the left node set and $\bar{p}^{(m)} \cdot V^{(m)}$ anchors in the right node set. By treating $\mathcal{G}^{(m)}$ as a general graph, its full affinity matrix $E^{(m)}\in\mathbb{R}^{(N+\bar{p}^{(m)}V^{(m)})\times (N+\bar{p}^{(m)}V^{(m)})}$ can be represented as
\begin{align}
	E^{(m)} = \begin{bmatrix}
		0 & B^{(m)}\\
		{(B^{(m)})}^{\top}& 0
	\end{bmatrix}
\end{align}
To partition this graph via conventional spectral clustering \cite{ng2002spectral}, the following generalized eigen-decomposition problem should be solved:
\begin{align}
	\label{eq:eigen_prob_1}
	L^{(m)}y^{(m)} = \lambda^{(m)} D^{(m)}y^{(m)},
\end{align}
where $L^{(m)}=D^{(m)}-E^{(m)}$ is the graph Laplacian, $D^{(m)}$ is the degree matrix with its $(i,i)$-th entry being the sum of the $i$-row in $E^{(m)}$. It takes $O((N+\bar{p}^{(m)}V^{(m)})^3)$ time to solve this problem via SVD, which is computationally expensive for large-scale datasets. 

Due to the imbalanced sizes of the left and right node sets in the bipartite graph, with $\bar{p}^{(m)}V^{(m)}\approx p\ll N$, the eigen-decomposition problem in Eq.~(\ref{eq:eigen_prob_1}) can be reduced to  an eigen-decomposition problem on a smaller graph $\mathcal{G}^{(m)}_s$ with a node set of $\bar{p}^{(m)}V^{(m)}$ anchors and an affinity matrix of $E_s^{(m)}=(B^{(m)})^{\top}(\hat{D}^{(m)})^{-1}B^{(m)}$, whose computation takes $O(NK^2)$ time, where $\hat{D}^{(m)}$ is a diagonal matrix with its $(i,i)$-th entry being the sum of the $i$-th row in the cross-affinity matrix $B^{(m)}$. The generalized eigen-decomposition problem on the reduced graph $\mathcal{G}^{(m)}_s$ can be formulated as
\begin{align}
	\label{eq:eigen_prob_2}
	L_s^{(m)}u^{(m)} = \delta^{(m)} D^{(m)}_su^{(m)},
\end{align}
where $L_s^{(m)}=D_s^{(m)}-E_s^{(m)}$ is the graph Laplacian, $D_s^{(m)}$ is the degree matrix with its $(i,i)$-th entry being the sum of the $i$-th row in $E_s^{(m)}$. 

By solving the eigen-decomposition problem in Eq.~(\ref{eq:eigen_prob_2}) with $O((\bar{p}^{(m)}V^{(m)})^3)\approx{O(p^3)}$ time, the first $k^{(m)}$ eigen-pairs can be obtained, denoted as $\{\delta^{(m)}_i,u^{(m)}_i\}_{i=1}^{k^{(m)}}$ with $0=\delta^{(m)}_1\leq \delta^{(m)}_2 \leq \cdots \leq \delta^{(m)}_{k^{(m)}}<1$. Let  $\{\lambda^{(m)}_i,y^{(m)}_i\}_{i=1}^{k^{(m)}}$ be the first $k$ eigen-pairs for the graph $\mathcal{G}^{(m)}$ with $0=\lambda^{(m)}_1\leq \lambda^{(m)}_2 \leq \cdots \leq \lambda^{(m)}_{k^{(m)}}<1$, which can be efficiently computed from $\{\delta^{(m)}_i,u^{(m)}_i\}_{i=1}^{k^{(m)}}$ with the following properties \cite{CVPR12_Li}:
\begin{align}
	\label{eq:tcut_property1}
	&\lambda^{(m)}_i (2-\lambda^{(m)}_i) = \delta^{(m)}_i,\\
	\label{eq:tcut_property2}
	&y^{(m)}_i = \begin{bmatrix}
		h^{(m)}_i\\
		u^{(m)}_i 
	\end{bmatrix}
\end{align}
where $h^{(m)}_i = (\hat{D}^{(m)})^{-1}B^{(m)}u^{(m)}_i/(1-\lambda^{(m)}_i)$. 

The time complexity of solving the eigen-decomposition problem in Eq.~(\ref{eq:eigen_prob_2}) is $O(p^3)$. Due to the sparsity of $B^{(m)}$, it takes $O(NK)$ time to compute $h^{(m)}_i$ and obtain each eigenvector $\{y^{(m)}_i\}_{i=1}^{k^{(m)}}$ from $\{u^{(m)}_i\}_{i=1}^{k^{(m)}}$. Then, the $k^{(m)}$ eigenvectors will be stacked as a matrix with each row treating as a new feature vector, upon which the $k$-means discretization can be performed to obtain the $m$-th base clustering with $O(N(k^{(m)})^2t)$ time. Since $k^{(m)}$ and $k$ are generally at  similar scales, the time complexity of generating the base clustering from a view-sharing bipartite graph is $O(N(k^2t+K^2+Kk)+p^3)$, and the space complexity is $O(N(k+K))$.

\subsection{Late-Stage Consensus Function}
\label{sec:consensus_function}

By partitioning the view-sharing bipartite graph of the $m$-th view group, we obtain a base clustering with $k^{(m)}$ clusters. Then the set of base clusterings generated for the $M$ view groups can be represented as 
\begin{align}
	\Pi = \{\pi^{(1)},\pi^{(2)},\cdots,\pi^{(M)}\},
\end{align}
where
$ \pi^{(m)} = \{C^{(m)}_1,C^{(m)}_2,\cdots,C^{(m)}_{k^{(m)}}\}$
is the $m$-th base clustering, and $C^{(m)}_i$ is the $i$-th cluster in $\pi^{(m)}$. Each base clustering consists of a certain number of clusters. For convenience, we represent the set of clusters in \emph{all} base clusterings as
\begin{align}
	\mathcal{C} = \{C_1,C_2,\cdots,C_{k_c}\},
\end{align}
where $C_i$ is the $i$-th cluster, and $k_c=\sum_{m=1}^M k^{(m)}$ is the total number of clusters in all base clusterings. 

To jointly formulate the information of multiple base clusterings, we construct a unified bipartite graph by treating both the data samples and the base clusters as graph nodes, denoted as
$\mathcal{G} = \{\mathcal{L}, \mathcal{R}, B\}$,
where $\mathcal{L}=\mathcal{X}$ is the left node set with $N$ data samples, $\mathcal{R}=\mathcal{C}$ is the right node set with $k_c$ base clusters, and $B\in\mathbb{R}^{N\times k_c}$ is the cross-affinity matrix. A link between two nodes exists if and only if one of them is a data sample and the other one is a base cluster that contains the sample. Thus, the $(i,j)$-th entry of the cross-affinity matrix $B$ can be defined as
\begin{align}
	b_{ij}=\begin{cases}
		1, &\text{if~}x_i\in C_j,\\
		0, &\text{otherwise},
	\end{cases}
\end{align}
Note that each data sample belongs to one and only one cluster  in a base clustering. With a total of $M$ base clusterings, each data sample will be linked to exactly $M$ clusters in the unified bipartite graph. That is, in each row of $B$, there are exactly $M$ non-zero entries. Therefore, $B$ is a matrix with $N\cdot M$ non-zero entries.

Similar to the partitioning of the view-sharing bipartite graph $\mathcal{G}^{(m)}$, the eigen-decomposition problem of the unified bipartite graph $\mathcal{G}$ can also be tackled by conducting the eigen-decomposition on a smaller graph $\mathcal{G}_s$ with an affinity matrix $E_s=B^{\top}\hat{D}^{-1}B$, whose computation takes $O(NM^2)$ time, where $\hat{D}$ is a diagonal matrix with its $(i,i)$-th entry being the sum of the $i$-th row in $B$.

To obtain the final clustering result with $k$ clusters, solving the eigen-decomposition of the graph $\mathcal{G}_s$ takes $O({k_c}^3)$ time. Then it takes $O(NM)$ time to compute \emph{each} of the $k$ eigenvectors of  $\mathcal{G}$ from the eigenvectors of $\mathcal{G}_s$. Thereafter, it takes $O(Nk^2t)$ time to perform the $k$-means discretization. Thus, the time complexity of the overall consensus function with the unified bipartite graph is $O(N(k^2t+M^2+Mk)+{k_c}^3)$, and the space complexity is $O(N(k+M))$.

\subsection{Complexity Analysis}
\label{sec:complexity}

In the following, we will analyze the time and space complexity of the proposed FastMICE approach.

\begin{table*}[!t]
	\centering
	\caption{Description of the benchmark datasets.}\vskip -0.15 in
	\label{table:datasets}
	\begin{center}
		\renewcommand\arraystretch{1}
		\begin{tabular}{p{1cm}<{\centering}|p{2cm}<{\centering}|p{1cm}<{\centering}p{1cm}<{\centering}p{1cm}<{\centering}p{7cm}<{\centering}}
			\toprule
			\multicolumn{2}{c|}{Dataset}         &\#Sample     &\#View      &\#Class     &Dimension\\
			\midrule
			\multirow{10}{*}{\emph{General}}
			&\emph{MSRCv1} 	&210 	&4 		&7		&CM (24), GIST (512), LBP (256), GENT (254)\\
			&\emph{Yale}		&165		&3		&15		&Intensity 4,096, LBP (3,304), Gabor (6,750)\\
			&\emph{ORL}		&400	&3		&40 		&Intensity (4,096), LBP (3,304), Gabor (6,750)\\
			&\emph{Movies}	&617		&2	&17	&Keywords (1,878), Actors (1,398)\\
			&\emph{BBCSport}		&544		&2		&5		&View1 (3,183), View2 (3,203)\\
			&\emph{NH-550}		&550		&3		&5		&Intensity (2,000), LBP (3,304), Gabor (6,750)\\
			&\emph{Out-Scene}	&2,688	&4	&8	&Color (432), GIST (512), HOG (256), LBP (48)\\
			&\emph{Citeseer}		&3,312		&2		&6		&Content (3,703), Citation (3,312)\\
			&\emph{NH-4660}		&4,660		&3		&5		&Intensity (2,000), LBP (3,304), Gabor (6,750)\\
			&\emph{ALOI}		&10,800		&4		&100		&Similarity (77), Haralick (13), HSV (64), RGB (125)\\
			\midrule
			\multirow{12}{*}{\emph{Large}}
			&\emph{NUS-WIDE} 	&30,000	&5 	&31 	&CH (64), CM (225), CORR (144), EDH (73), WT(128)\\
			&\emph{VGGFace2-50}		&34,027		&4		&50		&LBP (944), HOG (576), GIST(512), Gabor (640)\\
			&\emph{VGGFace2-100}		&65,342		&4		&100		&LBP (944), HOG (576), GIST(512), Gabor (640)\\
			&\emph{VGGFace2-200}		&124,880		&4		&200		&LBP (944), HOG (576), GIST(512), Gabor (640)\\
			&\emph{CIFAR-10}		&60,000		&4		&10		&LBP (944), HOG (576), GIST(512), Gabor (640)\\
			&\emph{CIFAR-100}		&60,000		&4		&100		&LBP (944), HOG (576), GIST(512), Gabor (640)\\
			&\emph{YTF-10}		&38,654		&4		&10		&LBP (944), HOG (576), GIST(512), Gabor (640)\\
			&\emph{YTF-20}		&63,896		&4		&20		&LBP (944), HOG (576), GIST(512), Gabor (640)\\
			&\emph{YTF-50}		&126,054		&4		&50		&LBP (944), HOG (576), GIST(512), Gabor (640)\\
			&\emph{YTF-100}		&195,537		&4		&100		&LBP (944), HOG (576), GIST(512), Gabor (640)\\
			&\emph{YTF-200}		&286,006		&4		&200		&LBP (944), HOG (576), GIST(512), Gabor (640)\\
			&\emph{YTF-400}		&398,191		&4		&400		&LBP (944), HOG (576), GIST(512), Gabor (640)\\
			\bottomrule
		\end{tabular} 
	\end{center}
\end{table*}

\subsubsection{Time Complexity}
This section analyzes the time complexity of FastMICE. The generation of random view groups in Section~\ref{sec:view_group_formation} takes $O(MV)$ time. Then it takes $O(Np^{1/2}V^{1/2})$ time to construct a view-sharing bipartite graph in each view group, and totally $O(NMp^{1/2}V^{1/2})$ time to construct the view-sharing bipartite graphs in all the $M$ view groups in Section~\ref{sec:view_sharing_bipartite_graph}. Thereafter, the ensemble generation  in Section~\ref{sec:ensemble_generation} (to produce $M$ base clusterings) takes $O(NM(k^2t+K^2+Kk)+Mp^3)$ time. Finally, the consensus function in Section~\ref{sec:consensus_function} takes $O(N(k^2t+M^2+Mk)+{k_c}^3)$ time. 

In large-scale scenarios,  the number of anchors may range from hundreds to thousands, which is much smaller than the data size $N$, but much larger than the number of clusters $k$ or the number of nearest neighbors $K$. With $k,K,M,V\ll p \ll N$, the overall time complexity of the FastMICE approach can be written as $O(NMp^{1/2}V^{1/2})$, which is linear to the data size $N$.

\subsubsection{Space Complexity}
This section analyzes the space complexity of FastMICE. The construction of a view-sharing bipartite graph takes $O(N(K+V))$ space, while its partitioning takes $O(N(k+K))$ space. Since the base clusterings can be generated in a serial processing manner, the space complexity of generating multiple base clusterings is still $O(N(k+K+V))$. The consensus function takes $O(N(k+M))$ space.  Thus, the overall space complexity of FastMICE is $O(N(k+K+V+M))$, which is also linear to the data size $N$.

\section{Experiments}
\label{sec:experiments}

In this section, we evaluate the proposed FastMICE approach against the state-of-the-art MVC approaches on a variety of general-scale and large-scale multi-view datasets. The experiments are conducted on a PC with an Intel i5-6600 CPU and 16GB of RAM.

\subsection{Benchmark Datasets}

In the experiments, 22 real-world multi-view datasets are used, including 10 general-scale datasets and 12 large-scale datasets (as shown in Table~\ref{table:datasets}). 

Specifically, the 10 general-scale datasets are
\emph{MSRCv1} \cite{chen21_if},
\emph{Yale}  \cite{chen21_if}, 
\emph{ORL} \cite{chen21_if},
\emph{Movies}  \cite{xie20_tkde},
\emph{BBCSport} \cite{ch20_tcyb}, 
\emph{NH-550} \cite{ch20_tcyb},
\emph{Out-Scene} \cite{hu20_if},
\emph{Citeseer} \cite{liu21_icml}, 
\emph{NH-4660} \cite{cao15_cvpr},
and \emph{ALOI} \cite{aloi05},
and the 12 large-scale datasets are
\emph{NUS-WIDE} \cite{zhang19_bmvc},
\emph{VGGFace2-50},
\emph{VGGFace2-100},
\emph{VGGFace2-200},
\emph{CIFAR-10},
\emph{CIFAR-100},
\emph{YTF-10}	,
\emph{YTF-20},
\emph{YTF-50},
\emph{YTF-100},
\emph{YTF-200},
and \emph{YTF-400}.
Note that \emph{VGGFace2-50}, \emph{VGGFace2-100}, and \emph{VGGFace2-200} are three versions of the 
\emph{VGGFace2}\footnote{\url{www.robots.ox.ac.uk/~vgg/data/vgg_face2/}} dataset. 
\emph{CIFAR-10} and 
\emph{CIFAR-100} are two versions of the \emph{CIFAR}\footnote{\url{https://www.cs.toronto.edu/~kriz/cifar.html}} dataset. 
And \emph{YTF-10}	,
\emph{YTF-20},
\emph{YTF-50},
\emph{YTF-100},
\emph{YTF-200},
and \emph{YTF-400} are six versions of the \emph{YouTube-Faces}\footnote{\url{https://www.cs.tau.ac.il/~wolf/ytfaces/}} (YTF) dataset.
The purpose of testing different versions (of different data sizes) of these large-scale datasets is to better evaluate the MVC algorithms with different levels of scalability. The details of the datasets are given in Table~\ref{table:datasets}.

\begin{table*}[!t]\scriptsize
	\centering
	\caption{Average NMI(\%) scores over 20 runs by different multi-view clustering algorithms.
		On each dataset, the best two scores are highlighted in \textbf{bold}, while the best one in [\textbf{bold and brackets}].}\vskip -0.1 in
	\label{table:compare_nmi}
	\begin{threeparttable}
		\renewcommand\arraystretch{1.02}
		\begin{tabular}{|m{1.56cm}<{\centering}|m{1cm}<{\centering}m{1cm}<{\centering}m{0.92cm}<{\centering}m{1.02cm}<{\centering}m{0.92cm}<{\centering}m{1cm}<{\centering}m{1.02cm}<{\centering}m{0.92cm}<{\centering}m{0.9cm}<{\centering}m{1.468cm}<{\centering}|m{1.36cm}<{\centering}|}
			\hline
			Dataset    &MVSC     &AMGL  &SwMC &MVEC     &M$^2$VEC$_{km}$     &M$^2$VEC$_{spec}$     &BMVC   &LMVSC  &SMVSC   &FPMVS-CAG  &FastMICE\\
			\hline
			\hline
			\emph{MSRCv1}	&49.34$_{\pm3.78}$	&58.84$_{\pm6.33}$	&\textbf{62.69}$_{\pm0.00}$	&62.44$_{\pm1.99}$	&59.88$_{\pm1.79}$	&60.09$_{\pm2.27}$	&53.62$_{\pm0.00}$	&27.94$_{\pm0.00}$	&61.09$_{\pm0.00}$	&56.62$_{\pm0.00}$	&[\textbf{71.04}$_{\pm3.77}$]\\
			\hline
			\emph{Yale}	&64.10$_{\pm3.25}$	&61.78$_{\pm2.21}$	&\textbf{66.28}$_{\pm0.00}$	&64.61$_{\pm1.11}$	&48.48$_{\pm9.28}$	&50.08$_{\pm3.73}$	&27.57$_{\pm0.00}$	&61.64$_{\pm0.00}$	&61.68$_{\pm0.00}$	&49.76$_{\pm0.00}$	&[\textbf{67.85}$_{\pm2.58}$]\\
			\hline
			\emph{ORL}	&84.69$_{\pm2.06}$	&\textbf{85.29}$_{\pm1.78}$	&83.31$_{\pm0.00}$	&83.29$_{\pm1.04}$	&77.59$_{\pm1.94}$	&75.16$_{\pm2.93}$	&39.29$_{\pm0.00}$	&76.42$_{\pm0.00}$	&75.26$_{\pm0.00}$	&74.30$_{\pm0.00}$	&[\textbf{87.01}$_{\pm1.15}$]\\
			\hline
			\emph{Movies}	&22.85$_{\pm1.50}$	&22.72$_{\pm0.88}$	&16.75$_{\pm0.00}$	&25.03$_{\pm1.15}$	&28.27$_{\pm0.42}$	&28.19$_{\pm1.12}$	&[\textbf{29.07}$_{\pm0.00}$]	&24.65$_{\pm0.00}$	&16.76$_{\pm0.00}$	&14.51$_{\pm0.00}$	&\textbf{28.59}$_{\pm1.33}$\\
			\hline
			\emph{BBCSport}	&69.29$_{\pm14.02}$	&87.32$_{\pm6.91}$	&46.93$_{\pm0.00}$	&[\textbf{90.57}$_{\pm1.16}$]	&\textbf{89.30}$_{\pm1.29}$	&89.18$_{\pm1.37}$	&75.97$_{\pm0.00}$	&82.88$_{\pm0.00}$	&14.83$_{\pm0.00}$	&10.37$_{\pm0.00}$	&88.95$_{\pm2.84}$\\
			\hline
			\emph{NH-550}	&77.09$_{\pm8.11}$	&59.40$_{\pm12.08}$	&82.04$_{\pm0.00}$	&\textbf{84.93}$_{\pm3.65}$	&79.62$_{\pm5.17}$	&79.22$_{\pm4.81}$	&4.14$_{\pm0.00}$	&72.59$_{\pm0.00}$	&82.49$_{\pm0.00}$	&71.35$_{\pm0.00}$	&[\textbf{93.17}$_{\pm4.99}$]\\
			\hline
			\emph{OutdoorScene}	&23.39$_{\pm14.73}$	&44.19$_{\pm3.96}$	&43.69$_{\pm0.00}$	&48.90$_{\pm0.55}$	&47.99$_{\pm0.42}$	&47.84$_{\pm0.59}$	&\textbf{53.66}$_{\pm0.00}$	&45.97$_{\pm0.00}$	&51.67$_{\pm0.00}$	&53.04$_{\pm0.00}$	&[\textbf{58.38}$_{\pm1.29}$]\\
			\hline
			\emph{Citeseer}	&0.40$_{\pm0.11}$	&0.53$_{\pm0.06}$	&1.22$_{\pm0.00}$	&29.12$_{\pm0.42}$	&\textbf{29.69}$_{\pm0.48}$	&28.56$_{\pm0.36}$	&25.29$_{\pm0.00}$	&8.75$_{\pm0.00}$	&17.86$_{\pm0.00}$	&14.73$_{\pm0.00}$	&[\textbf{31.71}$_{\pm2.03}$]\\
			\hline
			\emph{NH-4660}	&53.98$_{\pm19.66}$	&8.19$_{\pm2.82}$	&8.77$_{\pm0.00}$	&85.01$_{\pm4.17}$	&\textbf{85.05}$_{\pm0.38}$	&84.11$_{\pm2.91}$	&74.13$_{\pm0.00}$	&68.06$_{\pm0.00}$	&66.64$_{\pm0.00}$	&67.15$_{\pm0.00}$	&[\textbf{93.39}$_{\pm5.58}$]\\
			\hline
			\emph{ALOI}	&30.48$_{\pm2.14}$	&68.07$_{\pm1.72}$	&N/A	&N/A	&N/A	&N/A	&\textbf{75.30}$_{\pm0.00}$	&74.95$_{\pm0.00}$	&57.38$_{\pm0.00}$	&55.51$_{\pm0.00}$	&[\textbf{83.39}$_{\pm0.75}$]\\
			\hline
			\hline
			\emph{NUS-WIDE}	&{N/A}	&{N/A}	&{N/A}	&{N/A}	&{N/A}	&{N/A}	&[\textbf{15.48}$_{\pm0.00}$]	&8.55$_{\pm0.00}$	&11.43$_{\pm0.00}$	&12.32$_{\pm0.00}$ &\textbf{13.40}$_{\pm0.48}$\\
			\hline
			\emph{VGGFace2-50}	&N/A	&N/A	&N/A	&N/A	&N/A	&N/A	&\textbf{13.64}$_{\pm0.00}$	&13.62$_{\pm0.00}$	&11.16$_{\pm0.00}$	&12.33$_{\pm0.00}$	&[\textbf{14.75}$_{\pm0.34}$]\\
			\hline
			\emph{VGGFace2-100}	&N/A	&N/A	&N/A	&N/A	&N/A	&N/A	&14.30$_{\pm0.00}$	&\textbf{15.15}$_{\pm0.00}$	&11.93$_{\pm0.00}$	&11.82$_{\pm0.00}$	&[\textbf{15.71}$_{\pm0.29}$]\\
			\hline
			\emph{VGGFace2-200}	&{N/A}	&{N/A}	&{N/A}	&{N/A}	&{N/A}	&{N/A}	&14.42$_{\pm0.00}$	&\textbf{15.35}$_{\pm0.00}$	&11.13$_{\pm0.00}$	&9.58$_{\pm0.00}$	&[\textbf{15.50}$_{\pm0.36}$]\\
			\hline
			\emph{CIFAR-10}	&N/A	&N/A	&N/A	&N/A	&N/A	&N/A	&17.51$_{\pm0.00}$	&13.54$_{\pm0.00}$	&18.37$_{\pm0.00}$	&\textbf{18.41}$_{\pm0.00}$	&[\textbf{21.42}$_{\pm0.52}$]\\
			\hline
			\emph{CIFAR-100}	&N/A	&N/A	&N/A	&N/A	&N/A	&N/A	&\textbf{15.34}$_{\pm0.00}$	&14.50$_{\pm0.00}$	&13.93$_{\pm0.00}$	&13.75$_{\pm0.00}$	&[\textbf{18.37}$_{\pm0.34}$]\\
			\hline
			\emph{YTF-10}	&N/A	&N/A	&N/A	&N/A	&N/A	&N/A	&\textbf{79.80}$_{\pm0.00}$	&71.70$_{\pm0.00}$	&77.91$_{\pm0.00}$	&76.86$_{\pm0.00}$	&[\textbf{81.69}$_{\pm2.72}$]\\
			\hline
			\emph{YTF-20}	&N/A	&N/A	&N/A	&N/A	&N/A	&N/A	&74.17$_{\pm0.00}$	&75.06$_{\pm0.00}$	&\textbf{79.81}$_{\pm0.00}$	&77.82$_{\pm0.00}$	&[\textbf{80.03}$_{\pm1.73}$]\\
			\hline
			\emph{YTF-50}	&N/A	&N/A	&N/A	&N/A	&N/A	&N/A	&\textbf{82.29}$_{\pm0.00}$	&80.92$_{\pm0.00}$	&80.59$_{\pm0.00}$	&79.59$_{\pm0.00}$	&[\textbf{83.15}$_{\pm0.78}$]\\
			\hline
			\emph{YTF-100}	&N/A	&N/A	&N/A	&N/A	&N/A	&N/A	&\textbf{82.97}$_{\pm0.00}$	&79.62$_{\pm0.00}$	&75.97$_{\pm0.00}$	&67.27$_{\pm0.00}$	&[\textbf{83.09}$_{\pm0.69}$]\\
			\hline
			\emph{YTF-200}	&N/A	&N/A	&N/A	&N/A	&N/A	&N/A	&N/A	&\textbf{79.05}$_{\pm0.00}$	&70.64$_{\pm0.00}$	&65.10$_{\pm0.00}$	&[\textbf{82.01}$_{\pm0.23}$]\\
			\hline
			\emph{YTF-400}	&N/A	&N/A	&N/A	&N/A	&N/A	&N/A	&N/A	&N/A	&N/A	&N/A	&[\textbf{79.81}$_{\pm0.25}$]\\
			\hline
			\hline
			Average Rank	&6.55	&6.27	&6.14	&4.77	&5.32	&5.68	&4.41	&4.95	&4.77	&5.82	&1.23\\
			\hline
		\end{tabular}
		\begin{tablenotes}
			\item[*] Note that N/A indicates the out-of-memory error.
		\end{tablenotes}
	\end{threeparttable}\vskip -0.05 in
\end{table*}

\begin{table*}[!t]\scriptsize
	\centering
	\caption{Average ARI(\%) scores over 20 runs by different multi-view clustering algorithms.
		On each dataset, the best two scores are highlighted in \textbf{bold}, while the best one in [\textbf{bold and brackets}].}\vskip -0.1 in
	\label{table:compare_ari}
	\begin{threeparttable}
		\renewcommand\arraystretch{1.02}
		\begin{tabular}{|m{1.56cm}<{\centering}|m{1cm}<{\centering}m{1cm}<{\centering}m{0.92cm}<{\centering}m{1.02cm}<{\centering}m{0.92cm}<{\centering}m{1cm}<{\centering}m{1.02cm}<{\centering}m{0.92cm}<{\centering}m{0.9cm}<{\centering}m{1.468cm}<{\centering}|m{1.36cm}<{\centering}|}
			\hline
			Dataset    &MVSC     &AMGL  &SwMC &MVEC     &M$^2$VEC$_{km}$     &M$^2$VEC$_{spec}$     &BMVC   &LMVSC  &SMVSC   &FPMVS-CAG  &FastMICE\\
			\hline
			\hline
			\emph{MSRCv1}	&37.89$_{\pm4.62}$	&46.27$_{\pm9.59}$	&52.37$_{\pm0.00}$	&\textbf{54.81}$_{\pm2.44}$	&49.90$_{\pm2.06}$	&49.93$_{\pm2.36}$	&42.16$_{\pm0.00}$	&14.82$_{\pm0.00}$	&50.18$_{\pm0.00}$	&43.31$_{\pm0.00}$	&[\textbf{65.49}$_{\pm5.21}$]\\
			\hline
			\emph{Yale}	&40.15$_{\pm5.28}$	&34.35$_{\pm4.13}$	&\textbf{44.82}$_{\pm0.00}$	&42.67$_{\pm1.60}$	&20.39$_{\pm9.86}$	&23.60$_{\pm4.62}$	&1.77$_{\pm0.00}$	&37.25$_{\pm0.00}$	&37.56$_{\pm0.00}$	&25.30$_{\pm0.00}$	&[\textbf{46.65}$_{\pm3.90}$]\\
			\hline
			\emph{ORL}	&\textbf{56.95}$_{\pm6.96}$	&51.86$_{\pm6.76}$	&41.39$_{\pm0.00}$	&50.26$_{\pm4.93}$	&33.40$_{\pm7.19}$	&35.11$_{\pm5.33}$	&0.61$_{\pm0.00}$	&41.68$_{\pm0.00}$	&42.39$_{\pm0.00}$	&39.99$_{\pm0.00}$	&[\textbf{65.16}$_{\pm2.76}$]\\
			\hline
			\emph{Movies}	&5.80$_{\pm1.23}$	&5.89$_{\pm1.20}$	&1.86$_{\pm0.00}$	&10.44$_{\pm0.65}$	&10.10$_{\pm0.53}$	&9.91$_{\pm0.67}$	&[\textbf{13.30}$_{\pm0.00}$]	&8.33$_{\pm0.00}$	&4.89$_{\pm0.00}$	&2.19$_{\pm0.00}$	&\textbf{12.36}$_{\pm1.52}$\\
			\hline
			\emph{BBCSport}	&67.10$_{\pm18.72}$	&89.82$_{\pm10.05}$	&37.39$_{\pm0.00}$	&[\textbf{91.71}$_{\pm1.06}$]	&90.49$_{\pm1.05}$	&90.51$_{\pm1.11}$	&69.88$_{\pm0.00}$	&87.39$_{\pm0.00}$	&6.01$_{\pm0.00}$	&10.28$_{\pm0.00}$	&\textbf{91.04}$_{\pm3.36}$\\
			\hline
			\emph{NH-550}	&74.52$_{\pm9.88}$	&53.80$_{\pm15.83}$	&\textbf{84.30}$_{\pm0.00}$	&80.35$_{\pm8.34}$	&74.71$_{\pm9.44}$	&74.73$_{\pm8.71}$	&2.66$_{\pm0.00}$	&73.18$_{\pm0.00}$	&83.95$_{\pm0.00}$	&62.25$_{\pm0.00}$	&[\textbf{92.55}$_{\pm8.92}$]\\
			\hline
			\emph{OutdoorScene}	&18.95$_{\pm12.56}$	&33.84$_{\pm3.90}$	&33.82$_{\pm0.00}$	&41.01$_{\pm0.88}$	&38.55$_{\pm0.63}$	&38.19$_{\pm0.96}$	&42.94$_{\pm0.00}$	&38.04$_{\pm0.00}$	&42.91$_{\pm0.00}$	&\textbf{43.72}$_{\pm0.00}$	&[\textbf{49.79}$_{\pm1.95}$]\\
			\hline
			\emph{Citeseer}	&0.06$_{\pm0.05}$	&0.09$_{\pm0.03}$	&0.10$_{\pm0.00}$	&25.34$_{\pm0.70}$	&\textbf{26.69}$_{\pm0.60}$	&24.27$_{\pm0.71}$	&24.61$_{\pm0.00}$	&1.15$_{\pm0.00}$	&14.86$_{\pm0.00}$	&11.74$_{\pm0.00}$	&[\textbf{31.18}$_{\pm2.26}$]\\
			\hline
			\emph{NH-4660}	&48.17$_{\pm23.18}$	&2.76$_{\pm2.15}$	&2.88$_{\pm0.00}$	&\textbf{83.05}$_{\pm5.29}$	&82.23$_{\pm0.46}$	&81.26$_{\pm4.20}$	&68.54$_{\pm0.00}$	&62.11$_{\pm0.00}$	&61.11$_{\pm0.00}$	&60.93$_{\pm0.00}$	&[\textbf{92.58}$_{\pm7.71}$]\\
			\hline
			\emph{ALOI}	&5.97$_{\pm1.23}$	&11.69$_{\pm2.23}$	&N/A	&N/A	&N/A	&N/A	&42.71$_{\pm0.00}$	&\textbf{44.95}$_{\pm0.00}$	&16.98$_{\pm0.00}$	&16.07$_{\pm0.00}$	&[\textbf{66.40}$_{\pm1.50}$]\\
			\hline
			\hline
			\emph{NUS-WIDE}	&N/A	&N/A	&N/A	&N/A	&{N/A}	&{N/A}	&\textbf{6.17}$_{\pm0.00}$	&2.40$_{\pm0.00}$	&5.31$_{\pm0.00}$	&[\textbf{6.39}$_{\pm0.00}$]	&5.07$_{\pm0.41}$\\
			\hline
			\emph{VGGFace2-50}	&N/A	&N/A	&N/A	&N/A	&N/A	&N/A	&\textbf{3.68}$_{\pm0.00}$	&3.35$_{\pm0.00}$	&2.70$_{\pm0.00}$	&3.21$_{\pm0.00}$	&[\textbf{4.32}$_{\pm0.15}$]\\
			\hline
			\emph{VGGFace2-100}	&N/A	&N/A	&N/A	&N/A	&N/A	&N/A	&2.00$_{\pm0.00}$	&\textbf{2.46}$_{\pm0.00}$	&1.98$_{\pm0.00}$	&2.06$_{\pm0.00}$	&[\textbf{2.47}$_{\pm0.09}$]\\
			\hline
			\emph{VGGFace2-200}	&{N/A}	&{N/A}	&{N/A}	&{N/A}	&{N/A}	&{N/A}	&{1.06$_{\pm0.00}$}	&{\textbf{1.28}$_{\pm0.00}$}	&{0.92$_{\pm0.00}$}	&{0.86$_{\pm0.00}$}	&{[\textbf{1.29}$_{\pm0.05}$]}\\
			\hline
			\emph{CIFAR-10}	&N/A	&N/A	&N/A	&N/A	&N/A	&N/A	&10.95$_{\pm0.00}$	&8.09$_{\pm0.00}$	&12.56$_{\pm0.00}$	&\textbf{12.63}$_{\pm0.00}$	&[\textbf{13.56}$_{\pm0.49}$]\\
			\hline
			\emph{CIFAR-100}	&N/A	&N/A	&N/A	&N/A	&N/A	&N/A	&\textbf{3.39}$_{\pm0.00}$	&2.14$_{\pm0.00}$	&3.22$_{\pm0.00}$	&3.04$_{\pm0.00}$	&[\textbf{3.55}$_{\pm0.14}$]\\
			\hline
			\emph{YTF-10}	&N/A	&N/A	&N/A	&N/A	&N/A	&N/A	&\textbf{69.92}$_{\pm0.00}$	&56.96$_{\pm0.00}$	&66.04$_{\pm0.00}$	&65.04$_{\pm0.00}$	&[\textbf{70.79}$_{\pm5.57}$]\\
			\hline
			\emph{YTF-20}	&N/A	&N/A	&N/A	&N/A	&N/A	&N/A	&52.56$_{\pm0.00}$	&56.35$_{\pm0.00}$	&[\textbf{66.02}$_{\pm0.00}$]	&61.50$_{\pm0.00}$	&\textbf{63.50}$_{\pm4.48}$\\
			\hline
			\emph{YTF-50}	&N/A	&N/A	&N/A	&N/A	&N/A	&N/A	&\textbf{60.32}$_{\pm0.00}$	&56.97$_{\pm0.00}$	&60.10$_{\pm0.00}$	&58.05$_{\pm0.00}$	&[\textbf{61.24}$_{\pm2.34}$]\\
			\hline
			\emph{YTF-100}	&N/A	&N/A	&N/A	&N/A	&N/A	&N/A	&\textbf{58.51}$_{\pm0.00}$	&52.22$_{\pm0.00}$	&48.48$_{\pm0.00}$	&24.21$_{\pm0.00}$	&[\textbf{60.00}$_{\pm2.27}$]\\
			\hline
			\emph{YTF-200}	&N/A	&N/A	&N/A	&N/A	&N/A	&N/A	&N/A	&\textbf{49.30}$_{\pm0.00}$	&35.82$_{\pm0.00}$	&19.43$_{\pm0.00}$	&[\textbf{56.83}$_{\pm0.95}$]\\
			\hline
			\emph{YTF-400}	&N/A	&N/A	&N/A	&N/A	&N/A	&N/A	&N/A	&N/A	&N/A	&N/A	&[\textbf{51.49}$_{\pm1.14}$]\\
			\hline
			\hline
			Average Rank	&6.55	&6.45	&6.27	&4.64	&5.73	&5.77	&4.55	&5.00	&4.41	&5.27	&1.27\\
			\hline
		\end{tabular} 
		\begin{tablenotes}
			\item[*] Note that N/A indicates the out-of-memory error.
		\end{tablenotes}
	\end{threeparttable}\vskip -0.1 in
\end{table*}

\subsection{Baseline Methods and Experimental Settings}

Our proposed FastMICE method is experimentally compared against ten MVC methods, which are listed below.

\begin{itemize}
	\item \textbf{MVSC} \cite{li15_mvsc}: multi-view spectral clustering.
	\item \textbf{AMGL} \cite{nei16_amgl}: auto-weighted multiple graph learning.
	\item \textbf{SwMC} \cite{nie17_ijcai}: self-weighted multi-view clustering.
	\item \textbf{MVEC} \cite{tao17_ijcai}: multi-view ensemble clustering.
	\item \textbf{M$^2$VEC$_{km}$} \cite{tao20_tnnls}: marginalized multi-view ensemble clustering with $k$-means
	\item \textbf{M$^2$VEC$_{spec}$} \cite{tao20_tnnls}: marginalized multi-view ensemble clustering with spectral clustering
	\item \textbf{BMVC} \cite{zhang19_bmvc}: binary multi-view clustering.
	\item \textbf{LMVSC} \cite{kang2020_lmvsc}: large-scale multi-view subspace clustering.
	\item \textbf{SMVSC} \cite{sun21_smvsc}: scalable multi-view subspace clustering.
	\item \textbf{FPMVS-CAG} \cite{wang22_tip}: fast parameter-free multi-view subspace clustering with consensus anchor guidance.
\end{itemize}

\begin{table*}[!t]\scriptsize
	\centering
	\caption{Average ACC(\%) scores over 20 runs by different multi-view clustering algorithms.
		On each dataset, the best two scores are highlighted in \textbf{bold}, while the best one in [\textbf{bold and brackets}].}\vskip -0.1 in
	\label{table:compare_acc}
	\begin{threeparttable}
		\renewcommand\arraystretch{1.02}
		\begin{tabular}{|m{1.56cm}<{\centering}|m{1cm}<{\centering}m{1cm}<{\centering}m{0.92cm}<{\centering}m{1.02cm}<{\centering}m{0.92cm}<{\centering}m{1cm}<{\centering}m{1.02cm}<{\centering}m{0.92cm}<{\centering}m{0.9cm}<{\centering}m{1.468cm}<{\centering}|m{1.36cm}<{\centering}|}
			\hline
			Dataset    &MVSC     &AMGL  &SwMC &MVEC     &M$^2$VEC$_{km}$     &M$^2$VEC$_{spec}$     &BMVC   &LMVSC  &SMVSC   &FPMVS-CAG  &FastMICE\\
			\hline
			\hline
			\emph{MSRCv1}	&56.98$_{\pm5.33}$	&64.90$_{\pm8.10}$	&70.48$_{\pm0.00}$	&\textbf{70.81}$_{\pm1.75}$	&67.81$_{\pm2.22}$	&67.90$_{\pm2.21}$	&64.76$_{\pm0.00}$	&36.67$_{\pm0.00}$	&67.14$_{\pm0.00}$	&60.00$_{\pm0.00}$	&[\textbf{79.64}$_{\pm5.18}$]\\
			\hline
			\emph{Yale}	&59.03$_{\pm4.26}$	&60.12$_{\pm3.74}$	&61.21$_{\pm0.00}$	&\textbf{62.21}$_{\pm1.79}$	&45.18$_{\pm8.23}$	&44.03$_{\pm3.86}$	&22.42$_{\pm0.00}$	&55.15$_{\pm0.00}$	&56.97$_{\pm0.00}$	&44.24$_{\pm0.00}$	&[\textbf{64.21}$_{\pm3.69}$]\\
			\hline
			\emph{ORL}	&72.36$_{\pm3.81}$	&\textbf{72.61}$_{\pm2.25}$	&70.75$_{\pm0.00}$	&68.60$_{\pm1.43}$	&57.55$_{\pm2.38}$	&56.45$_{\pm4.81}$	&16.75$_{\pm0.00}$	&57.00$_{\pm0.00}$	&56.25$_{\pm0.00}$	&56.00$_{\pm0.00}$	&[\textbf{73.05}$_{\pm2.63}$]\\
			\hline
			\emph{Movies}	&25.19$_{\pm1.31}$	&26.17$_{\pm1.39}$	&20.91$_{\pm0.00}$	&27.07$_{\pm0.83}$	&27.49$_{\pm0.75}$	&27.78$_{\pm0.98}$	&[\textbf{29.34}$_{\pm0.00}$]	&26.74$_{\pm0.00}$	&21.39$_{\pm0.00}$	&17.02$_{\pm0.00}$	&\textbf{28.77}$_{\pm2.06}$\\
			\hline
			\emph{BBCSport}	&77.67$_{\pm12.79}$	&94.43$_{\pm7.15}$	&62.87$_{\pm0.00}$	&[\textbf{96.96}$_{\pm0.41}$]	&\textbf{96.50}$_{\pm0.41}$	&96.50$_{\pm0.43}$	&75.74$_{\pm0.00}$	&94.67$_{\pm0.00}$	&32.35$_{\pm0.00}$	&40.63$_{\pm0.00}$	&95.15$_{\pm5.54}$\\
			\hline
			\emph{NH-550}	&81.20$_{\pm5.86}$	&67.94$_{\pm11.39}$	&87.82$_{\pm0.00}$	&89.95$_{\pm4.43}$	&84.43$_{\pm9.12}$	&84.55$_{\pm8.81}$	&27.27$_{\pm0.00}$	&80.73$_{\pm0.00}$	&\textbf{91.45}$_{\pm0.00}$	&70.55$_{\pm0.00}$	&[\textbf{93.98}$_{\pm9.26}$]\\
			\hline
			\emph{OutdoorScene}	&31.82$_{\pm10.36}$	&49.91$_{\pm5.12}$	&45.46$_{\pm0.00}$	&62.55$_{\pm0.92}$	&59.58$_{\pm0.50}$	&59.69$_{\pm1.17}$	&\textbf{63.62}$_{\pm0.00}$	&59.86$_{\pm0.00}$	&63.32$_{\pm0.00}$	&63.58$_{\pm0.00}$	&[\textbf{69.85}$_{\pm3.57}$]\\
			\hline
			\emph{Citeseer}	&21.24$_{\pm0.17}$	&21.48$_{\pm0.06}$	&21.62$_{\pm0.00}$	&55.57$_{\pm0.48}$	&\textbf{56.20}$_{\pm0.42}$	&54.84$_{\pm0.47}$	&54.41$_{\pm0.00}$	&29.35$_{\pm0.00}$	&41.94$_{\pm0.00}$	&37.53$_{\pm0.00}$	&[\textbf{59.32}$_{\pm1.60}$]\\
			\hline
			\emph{NH-4660}	&62.93$_{\pm14.53}$	&36.29$_{\pm1.84}$	&33.88$_{\pm0.00}$	&\textbf{88.09}$_{\pm4.27}$	&86.00$_{\pm1.43}$	&85.70$_{\pm2.74}$	&85.94$_{\pm0.00}$	&73.76$_{\pm0.00}$	&71.61$_{\pm0.00}$	&71.31$_{\pm0.00}$	&[\textbf{94.38}$_{\pm7.13}$]\\
			\hline
			\emph{ALOI}	&12.36$_{\pm0.83}$	&60.00$_{\pm2.06}$	&N/A	&N/A	&N/A	&N/A	&\textbf{60.39}$_{\pm0.00}$	&56.61$_{\pm0.00}$	&34.30$_{\pm0.00}$	&31.53$_{\pm0.00}$	&[\textbf{75.62}$_{\pm1.58}$]\\
			\hline
			\hline
			{\emph{NUS-WIDE}}	&{N/A}	&{N/A}	&{N/A}	&{N/A}	&{N/A}	&{N/A}	&{14.32$_{\pm0.00}$}	&{12.06$_{\pm0.00}$}	&{\textbf{17.75}$_{\pm0.00}$}	&{[\textbf{19.41}$_{\pm0.00}$]}	&{14.44$_{\pm0.58}$}\\
			\hline
			\emph{VGGFace2-50}	&N/A	&N/A	&N/A	&N/A	&N/A	&N/A	&\textbf{11.86}$_{\pm0.00}$	&11.55$_{\pm0.00}$	&9.83$_{\pm0.00}$	&10.74$_{\pm0.00}$	&[\textbf{12.65}$_{\pm0.26}$]\\
			\hline
			\emph{VGGFace2-100}	&N/A	&N/A	&N/A	&N/A	&N/A	&N/A	&7.37$_{\pm0.00}$	&[\textbf{8.46}$_{\pm0.00}$]	&6.56$_{\pm0.00}$	&7.21$_{\pm0.00}$	&\textbf{8.09}$_{\pm0.26}$\\
			\hline
			{\emph{VGGFace2-200}}	&{N/A}	&{N/A}	&{N/A}	&{N/A}	&{N/A}	&{N/A}	&{4.68$_{\pm0.00}$}	&{[\textbf{5.14}$_{\pm0.00}$]}	&{4.16$_{\pm0.00}$}	&{4.22$_{\pm0.00}$}	&{\textbf{5.08}$_{\pm0.16}$}\\
			\hline
			\emph{CIFAR-10}	&N/A	&N/A	&N/A	&N/A	&N/A	&N/A	&28.03$_{\pm0.00}$	&25.81$_{\pm0.00}$	&29.20$_{\pm0.00}$	&\textbf{29.55}$_{\pm0.00}$	&[\textbf{33.28}$_{\pm1.24}$]\\
			\hline
			\emph{CIFAR-100}	&N/A	&N/A	&N/A	&N/A	&N/A	&N/A	&8.95$_{\pm0.00}$	&7.96$_{\pm0.00}$	&8.14$_{\pm0.00}$	&\textbf{9.06}$_{\pm0.00}$	&[\textbf{10.31}$_{\pm0.26}$]\\
			\hline
			\emph{YTF-10}	&N/A	&N/A	&N/A	&N/A	&N/A	&N/A	&[\textbf{78.62}$_{\pm0.00}$]	&69.32$_{\pm0.00}$	&73.81$_{\pm0.00}$	&71.49$_{\pm0.00}$	&\textbf{76.74}$_{\pm4.39}$\\
			\hline
			\emph{YTF-20}	&N/A	&N/A	&N/A	&N/A	&N/A	&N/A	&60.79$_{\pm0.00}$	&63.88$_{\pm0.00}$	&[\textbf{74.79}$_{\pm0.00}$]	&68.87$_{\pm0.00}$	&\textbf{71.47}$_{\pm3.78}$\\
			\hline
			\emph{YTF-50}	&N/A	&N/A	&N/A	&N/A	&N/A	&N/A	&[\textbf{70.06}$_{\pm0.00}$]	&\textbf{69.49}$_{\pm0.00}$	&67.44$_{\pm0.00}$	&63.73$_{\pm0.00}$	&69.21$_{\pm2.30}$\\
			\hline
			\emph{YTF-100}	&N/A	&N/A	&N/A	&N/A	&N/A	&N/A	&\textbf{65.67}$_{\pm0.00}$	&60.71$_{\pm0.00}$	&59.47$_{\pm0.00}$	&53.49$_{\pm0.00}$	&[\textbf{66.83}$_{\pm1.67}$]\\
			\hline
			\emph{YTF-200}	&N/A	&N/A	&N/A	&N/A	&N/A	&N/A	&N/A	&\textbf{55.00}$_{\pm0.00}$	&52.88$_{\pm0.00}$	&50.45$_{\pm0.00}$	&[\textbf{62.79}$_{\pm0.86}$]\\
			\hline
			\emph{YTF-400}	&N/A	&N/A	&N/A	&N/A	&N/A	&N/A	&N/A	&N/A	&N/A	&N/A	&[\textbf{56.40}$_{\pm0.82}$]\\
			\hline
			\hline
			Average Rank	&6.59	&6.14	&6.27	&4.68	&5.41	&5.64	&4.45	&4.82	&4.82	&5.55	&1.55\\
			\hline
		\end{tabular}
		\begin{tablenotes}
			\item[*] Note that N/A indicates the out-of-memory error.
		\end{tablenotes}
	\end{threeparttable}\vskip -0.05 in
\end{table*}

\begin{table*}[!t]\scriptsize
	\centering
	\caption{Average PUR(\%) scores over 20 runs by different multi-view clustering algorithms.
		On each dataset, the best two scores are highlighted in \textbf{bold}, while the best one in [\textbf{bold and brackets}].}\vskip -0.1 in
	\label{table:compare_pur}
	\begin{threeparttable}
		\renewcommand\arraystretch{1.02}
		\begin{tabular}{|m{1.56cm}<{\centering}|m{1cm}<{\centering}m{1cm}<{\centering}m{0.92cm}<{\centering}m{1.02cm}<{\centering}m{0.92cm}<{\centering}m{1cm}<{\centering}m{1.02cm}<{\centering}m{0.92cm}<{\centering}m{0.9cm}<{\centering}m{1.468cm}<{\centering}|m{1.36cm}<{\centering}|}
			\hline
			Dataset    &MVSC     &AMGL  &SwMC &MVEC     &M$^2$VEC$_{km}$     &M$^2$VEC$_{spec}$     &BMVC   &LMVSC  &SMVSC   &FPMVS-CAG  &FastMICE\\
			\hline
			\hline
			\emph{MSRCv1}	&59.31$_{\pm5.06}$	&68.10$_{\pm7.17}$	&73.33$_{\pm0.00}$	&\textbf{74.29}$_{\pm1.96}$	&71.79$_{\pm2.05}$	&71.69$_{\pm2.31}$	&66.67$_{\pm0.00}$	&40.95$_{\pm0.00}$	&69.05$_{\pm0.00}$	&62.86$_{\pm0.00}$	&[\textbf{81.98}$_{\pm3.13}$]\\
			\hline
			\emph{Yale}	&60.33$_{\pm3.59}$	&61.27$_{\pm2.86}$	&61.82$_{\pm0.00}$	&\textbf{63.70}$_{\pm1.58}$	&46.94$_{\pm7.71}$	&46.39$_{\pm3.32}$	&24.24$_{\pm0.00}$	&56.97$_{\pm0.00}$	&56.97$_{\pm0.00}$	&46.67$_{\pm0.00}$	&[\textbf{65.27}$_{\pm3.10}$]\\
			\hline
			\emph{ORL}	&77.05$_{\pm2.47}$	&[\textbf{78.08}$_{\pm1.87}$]	&76.75$_{\pm0.00}$	&75.41$_{\pm1.08}$	&67.85$_{\pm1.60}$	&62.78$_{\pm3.65}$	&18.00$_{\pm0.00}$	&63.50$_{\pm0.00}$	&60.25$_{\pm0.00}$	&60.00$_{\pm0.00}$	&\textbf{77.33}$_{\pm2.03}$\\
			\hline
			\emph{Movies}	&28.11$_{\pm1.37}$	&28.50$_{\pm1.19}$	&23.34$_{\pm0.00}$	&30.26$_{\pm0.77}$	&30.86$_{\pm0.57}$	&31.15$_{\pm1.25}$	&[\textbf{32.25}$_{\pm0.00}$]	&28.36$_{\pm0.00}$	&23.50$_{\pm0.00}$	&18.64$_{\pm0.00}$	&\textbf{32.00}$_{\pm1.87}$\\
			\hline
			\emph{BBCSport}	&81.53$_{\pm9.87}$	&94.99$_{\pm5.92}$	&64.15$_{\pm0.00}$	&[\textbf{96.96}$_{\pm0.41}$]	&\textbf{96.50}$_{\pm0.41}$	&96.50$_{\pm0.43}$	&85.29$_{\pm0.00}$	&94.67$_{\pm0.00}$	&42.83$_{\pm0.00}$	&48.16$_{\pm0.00}$	&95.89$_{\pm3.27}$\\
			\hline
			\emph{NH-550}	&83.55$_{\pm6.19}$	&69.02$_{\pm10.93}$	&87.82$_{\pm0.00}$	&89.95$_{\pm4.43}$	&86.89$_{\pm4.76}$	&87.01$_{\pm4.54}$	&35.64$_{\pm0.00}$	&84.00$_{\pm0.00}$	&\textbf{91.45}$_{\pm0.00}$	&82.73$_{\pm0.00}$	&[\textbf{96.02}$_{\pm4.77}$]\\
			\hline
			\emph{OutdoorScene}	&32.03$_{\pm10.40}$	&50.47$_{\pm5.32}$	&45.57$_{\pm0.00}$	&62.80$_{\pm0.32}$	&62.00$_{\pm0.47}$	&61.77$_{\pm0.47}$	&63.62$_{\pm0.00}$	&59.86$_{\pm0.00}$	&66.00$_{\pm0.00}$	&\textbf{67.08}$_{\pm0.00}$	&[\textbf{70.50}$_{\pm2.73}$]\\
			\hline
			\emph{Citeseer}	&21.39$_{\pm0.11}$	&21.57$_{\pm0.05}$	&22.04$_{\pm0.00}$	&57.22$_{\pm0.46}$	&\textbf{57.90}$_{\pm0.41}$	&56.44$_{\pm0.47}$	&56.40$_{\pm0.00}$	&31.76$_{\pm0.00}$	&47.01$_{\pm0.00}$	&42.81$_{\pm0.00}$	&[\textbf{61.48}$_{\pm1.78}$]\\
			\hline
			\emph{NH-4660}	&65.24$_{\pm13.91}$	&37.07$_{\pm2.10}$	&37.10$_{\pm0.00}$	&\textbf{88.23}$_{\pm3.68}$	&87.24$_{\pm0.01}$	&86.78$_{\pm1.96}$	&85.94$_{\pm0.00}$	&81.20$_{\pm0.00}$	&78.41$_{\pm0.00}$	&78.50$_{\pm0.00}$	&[\textbf{95.41}$_{\pm4.90}$]\\
			\hline
			\emph{ALOI}	&13.57$_{\pm0.90}$	&\textbf{63.44}$_{\pm1.55}$	&N/A	&N/A	&N/A	&N/A	&63.27$_{\pm0.00}$	&59.76$_{\pm0.00}$	&35.28$_{\pm0.00}$	&32.13$_{\pm0.00}$	&[\textbf{76.98}$_{\pm1.35}$]\\
			\hline
			\hline
			{\emph{NUS-WIDE}}	&{N/A}	&{N/A}	&{N/A}	&{N/A}	&{N/A}	&{N/A}	&{[\textbf{28.57}$_{\pm0.00}$]}	&{20.09$_{\pm0.00}$}	&{22.81$_{\pm0.00}$}	&{23.61$_{\pm0.00}$}	&{\textbf{24.75}$_{\pm0.57}$}\\
			\hline
			\emph{VGGFace2-50}	&N/A	&N/A	&N/A	&N/A	&N/A	&N/A	&\textbf{12.87}$_{\pm0.00}$	&12.29$_{\pm0.00}$	&10.28$_{\pm0.00}$	&11.15$_{\pm0.00}$	&[\textbf{13.67}$_{\pm0.27}$]\\
			\hline
			\emph{VGGFace2-100}	&N/A	&N/A	&N/A	&N/A	&N/A	&N/A	&8.19$_{\pm0.00}$	&[\textbf{9.28}$_{\pm0.00}$]	&6.74$_{\pm0.00}$	&7.27$_{\pm0.00}$	&\textbf{9.04}$_{\pm0.24}$\\
			\hline
			{\emph{VGGFace2-200}}	&{N/A}	&{N/A}	&{N/A}	&{N/A}	&{N/A}	&{N/A}	&{5.40$_{\pm0.00}$}	&{\textbf{5.71}$_{\pm0.00}$}	&{4.37$_{\pm0.00}$}	&{4.26$_{\pm0.00}$}	&{[\textbf{5.79}$_{\pm0.18}$]}\\
			\hline
			\emph{CIFAR-10}	&N/A	&N/A	&N/A	&N/A	&N/A	&N/A	&30.55$_{\pm0.00}$	&27.48$_{\pm0.00}$	&32.40$_{\pm0.00}$	&\textbf{32.85}$_{\pm0.00}$	&[\textbf{35.28}$_{\pm0.73}$]\\
			\hline
			\emph{CIFAR-100}	&N/A	&N/A	&N/A	&N/A	&N/A	&N/A	&\textbf{10.05}$_{\pm0.00}$	&9.37$_{\pm0.00}$	&8.64$_{\pm0.00}$	&9.45$_{\pm0.00}$	&[\textbf{11.71}$_{\pm0.26}$]\\
			\hline
			\emph{YTF-10}	&N/A	&N/A	&N/A	&N/A	&N/A	&N/A	&\textbf{81.40}$_{\pm0.00}$	&73.09$_{\pm0.00}$	&78.89$_{\pm0.00}$	&76.22$_{\pm0.00}$	&[\textbf{81.95}$_{\pm3.03}$]\\
			\hline
			\emph{YTF-20}	&N/A	&N/A	&N/A	&N/A	&N/A	&N/A	&69.33$_{\pm0.00}$	&72.33$_{\pm0.00}$	&[\textbf{77.71}$_{\pm0.00}$]	&75.06$_{\pm0.00}$	&\textbf{77.08}$_{\pm2.61}$\\
			\hline
			\emph{YTF-50}	&N/A	&N/A	&N/A	&N/A	&N/A	&N/A	&[\textbf{75.37}$_{\pm0.00}$]	&73.61$_{\pm0.00}$	&70.84$_{\pm0.00}$	&66.90$_{\pm0.00}$	&\textbf{75.18}$_{\pm1.68}$\\
			\hline
			\emph{YTF-100}	&N/A	&N/A	&N/A	&N/A	&N/A	&N/A	&[\textbf{74.76}$_{\pm0.00}$]	&69.37$_{\pm0.00}$	&62.21$_{\pm0.00}$	&54.76$_{\pm0.00}$	&\textbf{73.59}$_{\pm1.43}$\\
			\hline
			\emph{YTF-200}	&N/A	&N/A	&N/A	&N/A	&N/A	&N/A	&N/A	&\textbf{65.14}$_{\pm0.00}$	&57.19$_{\pm0.00}$	&51.55$_{\pm0.00}$	&[\textbf{70.94}$_{\pm0.55}$]\\
			\hline
			\emph{YTF-400}	&N/A	&N/A	&N/A	&N/A	&N/A	&N/A	&N/A	&N/A	&N/A	&N/A	&[\textbf{68.11}$_{\pm0.39}$]\\
			\hline
			\hline
			Average Rank	&6.68	&6.00	&6.23	&4.68	&5.27	&5.64	&4.41	&4.95	&4.95	&5.64	&1.45\\
			\hline
		\end{tabular}
		\begin{tablenotes}
			\item[*] Note that N/A indicates the out-of-memory error.
		\end{tablenotes}
	\end{threeparttable}\vskip -0.1 in
\end{table*}

Among these baseline methods,  BMVC, LMVSC, SMVSC, and FPMVS-CAG are four large-scale MVC methods, MVEC, M$^2$VEC$_{km}$, and M$^2$VEC$_{spec}$ are three EC based MVC methods, and AMGL, SwMC, and FPMVS-CAG are three tuning-free MVC methods. For the baseline methods as well as the proposed method, if the distance metric can be customized, then the cosine distance will be adopted for the document datasets, such as \emph{Movies}, \emph{BBCSport}, and \emph{Citeseer}. Otherwise, their suggested distance (mostly Euclidean distance) will be adopted.

For each of the baseline methods, if the dataset-specific tuning is needed, then each of its hyperparameters will be tuned in the range of $\{10^{-5},10^{-4},\cdots,10^{5}\}$, unless the tuning range is specifically suggested by the corresponding paper. To avoid the expensive or even unaffordable computational costs of hyperparameter-tuning on the entire large-scale datasets, for all the baseline methods except M$^2$VEC$_{km}$  and M$^2$VEC$_{spec}$,  if $N>$ 10,000,  the tuning will be conducted on a random subset of 10,000 samples. For M$^2$VEC$_{km}$  and M$^2$VEC$_{spec}$,  whose computational costs rapidly increase with larger data sizes, if $N>$ 1,000,  the tuning will be conducted on a random subset of 1,000 samples. 

Note that the proposed FastMICE method does not require dataset-specific tuning. Though there exist several parameters in FastMICE, yet these parameters can safely be set to some common values (or randomized in some common ranges) across various datasets. Specifically, the feature sampling ratio $\tau^{(m)}_v$ for each view member is randomized in $[0.2,0.8]$. The number of clusters $k^{(m)}$ in each base clustering is randomized in $[k,2k]$, where $k$ is the desired number of clusters in the final clustering. Besides these randomizations, the number of base clusterings $M=20$, the number of anchors $p=\min\{\text{1,000},N\}$, and the number of nearest neighbors $K=5$ are used in the experiments on all benchmark datasets. 

To evaluate the clustering performances of different MVC methods, we adopt four wide-used evaluation metrics in our experiments, namely, normalized mutual information (NMI) \cite{strehl02}, adjusted Rand index (ARI)\cite{huang21_tcyb}, accuracy (ACC) \cite{ch20_tcyb}, and purity (PUR) \cite{ch20_tcyb}. Larger values of these metrics indicate better clustering results.

\begin{figure}[!t] \footnotesize 
	\begin{center}
		{\subfigure[\scriptsize \emph{MSRCv1}]
			{\includegraphics[width=0.24\columnwidth]{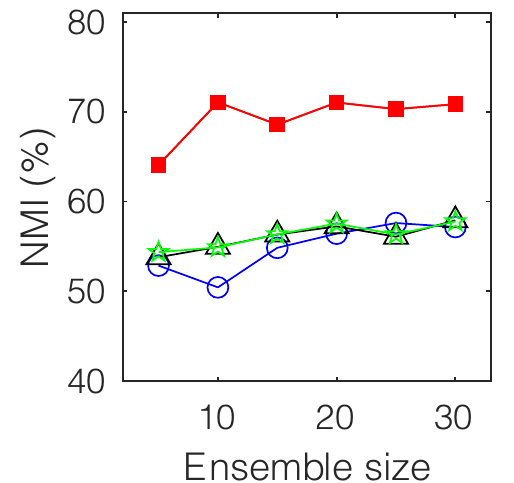}}}
		{\subfigure[\scriptsize \emph{Yale}]
			{\includegraphics[width=0.24\columnwidth]{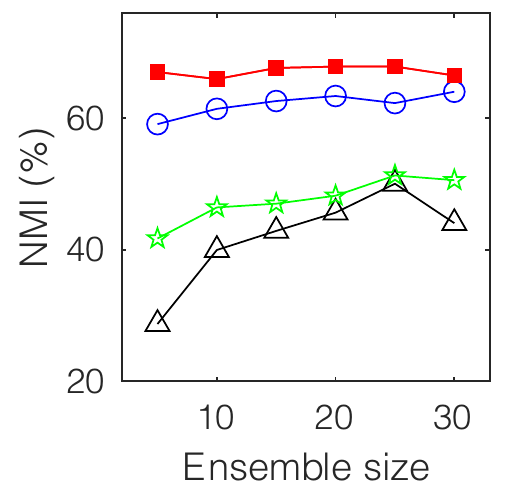}}}
		{\subfigure[\scriptsize \emph{ORL}]
			{\includegraphics[width=0.24\columnwidth]{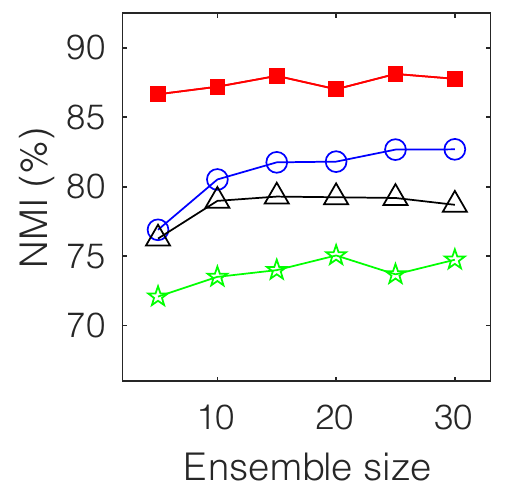}}}
		{\subfigure[\scriptsize \emph{Movies}]
			{\includegraphics[width=0.24\columnwidth]{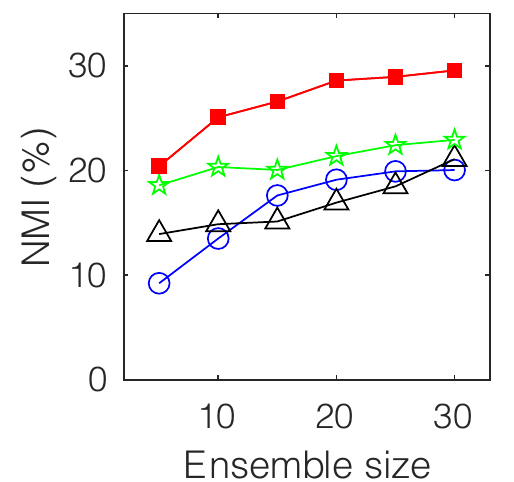}}}\\ 
		{\subfigure[\scriptsize \emph{BBCSport}]
			{\includegraphics[width=0.24\columnwidth]{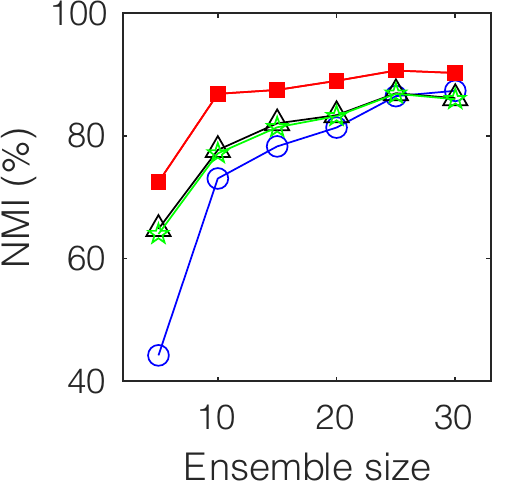}}}
		{\subfigure[\scriptsize \emph{NH-550}]
			{\includegraphics[width=0.24\columnwidth]{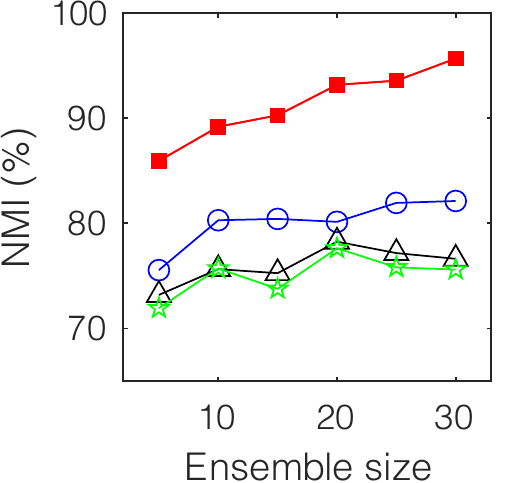}}}
		{\subfigure[\scriptsize \emph{Out-Scene}]
			{\includegraphics[width=0.24\columnwidth]{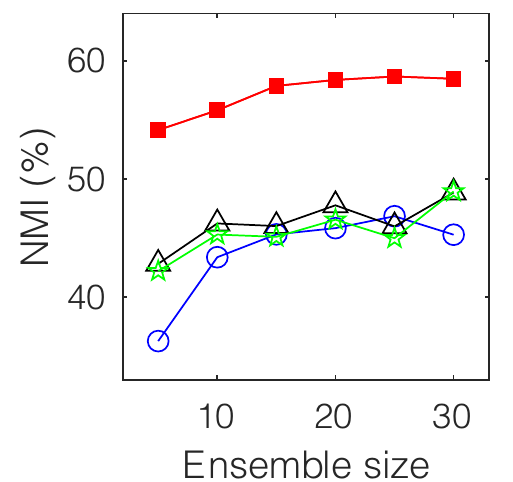}}}
		{\subfigure[\scriptsize \emph{Citeseer}]
			{\includegraphics[width=0.24\columnwidth]{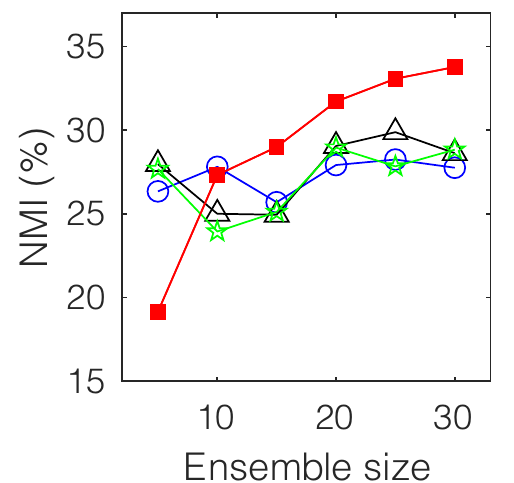}}}\\ 
		{\subfigure[\scriptsize \emph{NH-4660}]
			{\includegraphics[width=0.24\columnwidth]{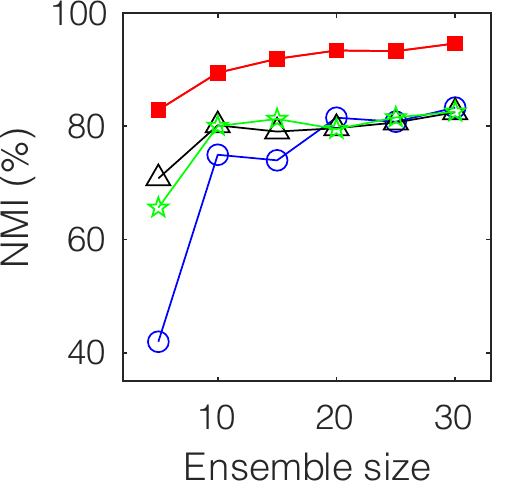}}}
		{\subfigure[\scriptsize \emph{ALOI}]
			{\includegraphics[width=0.24\columnwidth]{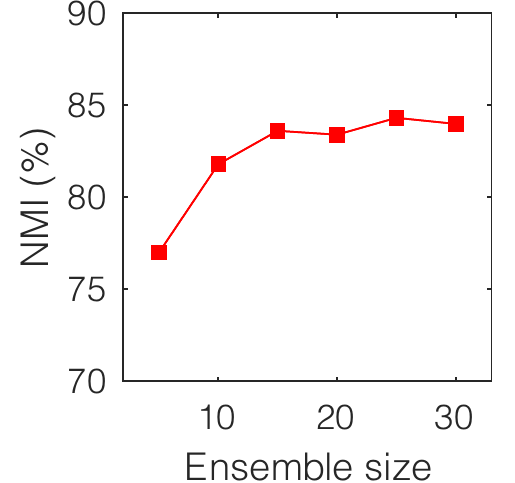}}}
		{\subfigure[\scriptsize {\emph{NUS-WIDE}}]
			{\includegraphics[width=0.24\columnwidth]{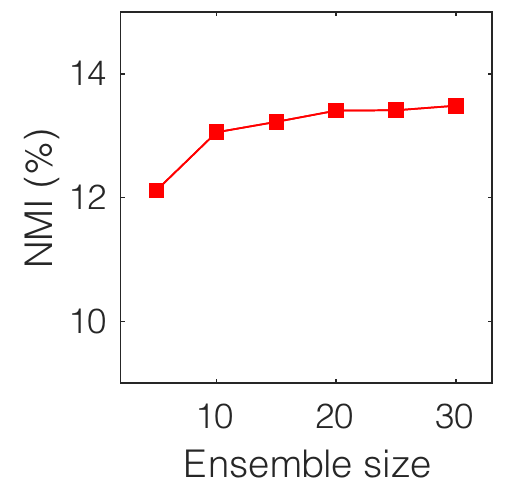}}}
		{\subfigure[\scriptsize \emph{VGGFace2-50}]
			{\includegraphics[width=0.24\columnwidth]{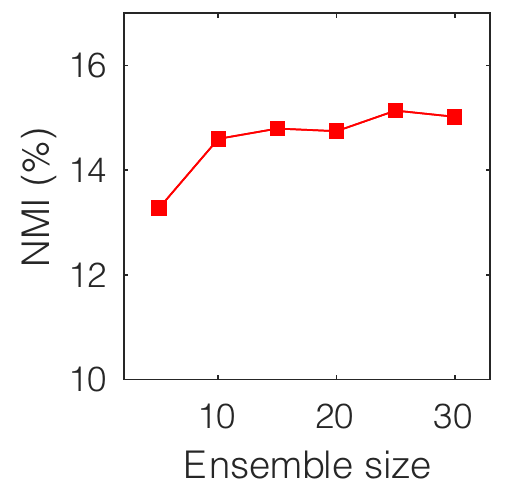}}}\\
		{\subfigure[\scriptsize \emph{VGGFace2-100}]
			{\includegraphics[width=0.24\columnwidth]{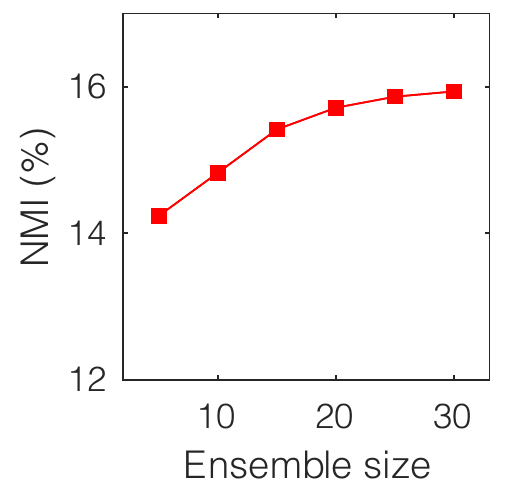}}}
		{\subfigure[\scriptsize{\emph{VGGFace2-200}}]
			{\includegraphics[width=0.24\columnwidth]{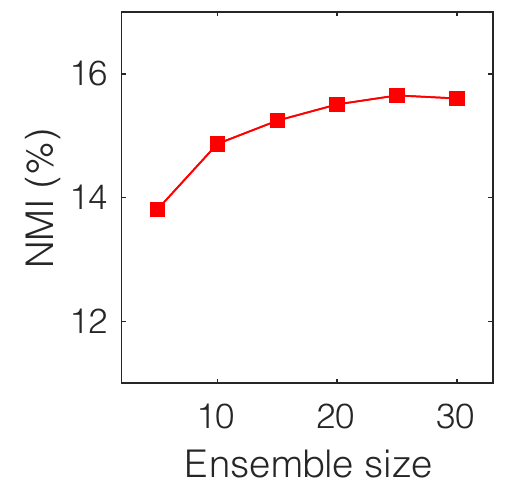}}}
		{\subfigure[\scriptsize \emph{CIFAR-10}]
			{\includegraphics[width=0.24\columnwidth]{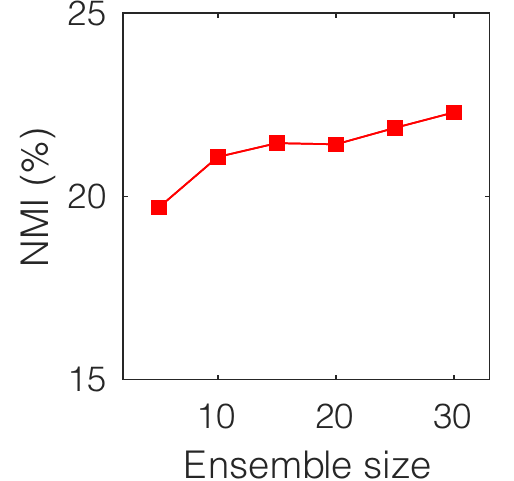}}}
		{\subfigure[\scriptsize \emph{CIFAR-100}]
			{\includegraphics[width=0.24\columnwidth]{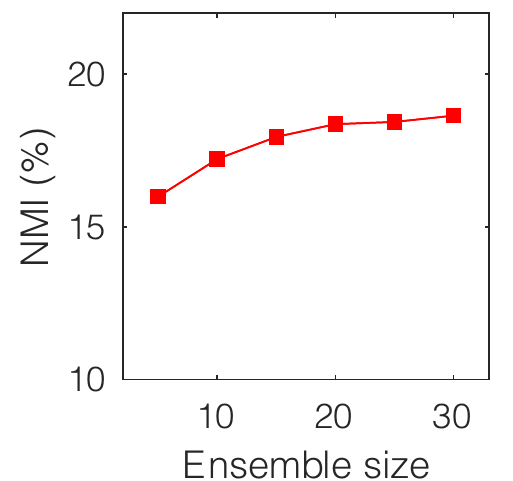}}}\\ 
		{\subfigure[\scriptsize \emph{YTF-10}]
			{\includegraphics[width=0.24\columnwidth]{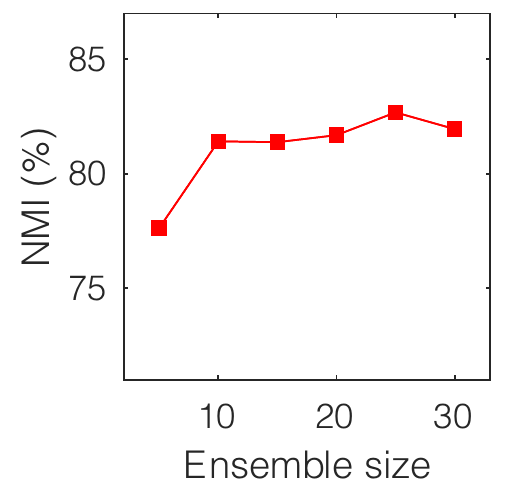}}}
		{\subfigure[\scriptsize \emph{YTF-20}]
			{\includegraphics[width=0.24\columnwidth]{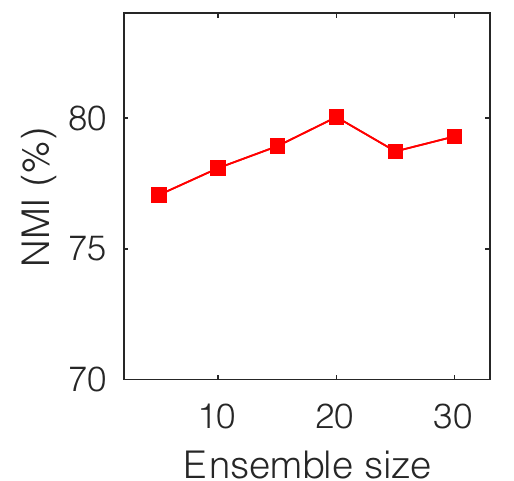}}}
		{\subfigure[\scriptsize \emph{YTF-50}]
			{\includegraphics[width=0.24\columnwidth]{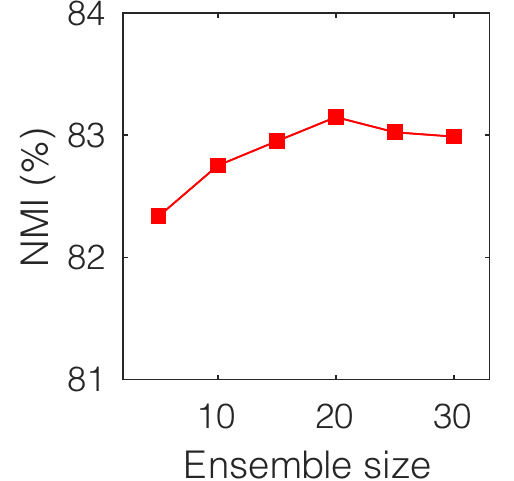}}}
		{\subfigure[\scriptsize \emph{YTF-100}]
			{\includegraphics[width=0.24\columnwidth]{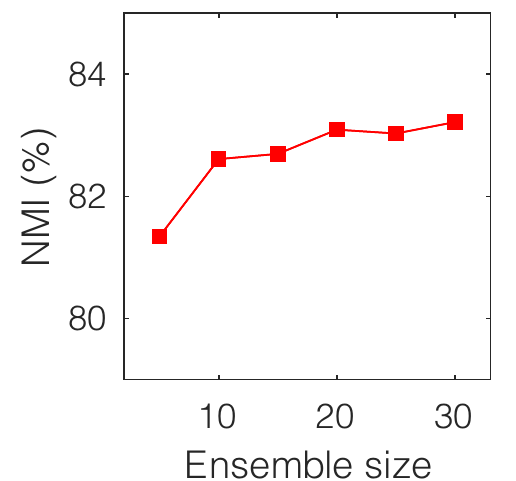}}}\\ 
		{\subfigure[\scriptsize \emph{YTF-200}]
			{\includegraphics[width=0.24\columnwidth]{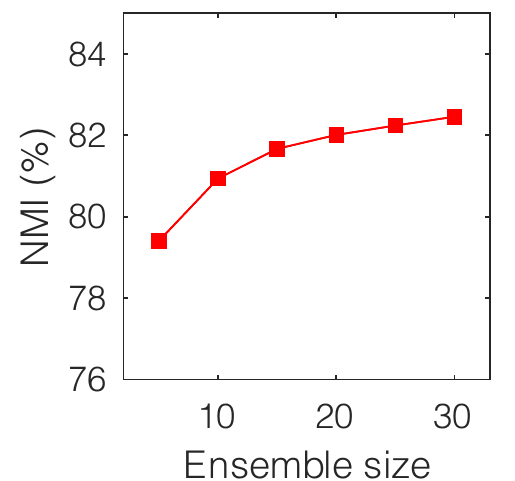}}}
		{\subfigure[\scriptsize \emph{YTF-400}]
			{\includegraphics[width=0.24\columnwidth]{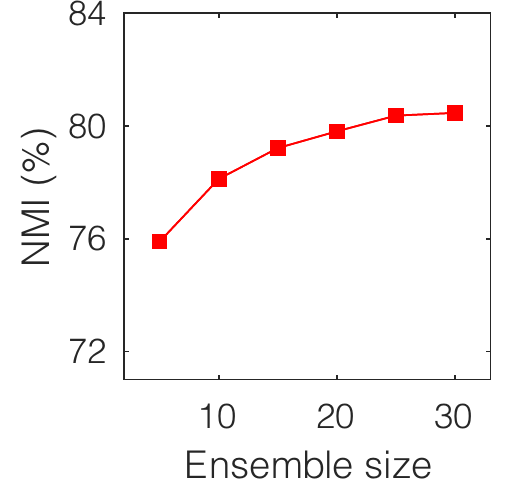}}}
		{\subfigure
			{\includegraphics[width=0.75\columnwidth]{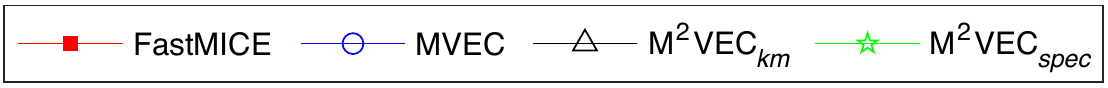}}}
		\caption{The average NMI (\%) scores over 20 runs by different EC based MVC algorithms with varying ensemble sizes $M$. Note that if an algorithm is not computationally feasible on a dataset, then its curve will not appear in the corresponding sub-figure.}
		\label{fig:comp_nmi_Msize}
	\end{center}
\end{figure}

\begin{figure}[!th] \scriptsize 
	\begin{center}
		{\subfigure[\scriptsize \emph{MSRCv1}]
			{\includegraphics[width=0.238\columnwidth]{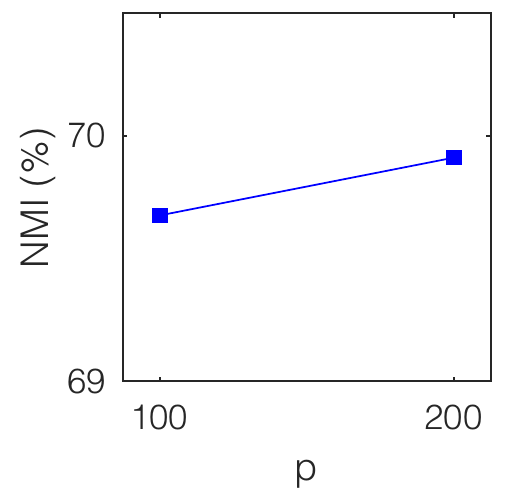}}}
		{\subfigure[\scriptsize \emph{Yale}]
			{\includegraphics[width=0.238\columnwidth]{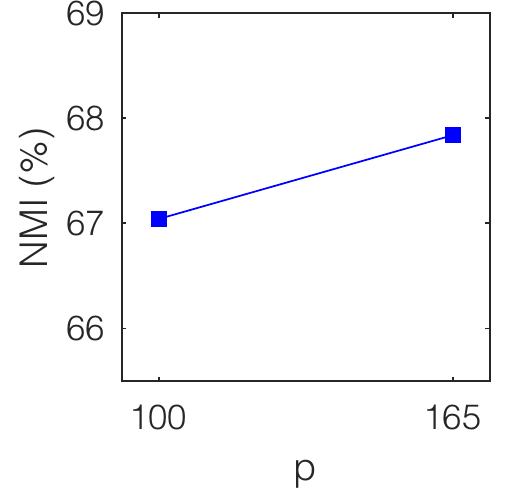}}}
		{\subfigure[\scriptsize \emph{ORL}]
			{\includegraphics[width=0.238\columnwidth]{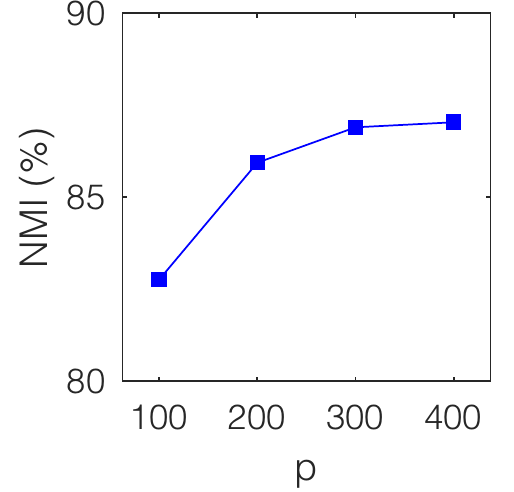}}}
		{\subfigure[\scriptsize \emph{Movies}]
			{\includegraphics[width=0.238\columnwidth]{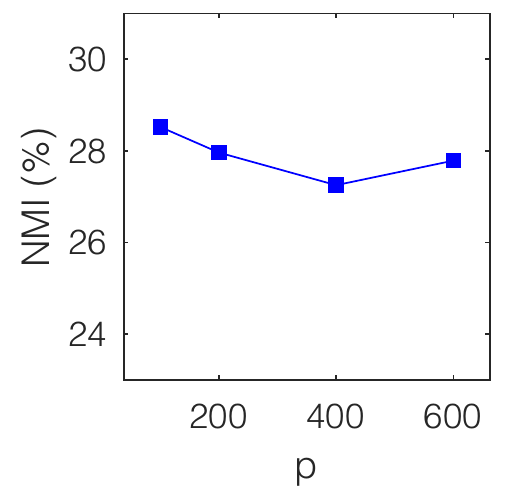}}}\\ 
		{\subfigure[\scriptsize \emph{BBCSport}]
			{\includegraphics[width=0.238\columnwidth]{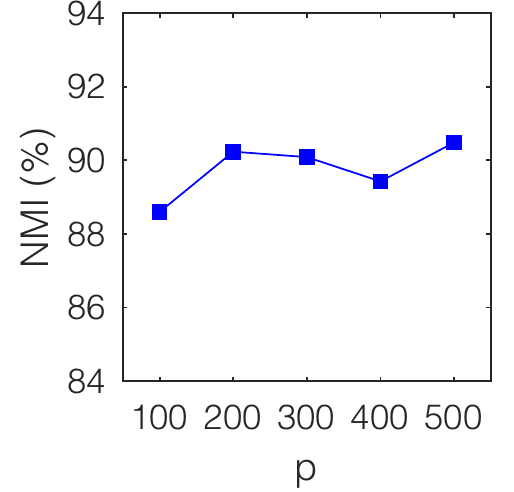}}}
		{\subfigure[\scriptsize \emph{NH-550}]
			{\includegraphics[width=0.238\columnwidth]{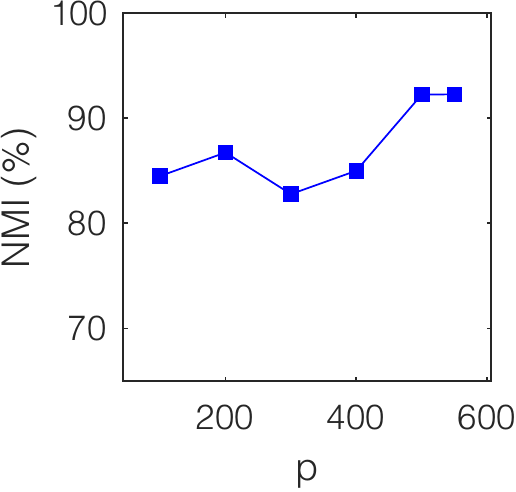}}}
		{\subfigure[\scriptsize \emph{Out-Scene}]
			{\includegraphics[width=0.238\columnwidth]{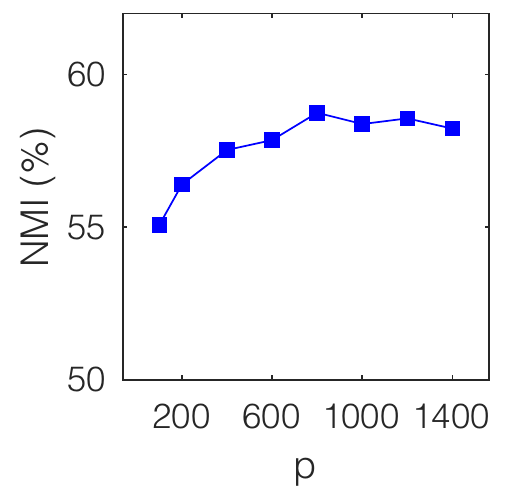}}}
		{\subfigure[\scriptsize \emph{Citeseer}]
			{\includegraphics[width=0.238\columnwidth]{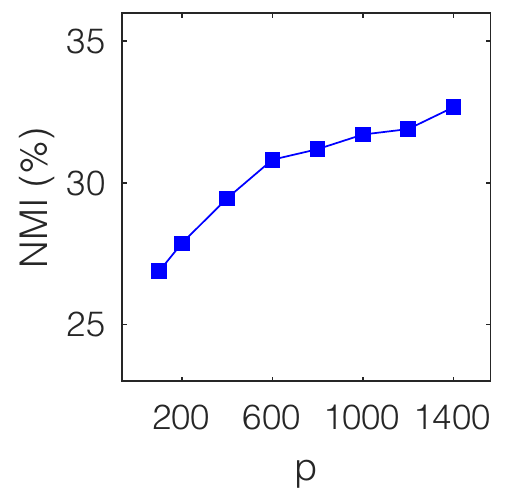}}}\\ 
		{\subfigure[\scriptsize \emph{NH-4660}]
			{\includegraphics[width=0.238\columnwidth]{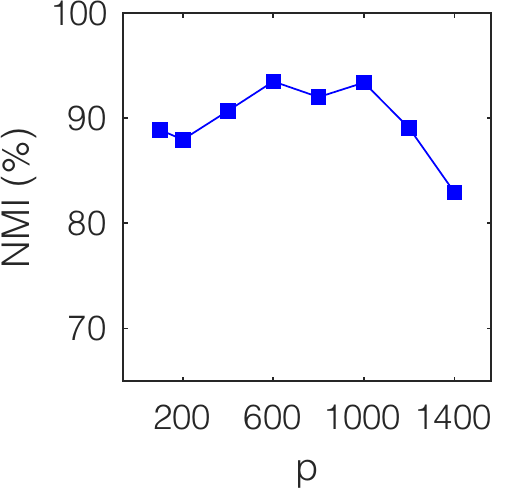}}}
		{\subfigure[\scriptsize \emph{ALOI}]
			{\includegraphics[width=0.238\columnwidth]{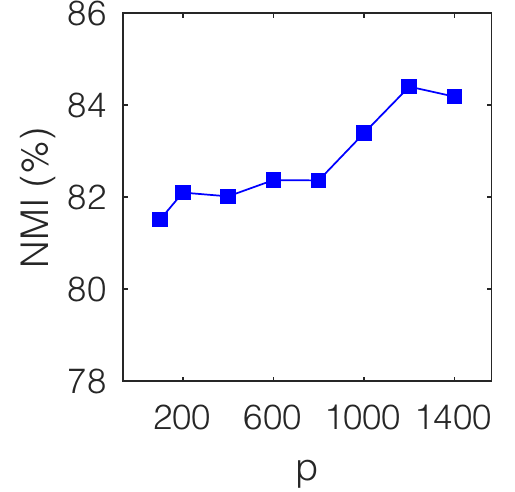}}}
		{\subfigure[\scriptsize {\emph{NUS-WIDE}}]
			{\includegraphics[width=0.238\columnwidth]{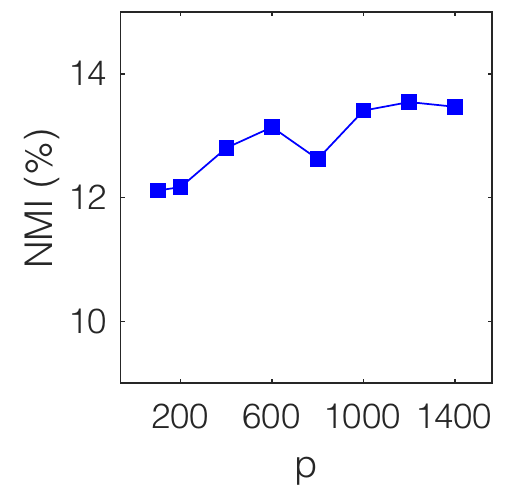}}}
		{\subfigure[\scriptsize \emph{VGGFace2-50}]
			{\includegraphics[width=0.238\columnwidth]{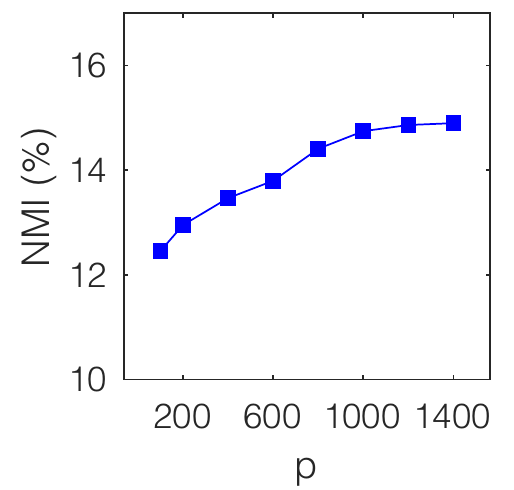}}}\\
		{\subfigure[\scriptsize \emph{VGGFace2-100}]
			{\includegraphics[width=0.238\columnwidth]{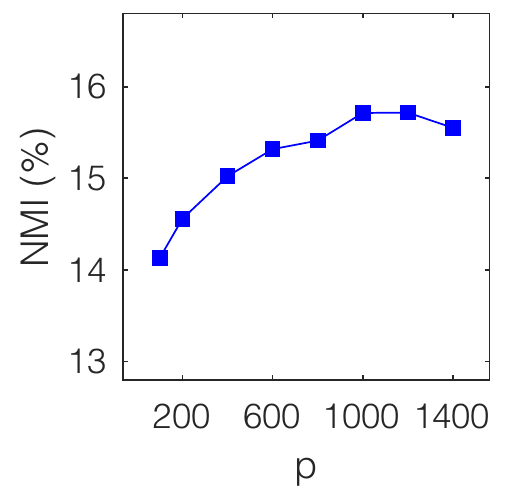}}}
		{\subfigure[\scriptsize {\emph{VGGFace2-200}}]
			{\includegraphics[width=0.238\columnwidth]{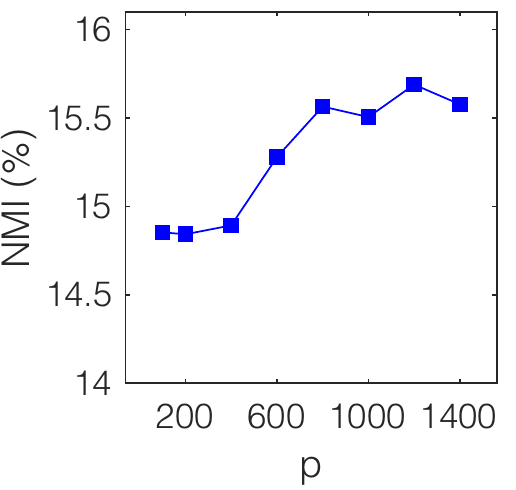}}}
		{\subfigure[\scriptsize \emph{CIFAR-10}]
			{\includegraphics[width=0.238\columnwidth]{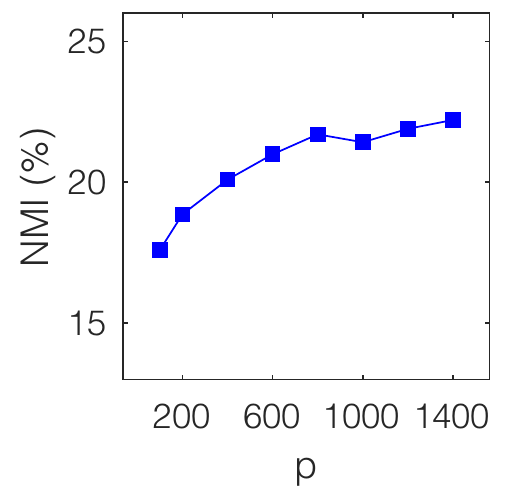}}}
		{\subfigure[\scriptsize \emph{CIFAR-100}]
			{\includegraphics[width=0.238\columnwidth]{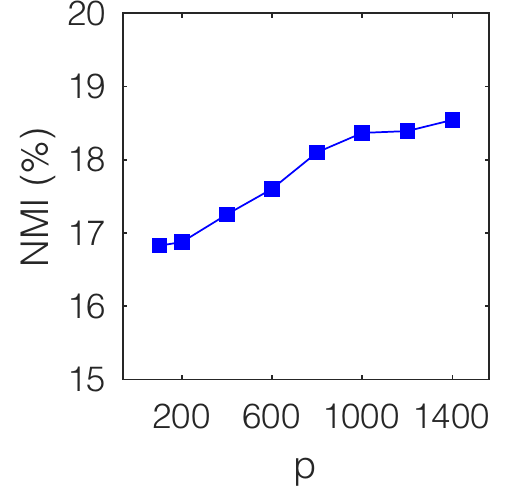}}}
		{\subfigure[\scriptsize \emph{YTF-10}]
			{\includegraphics[width=0.238\columnwidth]{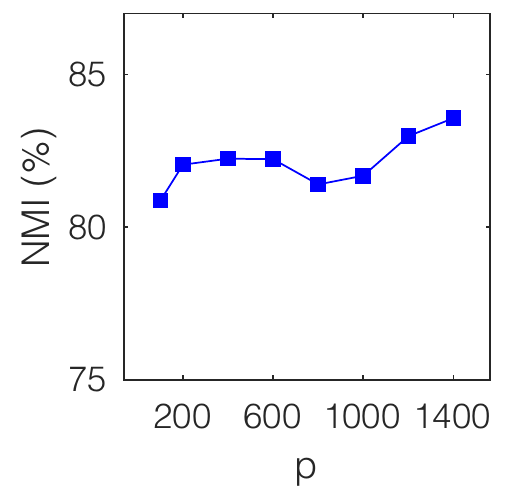}}}
		{\subfigure[\scriptsize \emph{YTF-20}]
			{\includegraphics[width=0.238\columnwidth]{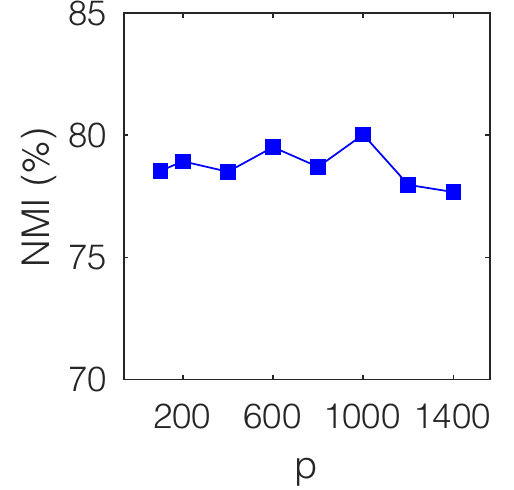}}}
		{\subfigure[\scriptsize \emph{YTF-50}]
			{\includegraphics[width=0.238\columnwidth]{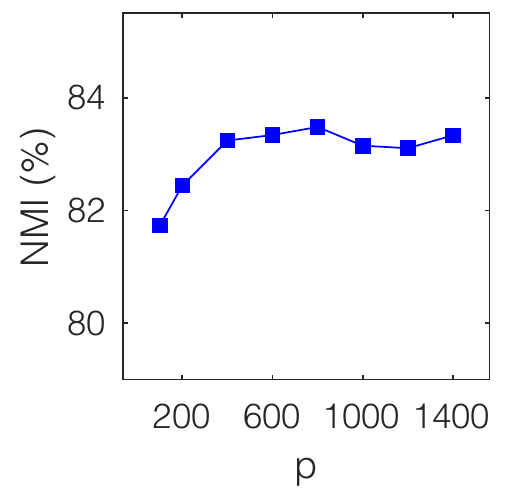}}}
		{\subfigure[\scriptsize \emph{YTF-100}]
			{\includegraphics[width=0.238\columnwidth]{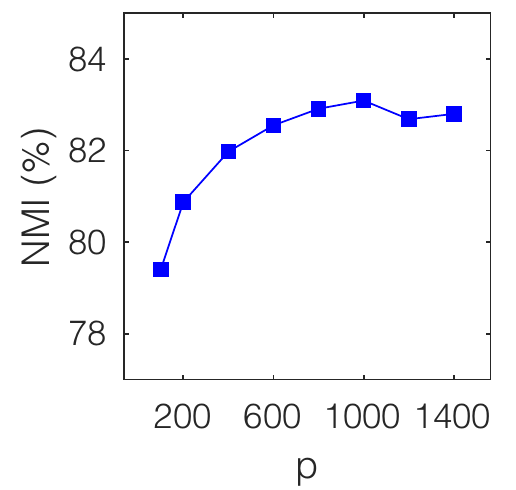}}}\\
		{\subfigure[\scriptsize \emph{YTF-200}]
			{\includegraphics[width=0.238\columnwidth]{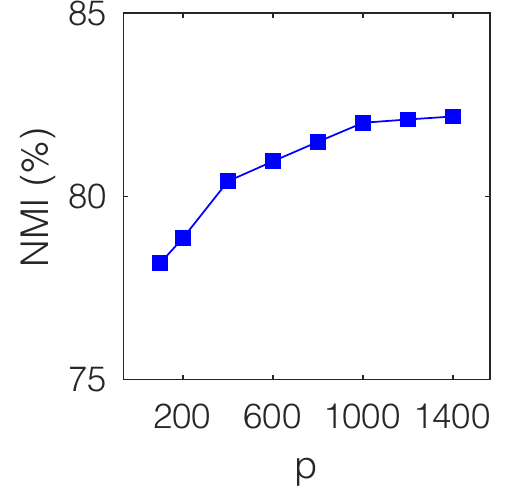}}}
		{\subfigure[\scriptsize \emph{YTF-400}]
			{\includegraphics[width=0.238\columnwidth]{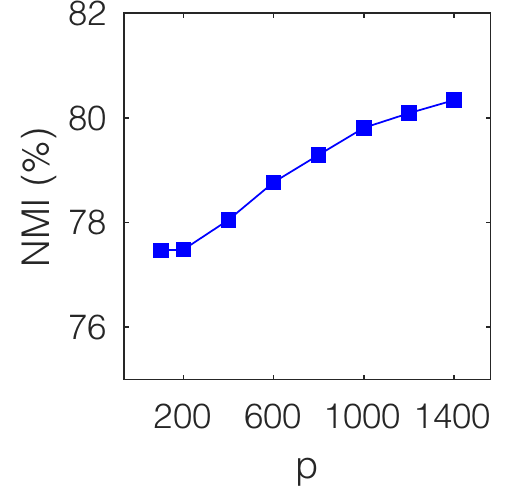}}}\\
		\caption{The average NMI (\%) scores over 20 runs by the proposed algorithm with varying number of anchors $p$.}
		\label{fig:comp_nmi_Psize}
	\end{center}
\end{figure}

\subsection{Performance Comparison and Analysis}

In this section, we experimentally compare our FastMICE method against ten baseline MVC methods. The comparison results w.r.t. NMI, ARI, ACC, and PUR are reported in Tables~\ref{table:compare_nmi}, \ref{table:compare_ari}, \ref{table:compare_acc}, and \ref{table:compare_pur}, respectively.

As shown in Table~\ref{table:compare_nmi}, FastMICE achieves the best performance w.r.t. NMI on 19 out of the 22 datasets. Though the MVEC method yields a higher NMI score than FastMICE on the \emph{BBCSport} dataset, yet FastMICE outperforms or significantly outperforms MVEC on all the other datasets. In comparison with the four large-scale baseline methods, namely, BMVC, LMVSC, SMVSC, and FPMVS-CAG, our FastMICE method outperforms these methods on all benchmark datasets except \emph{Movies} and \emph{NUS-WIDE}. 

Further, we report the average ranks (across the twenty datasets) of the proposed method and the ten baseline methods in Table~\ref{table:compare_nmi}. Note that, for a dataset, if four MVC methods are computationally feasible and seven MVC methods are computationally infeasible due to the out-of-memory error, then the seven infeasible methods will equally rank in the fifth position on this dataset. As can be seen in Table~\ref{table:compare_nmi}, the proposed FastMICE method achieves an average rank of 1.23, which significantly outperforms the second best method with an average rank of 4.41.

Similar advantages of the proposed method over the baselines can also be observed in Tables~\ref{table:compare_ari}, \ref{table:compare_acc}, and \ref{table:compare_pur}.  In terms of ARI, our FastMICE method achieves the best scores on 18 out of the 22 datasets, with an average rank of 1.27. In terms of ACC, our FastMICE method achieves the best scores on 14 out of the 22 datasets, with an average rank of 1.55. In terms of PUR, our FastMICE method achieves the best scores on 14 out of the twenty datasets, with an average rank of 1.45. The experimental results in Tables~\ref{table:compare_nmi}, \ref{table:compare_ari}, \ref{table:compare_acc}, and \ref{table:compare_pur} have confirmed the robust performance of our FastMICE method when compared with the other MVC methods.

\begin{figure}[!t] \scriptsize 
	\begin{center}
		{\subfigure[\scriptsize \emph{MSRCv1}]
			{\includegraphics[width=0.238\columnwidth]{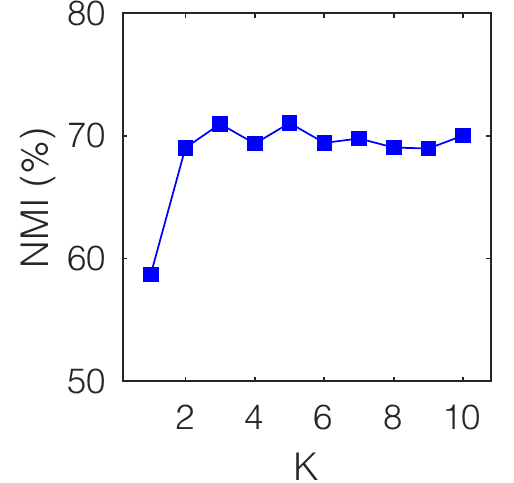}}}
		{\subfigure[\scriptsize \emph{Yale}]
			{\includegraphics[width=0.238\columnwidth]{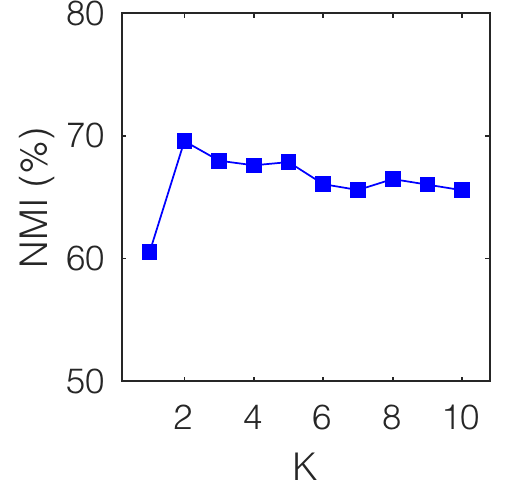}}}
		{\subfigure[\scriptsize \emph{ORL}]
			{\includegraphics[width=0.238\columnwidth]{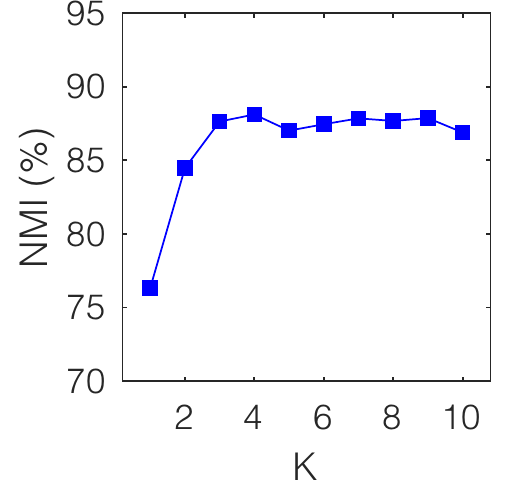}}}
		{\subfigure[\scriptsize \emph{Movies}]
			{\includegraphics[width=0.238\columnwidth]{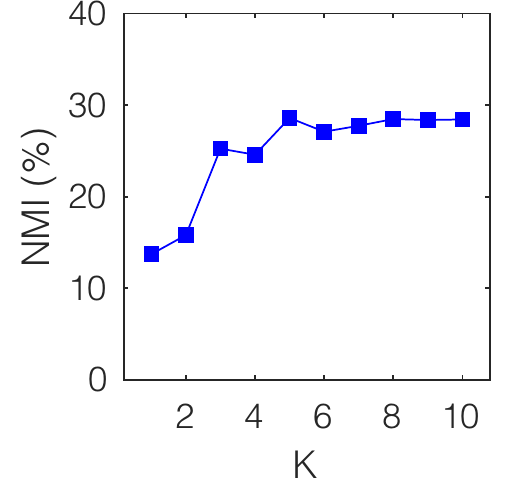}}}\\ 
		{\subfigure[\scriptsize \emph{BBCSport}]
			{\includegraphics[width=0.238\columnwidth]{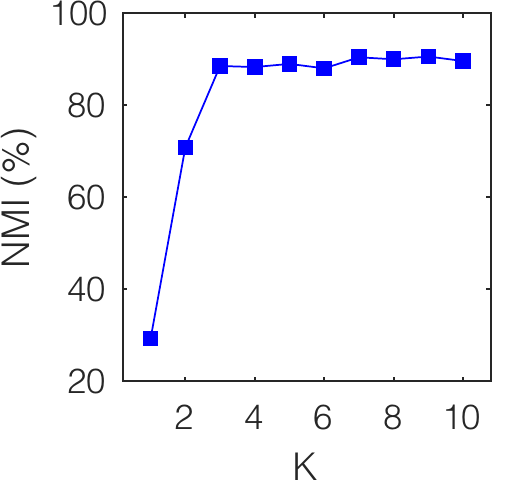}}}
		{\subfigure[\scriptsize \emph{NH-550}]
			{\includegraphics[width=0.238\columnwidth]{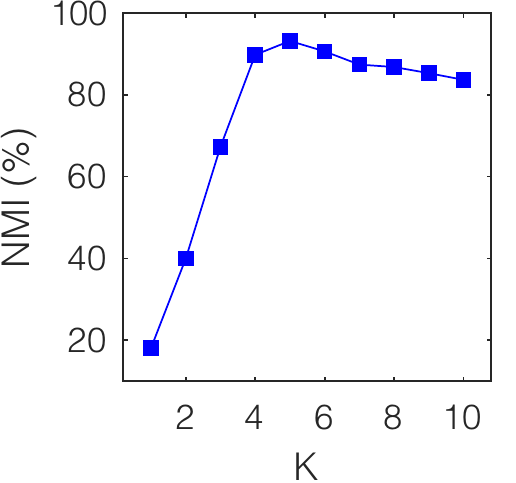}}}
		{\subfigure[\scriptsize \emph{Out-Scene}]
			{\includegraphics[width=0.238\columnwidth]{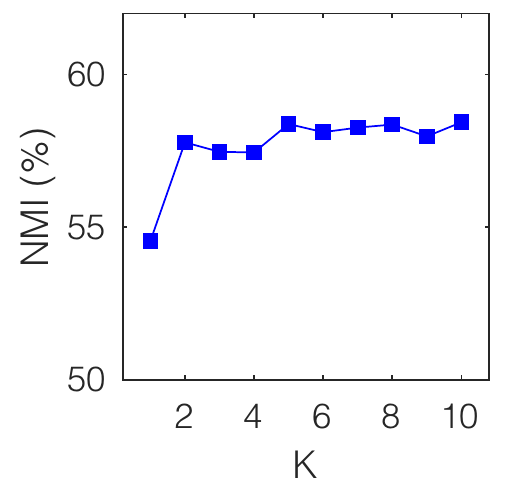}}}
		{\subfigure[\scriptsize \emph{Citeseer}]
			{\includegraphics[width=0.238\columnwidth]{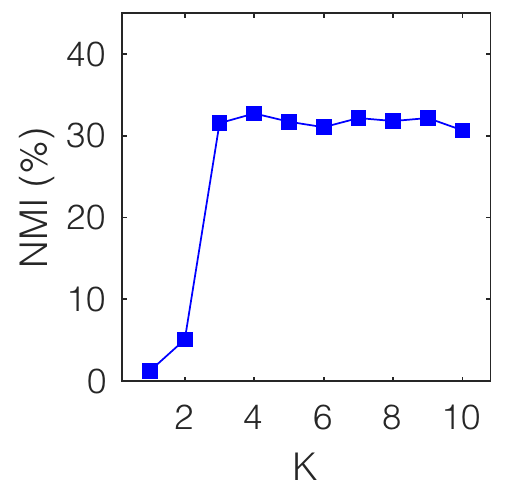}}}\\
		{\subfigure[\scriptsize \emph{NH-4660}]
			{\includegraphics[width=0.238\columnwidth]{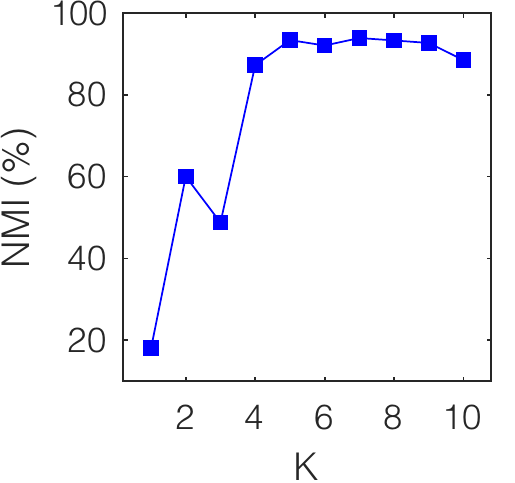}}}
		{\subfigure[\scriptsize \emph{ALOI}]
			{\includegraphics[width=0.238\columnwidth]{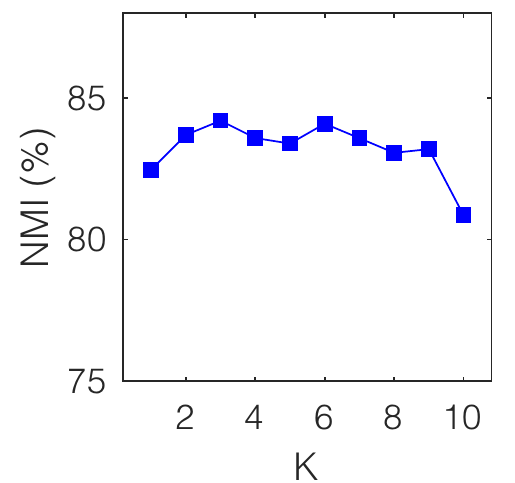}}}
		{\subfigure[\scriptsize {\emph{NUS-WIDE}}]
			{\includegraphics[width=0.238\columnwidth]{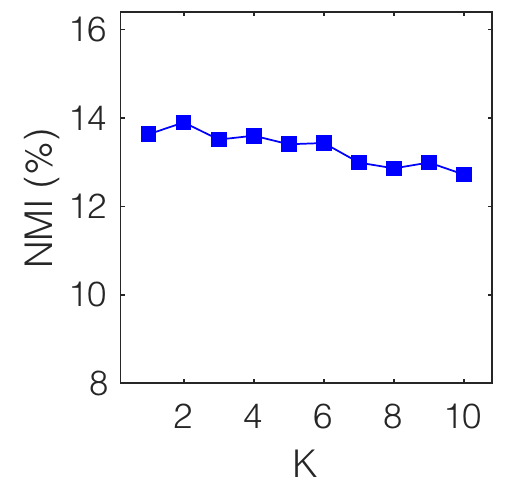}}}
		{\subfigure[\scriptsize \emph{VGGFace2-50}]
			{\includegraphics[width=0.238\columnwidth]{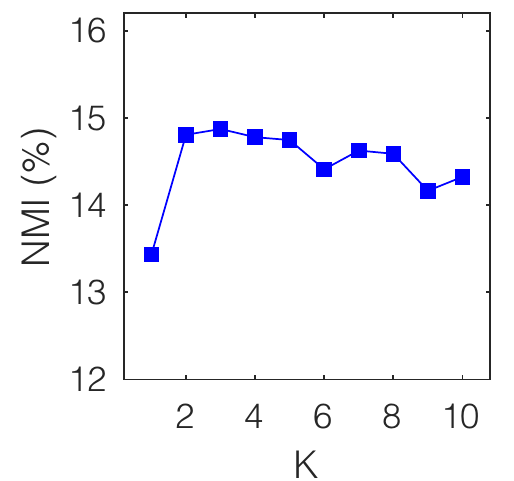}}}\\
		{\subfigure[\scriptsize \emph{VGGFace2-100}]
			{\includegraphics[width=0.238\columnwidth]{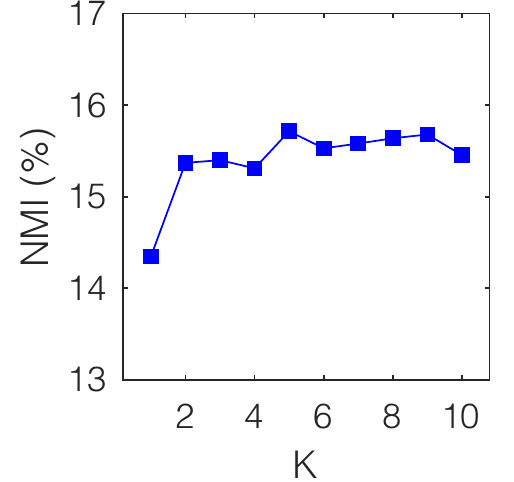}}}
		{\subfigure[\scriptsize {\emph{VGGFace2-200}}]
			{\includegraphics[width=0.238\columnwidth]{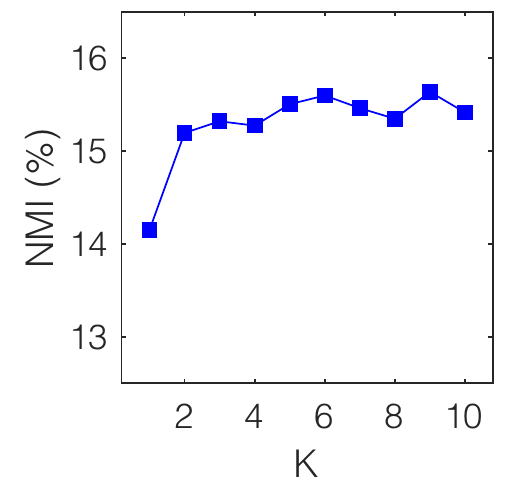}}}
		{\subfigure[\scriptsize \emph{CIFAR-10}]
			{\includegraphics[width=0.238\columnwidth]{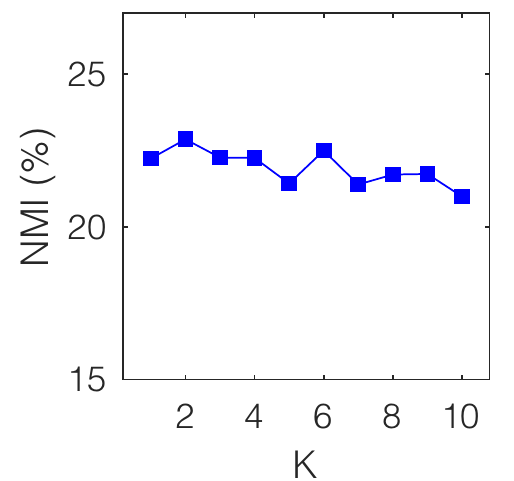}}}
		{\subfigure[\scriptsize \emph{CIFAR-100}]
			{\includegraphics[width=0.238\columnwidth]{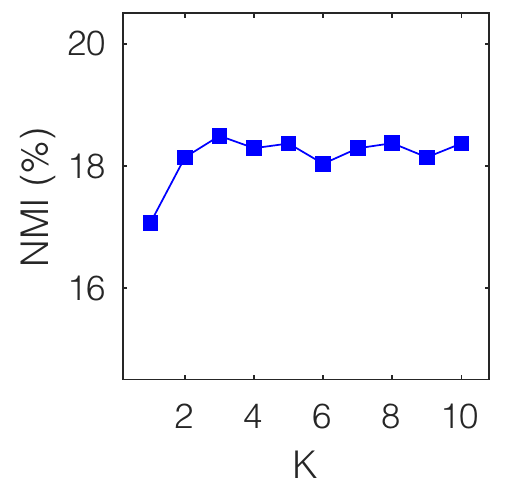}}}\\ 
		{\subfigure[\scriptsize \emph{YTF-10}]
			{\includegraphics[width=0.238\columnwidth]{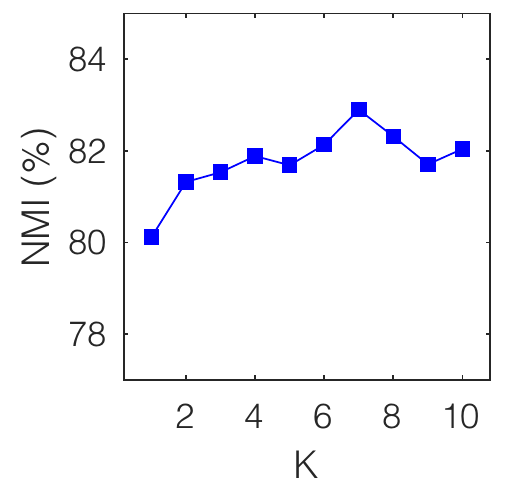}}}
		{\subfigure[\scriptsize \emph{YTF-20}]
			{\includegraphics[width=0.238\columnwidth]{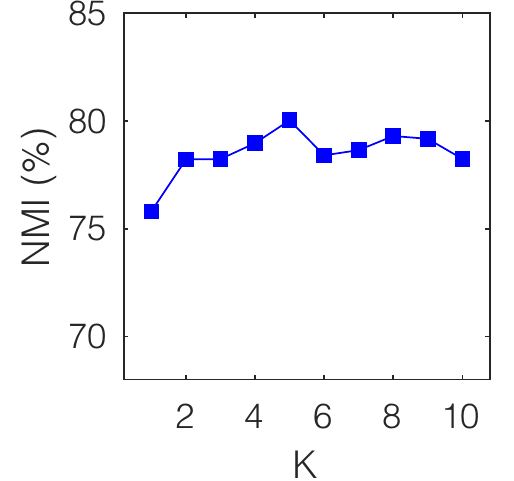}}}
		{\subfigure[\scriptsize \emph{YTF-50}]
			{\includegraphics[width=0.238\columnwidth]{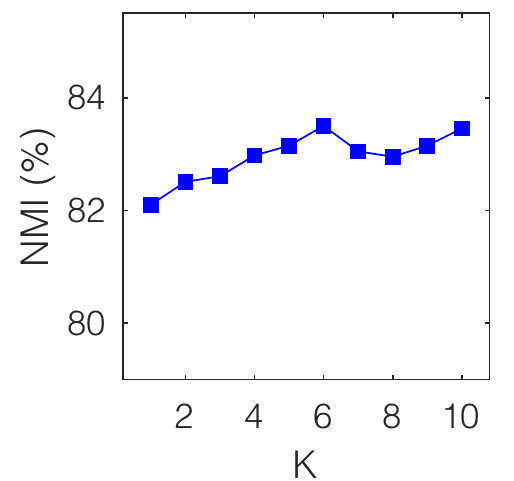}}}
		{\subfigure[\scriptsize \emph{YTF-100}]
			{\includegraphics[width=0.238\columnwidth]{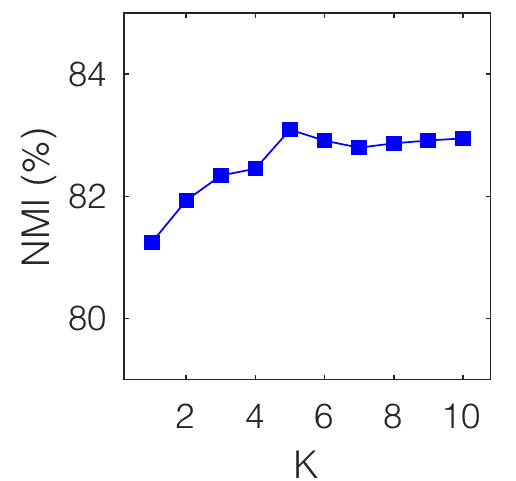}}}\\ 
		{\subfigure[\scriptsize \emph{YTF-200}]
			{\includegraphics[width=0.238\columnwidth]{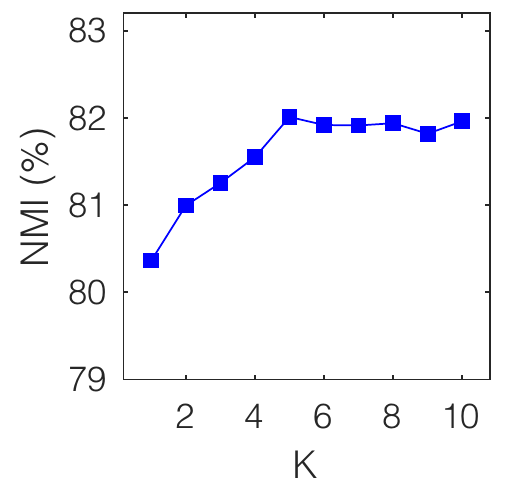}}}
		{\subfigure[\scriptsize \emph{YTF-400}]
			{\includegraphics[width=0.238\columnwidth]{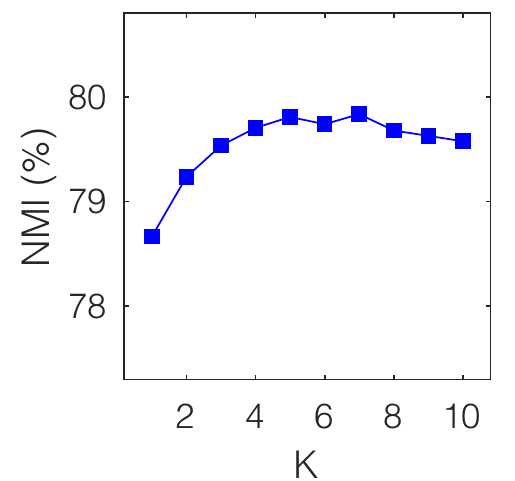}}}\\
		\caption{The average NMI (\%) scores over 20 runs by the proposed algorithm with varying number of nearest neighbors $K$.}
		\label{fig:comp_nmi_Knn}
	\end{center}
\end{figure}

\begin{figure}[!t] \small 
	\begin{center}
		{\subfigure[\scriptsize \emph{MSRCv1}]
			{\includegraphics[width=0.238\columnwidth]{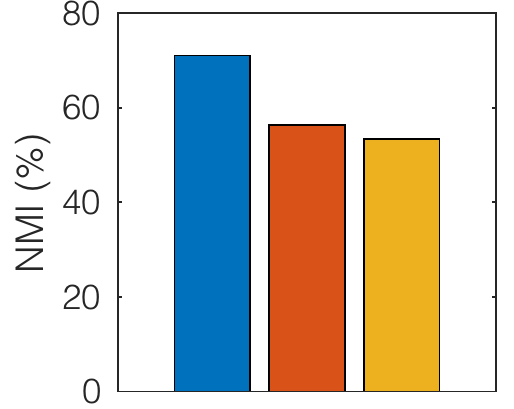}}}
		{\subfigure[\scriptsize \emph{Yale}]
			{\includegraphics[width=0.238\columnwidth]{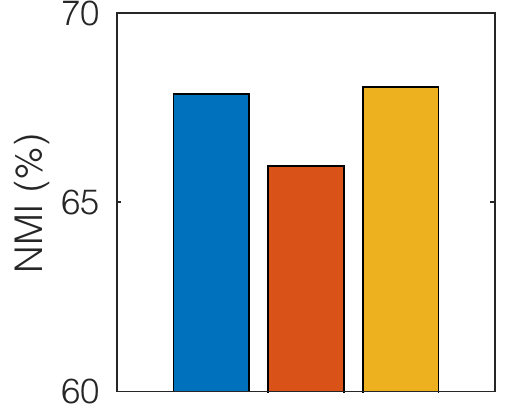}}}
		{\subfigure[\scriptsize \emph{ORL}]
			{\includegraphics[width=0.238\columnwidth]{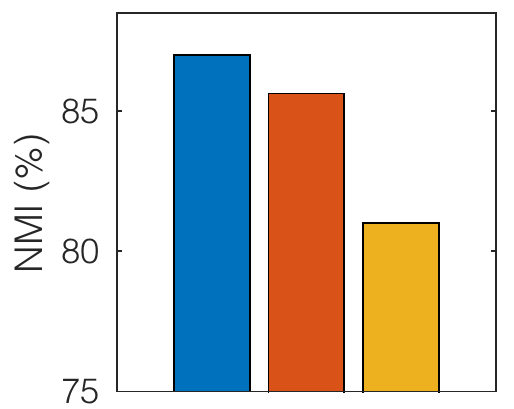}}}
		{\subfigure[\scriptsize \emph{Movies}]
			{\includegraphics[width=0.238\columnwidth]{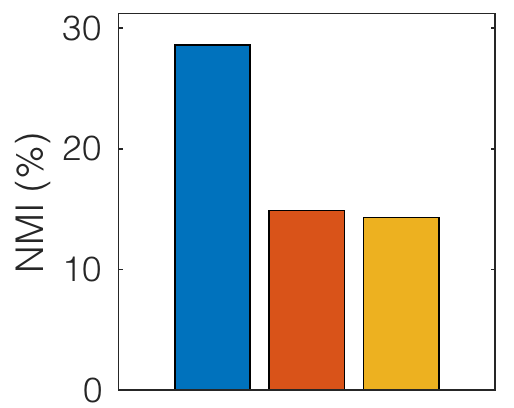}}}\\
		{\subfigure[\scriptsize \emph{BBCSport}]
			{\includegraphics[width=0.238\columnwidth]{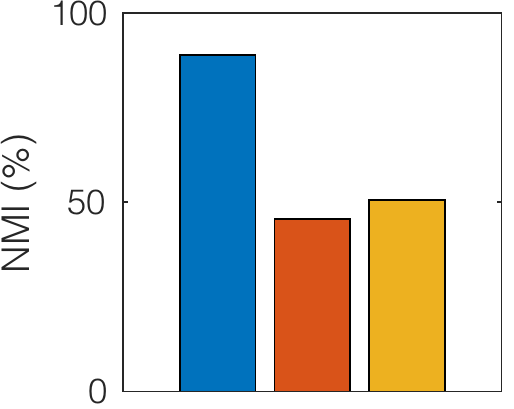}}}
		{\subfigure[\scriptsize \emph{NH-550}]
			{\includegraphics[width=0.238\columnwidth]{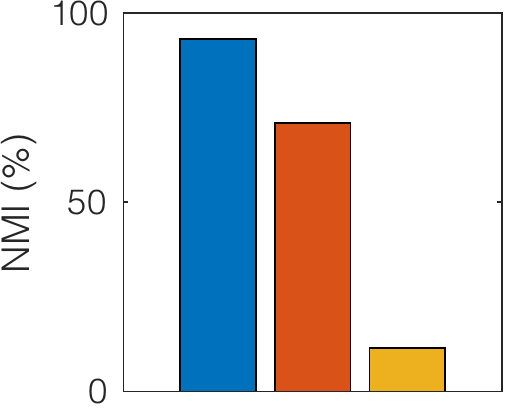}}}
		{\subfigure[\scriptsize \emph{Out-Scene}]
			{\includegraphics[width=0.238\columnwidth]{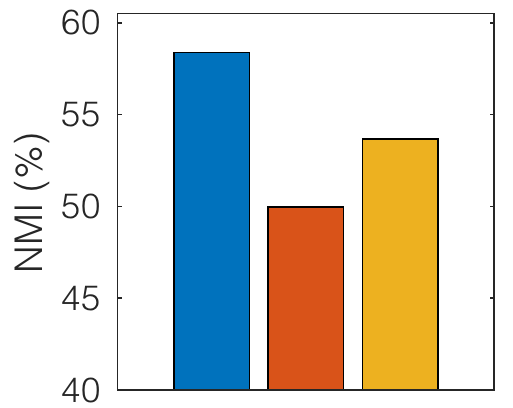}}}
		{\subfigure[\scriptsize \emph{Citeseer}]
			{\includegraphics[width=0.238\columnwidth]{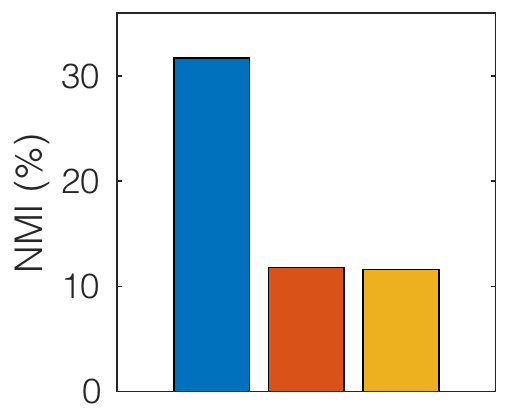}}}\\ 
		{\subfigure[\scriptsize \emph{NH-4660}]
			{\includegraphics[width=0.238\columnwidth]{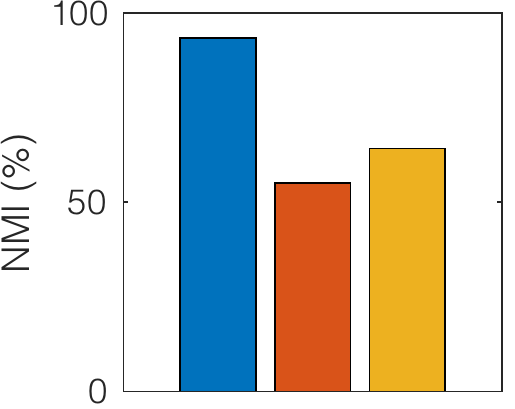}}}
		{\subfigure[\scriptsize \emph{ALOI}]
			{\includegraphics[width=0.238\columnwidth]{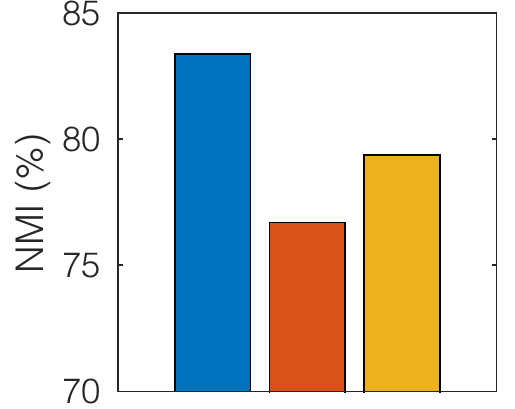}}}
		{\subfigure[\scriptsize {\emph{NUS-WIDE}}]
			{\includegraphics[width=0.238\columnwidth]{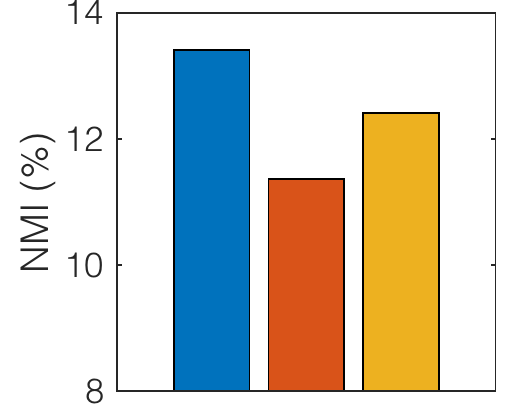}}}
		{\subfigure[\scriptsize \emph{VGGFace2-50}]
			{\includegraphics[width=0.238\columnwidth]{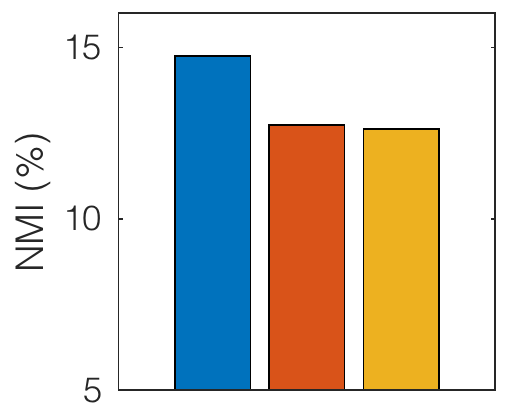}}}\\ 
		{\subfigure[\scriptsize \emph{VGGFace2-100}]
			{\includegraphics[width=0.238\columnwidth]{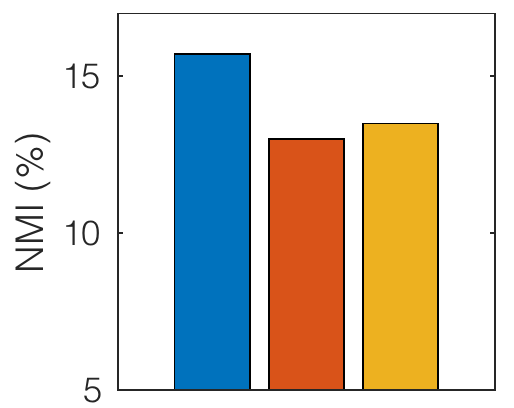}}}
		{\subfigure[\scriptsize {\emph{VGGFace2-200}}]
			{\includegraphics[width=0.238\columnwidth]{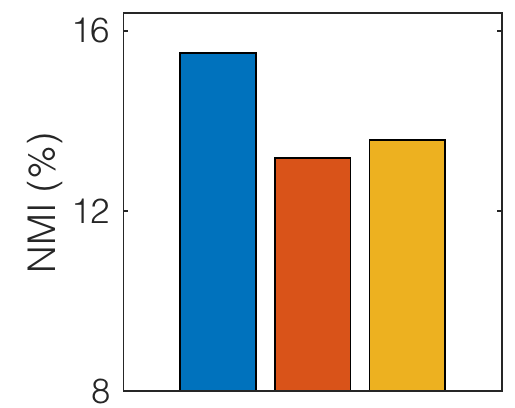}}}
		{\subfigure[\scriptsize \emph{CIFAR-10}]
			{\includegraphics[width=0.238\columnwidth]{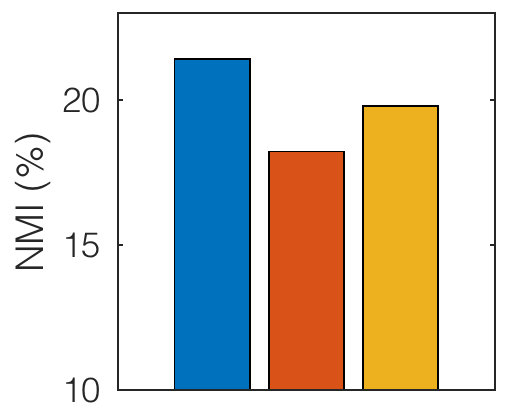}}}
		{\subfigure[\scriptsize \emph{CIFAR-100}]
			{\includegraphics[width=0.238\columnwidth]{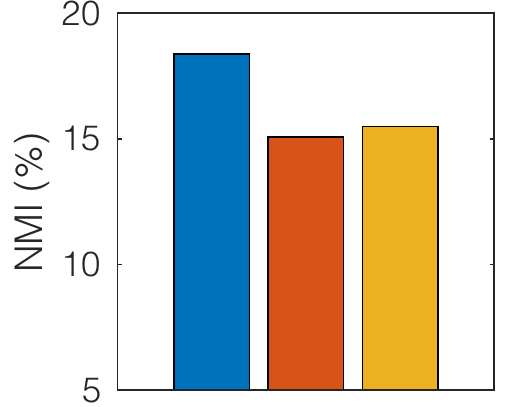}}}\\ 
		{\subfigure[\scriptsize \emph{YTF-10}]
			{\includegraphics[width=0.238\columnwidth]{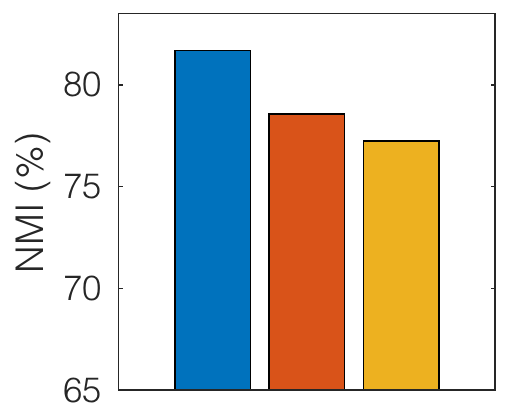}}}
		{\subfigure[\scriptsize \emph{YTF-20}]
			{\includegraphics[width=0.238\columnwidth]{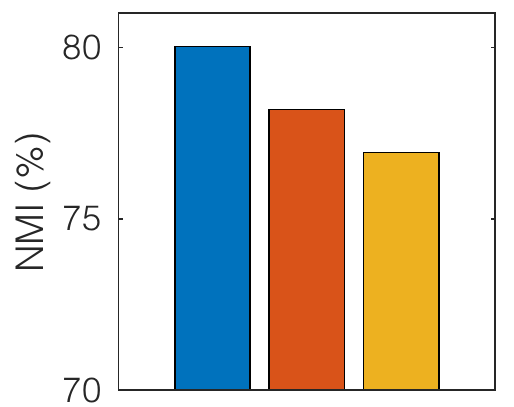}}}
		{\subfigure[\scriptsize \emph{YTF-50}]
			{\includegraphics[width=0.238\columnwidth]{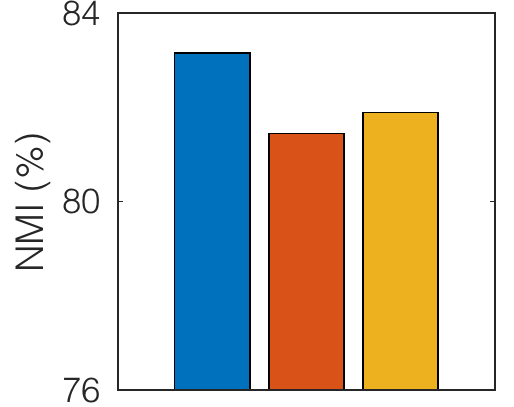}}}
		{\subfigure[\scriptsize \emph{YTF-100}]
			{\includegraphics[width=0.238\columnwidth]{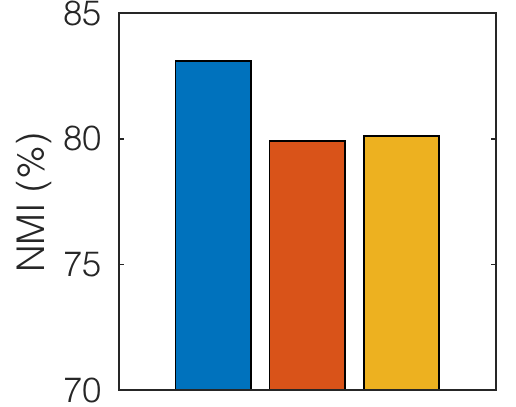}}}\\ 
		{\subfigure[\scriptsize \emph{YTF-200}]
			{\includegraphics[width=0.238\columnwidth]{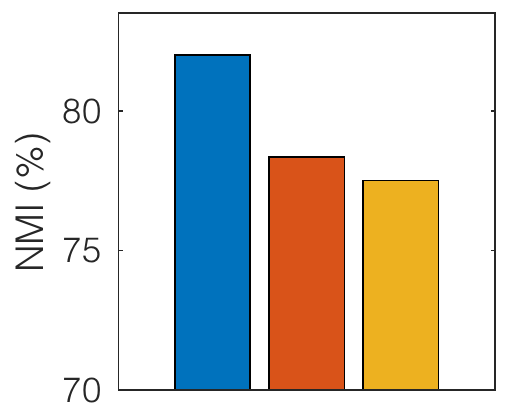}}}
		{\subfigure[\scriptsize \emph{YTF-400}]
			{\includegraphics[width=0.238\columnwidth]{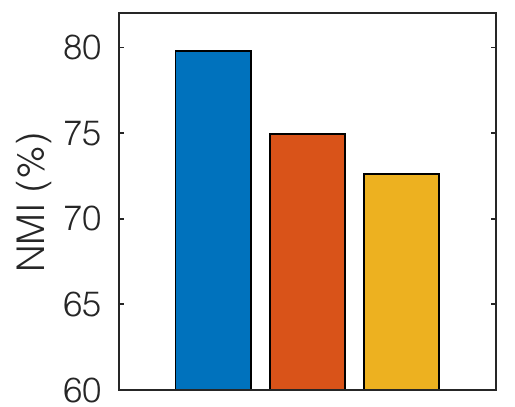}}}\\ 
		{\subfigure
			{\includegraphics[width=0.9\columnwidth]{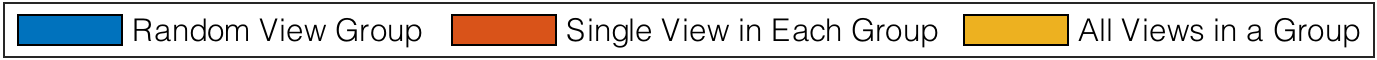}}}
		\caption{The comparison (w.r.t. NMI (\%)) of the proposed algorithm using random view groups against treating each single view as a group and treating all views as a group.}
		\label{fig:comp_nmi_sglFull}
	\end{center}
\end{figure}

\begin{table*}[!t]\scriptsize
	\centering
	\caption{Time costs (in seconds) of different multi-view clustering algorithms. \emph{\textbf{TuneTime}}: Time cost for dataset-specific hyperparameter tuning. \emph{\textbf{RunTime}}: Time cost for running the algorithm.}\vskip -0.1 in
	\label{table:compare_time}
	\begin{threeparttable}\renewcommand\arraystretch{1.05}
		\begin{tabular}{|m{1.45cm}<{\centering}|m{0.92cm}<{\centering}|m{0.79cm}<{\centering}|m{0.79cm}<{\centering}|m{0.79cm}<{\centering}|m{1.15cm}<{\centering}|m{1cm}<{\centering}m{1.19cm}<{\centering}|m{0.81cm}<{\centering}|m{0.79cm}<{\centering}|m{0.95cm}<{\centering}|m{0.91cm}<{\centering}|m{1cm}<{\centering}|}
			\hline
			Dataset    & &MVSC     &AMGL  &SwMC &MVEC     &M$^2$VEC$_{km}$     &M$^2$VEC$_{spec}$     &BMVC   &LMVSC  &SMVSC   &FPMVS-CAG  &FastMICE\\
			\hline
			\hline
			\multirow{2}{*}{\emph{MSRCv1}}	&\emph{TuneTime} 		&2.10	&-	&-	&226.19	&\multicolumn{2}{c|}{8,252.72}	&54.95	&8.65	&3.75	&-	&-\\
			\cline{7-8}
			&\emph{RunTime} 		&0.26	&0.16	&2.46	&3.19	&12.83	&12.86	&0.06	&0.46	&1.17	&2.31	&0.33\\
			\hline
			\multirow{2}{*}{\emph{Yale}}	&\emph{TuneTime} 		&5.07	&-	&-	&141.30	&\multicolumn{2}{c|}{3,663.22}	&98.64	&7.75	&8.55	&-	&-\\
			\cline{7-8}
			&\emph{RunTime} 		&0.53	&0.08	&1.40	&4.78	&4.65	&4.70	&0.08	&0.56	&2.57	&3.21	&1.81\\
			\hline
			\multirow{2}{*}{\emph{ORL}}	&\emph{TuneTime} 		&27.36	&-	&-	&1,442.16	&\multicolumn{2}{c|}{55,123.16}	&233.75	&16.63	&56.55	&-	&-\\
			\cline{7-8}
			&\emph{RunTime} 		&2.77	&0.40	&7.39	&31.40	&116.87	&116.92	&0.24	&1.50	&10.73	&22.90	&6.67\\
			\hline
			\multirow{2}{*}{\emph{Movies}}	&\emph{TuneTime} 		&13.86	&-	&-	&3,823.81	&\multicolumn{2}{c|}{94,912.10}	&297.74	&16.96	&13.84	&-	&-\\
			\cline{7-8}
			&\emph{RunTime} 		&1.33	&1.59	&15.54	&54.12	&178.80	&179.01	&0.21	&0.74	&4.83	&7.54	&3.18\\
			\hline
			\multirow{2}{*}{\emph{BBCSport}}	&\emph{TuneTime} 		&18.33	&-	&-	&2,267.14	&\multicolumn{2}{c|}{111,283.96}	&234.11	&13.20	&11.18	&-	&-\\
			\cline{7-8}
			&\emph{RunTime} 		&1.81	&0.91	&16.48	&32.82	&178.83	&179.00	&0.22	&0.75	&3.94	&7.09	&6.85\\
			\hline
			\multirow{2}{*}{\emph{NH-550}}	&\emph{TuneTime} 		&33.17	&-	&-	&2,350.40	&\multicolumn{2}{c|}{117,688.03}	&364.87	&19.12	&12.75	&-	&-\\
			\cline{7-8}
			&\emph{RunTime} 		&3.23	&1.53	&22.92	&36.27	&208.96	&209.08	&0.46	&1.78	&4.10	&7.98	&6.38\\
			\hline
			\multirow{2}{*}{\emph{Out-Scene}}	&\emph{TuneTime} 		&92.35	&-	&-	&218,154.40	&\multicolumn{2}{c|}{1,490,922.44}	&1,857.73	&79.59	&47.82	&-	&-\\
			\cline{7-8}
			&\emph{RunTime} 		&9.20	&53.79	&1,644.44	&4,495.78	&34,118.99	&34,140.06	&1.31	&4.97	&14.63	&29.12	&4.93\\
			\hline
			\multirow{2}{*}{\emph{Citeseer}}	&\emph{TuneTime} 		&179.54	&-	&-	&37,1374.01	&\multicolumn{2}{c|}{1,131,867.93}	&2,209.50	&117.07	&66.03	&-	&-\\
			\cline{7-8}
			&\emph{RunTime} 		&17.50	&191.80	&816.35	&7,984.33	&46,788.20	&46,826.12	&1.25	&8.10	&21.41	&42.38	&28.26\\
			\hline
			\multirow{2}{*}{\emph{NH-4660}}	&\emph{TuneTime} 		&522.87	&-	&-	&1,256,609.02	&\multicolumn{2}{c|}{1,418,874.08}	&3,962.23	&249.21	&103.99	&-	&-\\
			\cline{7-8}
			&\emph{RunTime} 		&52.98	&388.50	&5,709.74	&19,621.66	&142,392.16	&142,472.40	&2.20	&23.75	&33.95	&66.69	&27.60\\
			\hline
			\multirow{2}{*}{\emph{ALOI}}	&\emph{TuneTime} 		&2,073.97	&-	&N/A	&N/A	&\multicolumn{2}{c|}{N/A}	&5,822.19	&286.58	&1,355.84	&-	&-\\
			\cline{7-8}
			&\emph{RunTime} 		&205.71	&4,760.38	&N/A	&N/A	&N/A	&N/A	&4.23	&25.46	&124.94	&241.03	&10.32\\
			\hline
			\multirow{2}{*}{\emph{NUS-WIDE}}	&{\emph{TuneTime}} 		&{N/A}	&{N/A}	&{N/A}	&{N/A}	&\multicolumn{2}{c|}{{N/A}}	&{6,478.61}	&{375.52}	&{267.60}	&{-}	&{-}\\
			\cline{7-8}
			&{\emph{RunTime}} 		&{N/A}	&{N/A}	&{N/A}	&{N/A}	&{N/A}	&{N/A}	&{12.73}	&{91.99}	&{203.15}	&{386.47}	&{14.43}\\
			\hline
			\multirow{2}{*}{\emph{VGGFace2-50}}	&\emph{TuneTime} 		&N/A	&N/A	&N/A	&N/A	&\multicolumn{2}{c|}{N/A}	&5,677.58	&348.02	&587.33	&-	&-\\
			\cline{7-8}
			&\emph{RunTime} 		&N/A	&N/A	&N/A	&N/A	&N/A	&N/A	&13.34	&249.97	&491.59	&571.76	&30.69\\
			\hline
			\multirow{2}{*}{\emph{VGGFace2-100}}	&\emph{TuneTime} 		&N/A	&N/A	&N/A	&N/A	&\multicolumn{2}{c|}{N/A}	&5,870.30	&400.46	&2,181.60	&-	&-\\
			\cline{7-8}
			&\emph{RunTime} 		&N/A	&N/A	&N/A	&N/A	&N/A	&N/A	&26.12	&663.49	&3,989.61	&1,788.38	&62.63\\
			\hline
			\multirow{2}{*}{{\emph{VGGFace2-200}}}	&{\emph{TuneTime}} 		&{N/A}	&{N/A}	&{N/A}	&{N/A}	&\multicolumn{2}{c|}{{N/A}}	&{6,334.40}	&{853.80}	&{9,613.44}	&{-}	&{-}\\
			\cline{7-8}
			&{\emph{RunTime}} 		&{N/A}	&{N/A}	&{N/A}	&{N/A}	&{N/A}	&{N/A}	&{53.58}	&{1,091.23}	&{34,208.76}	&{14,987.48}	&{157.55}\\
			\hline
			\multirow{2}{*}{\emph{CIFAR-10}}	&\emph{TuneTime} 		&N/A	&N/A	&N/A	&N/A	&\multicolumn{2}{c|}{N/A}	&5,930.36	&388.57	&200.48	&-	&-\\
			\cline{7-8}
			&\emph{RunTime} 		&N/A	&N/A	&N/A	&N/A	&N/A	&N/A	&22.75	&303.92	&376.17	&719.82	&47.53\\
			\hline
			\multirow{2}{*}{\emph{CIFAR-100}}	&\emph{TuneTime} 		&N/A	&N/A	&N/A	&N/A	&\multicolumn{2}{c|}{N/A}	&6,649.10	&681.23	&2,056.21	&-	&-\\
			\cline{7-8}
			&\emph{RunTime} 		&N/A	&N/A	&N/A	&N/A	&N/A	&N/A	&24.00	&625.90	&3,569.92	&1,680.07	&58.84\\
			\hline
			\multirow{2}{*}{\emph{YTF-10}}	&\emph{TuneTime} 		&N/A	&N/A	&N/A	&N/A	&\multicolumn{2}{c|}{N/A}	&5,786.88	&315.67	&198.01	&-	&-\\
			\cline{7-8}
			&\emph{RunTime} 		&N/A	&N/A	&N/A	&N/A	&N/A	&N/A	&13.74	&198.22	&258.70	&460.84	&23.53\\
			\hline
			\multirow{2}{*}{\emph{YTF-20}}	&\emph{TuneTime} 		&N/A	&N/A	&N/A	&N/A	&\multicolumn{2}{c|}{N/A}	&5,537.66	&335.58	&236.48	&-	&-\\
			\cline{7-8}
			&\emph{RunTime} 		&N/A	&N/A	&N/A	&N/A	&N/A	&N/A	&22.69	&324.86	&591.16	&831.06	&35.05\\
			\hline
			\multirow{2}{*}{\emph{YTF-50}}	&\emph{TuneTime} 		&N/A	&N/A	&N/A	&N/A	&\multicolumn{2}{c|}{N/A}	&5,864.39	&261.78	&549.79	&-	&-\\
			\cline{7-8}
			&\emph{RunTime} 		&N/A	&N/A	&N/A	&N/A	&N/A	&N/A	&47.57	&751.22	&1,107.12	&2,143.80	&70.89\\
			\hline
			\multirow{2}{*}{\emph{YTF-100}}	&\emph{TuneTime} 		&N/A	&N/A	&N/A	&N/A	&\multicolumn{2}{c|}{N/A}	&6,131.04	&323.94	&1,970.67	&-	&-\\
			\cline{7-8}
			&\emph{RunTime} 		&N/A	&N/A	&N/A	&N/A	&N/A	&N/A	&72.69	&1,242.44	&2,863.95	&5,415.22	&124.48\\
			\hline
			\multirow{2}{*}{\emph{YTF-200}}	&\emph{TuneTime} 		&N/A	&N/A	&N/A	&N/A	&\multicolumn{2}{c|}{N/A}	&N/A	&816.28	&8,587.07	&-	&-\\
			\cline{7-8}
			&\emph{RunTime} 		&N/A	&N/A	&N/A	&N/A	&N/A	&N/A	&N/A	&3,184.79	&74,063.30	&32,974.03	&263.81\\
			\hline
			\multirow{2}{*}{\emph{YTF-400}}	&\emph{TuneTime} 		&N/A	&N/A	&N/A	&N/A	&\multicolumn{2}{c|}{N/A}	&N/A	&N/A	&N/A	&N/A	&-\\
			\cline{7-8}
			&\emph{RunTime} 		&N/A	&N/A	&N/A	&N/A	&N/A	&N/A	&N/A	&N/A	&N/A	&N/A	&652.05\\
			\hline
		\end{tabular}
		\begin{tablenotes}
			\item[*] Note that N/A indicates the out-of-memory error, and the symbol ``-'' indicates that no dataset-specific tuning is needed.
			\item[**] If the dataset-specific tuning is needed, the hyperparameters will be tuned in $\{10^{-5},10^{-4},\cdots,10^{5}\}$, unless the tuning range is specifically given by the corresponding paper.
			\item[***] For the baseline methods except M$^2$VEC$_{km}$  and M$^2$VEC$_{spec}$,  if $N>$ 10,000,  the tuning (if needed) will be conducted on a random subset of 10,000 samples. For M$^2$VEC$_{km}$  and M$^2$VEC$_{spec}$,  if $N>$ 1,000,  the tuning will be conducted on a random subset of 1,000 samples.
			\item[****] The difference between M$^2$VEC$_{km}$  and M$^2$VEC$_{spec}$ lies in the last step, where either $k$-means or spectral clustering is used. Thus we tune them together.
		\end{tablenotes}
	\end{threeparttable}
\end{table*}

\subsection{Influence of Ensemble Size $M$}

In this section, we test the influence of the ensemble size $M$, which corresponds to the number of base clusterings and also the number of random view groups in FastMICE. The three EC based baseline methods, namely, MVEC, M$^2$VEC$_{km}$, and M$^2$VEC$_{spec}$, similarly involve the ensemble size $M$, which will also be tested in the comparison.

The performances of FastMICE and the three baselines are illustrated in Fig.~\ref{fig:comp_nmi_Msize}. Note that the three EC based baseline methods are not computationally feasible for datasets larger than \emph{NH-4660}, so their curves will be absent in the corresponding sub-figures. As shown in Fig.~\ref{fig:comp_nmi_Msize}, the proposed FastMICE method yields consistently high-quality clustering performance with varying ensemble sizes. When compared with the other EC based methods, FastMICE achieves overall better performance than the baselines on the benchmark datasets.  Empirically, a relatively larger ensemble size is beneficial. In our experiments, we use the ensemble size $M=20$ on all benchmark datasets.

\subsection{Influence of Number of Anchors $p$}

In this section, we test the influence of the number of anchors $p$ in the FastMICE method. As can be  seen in Fig.~\ref{fig:comp_nmi_Psize}, the proposed FastMICE method shows consistent performance as the number of anchors goes from 100 to 1400. Empirically, a larger number of anchors can be beneficial on most of the datasets, especially on the large-scale ones such as \emph{YTF-200} and \emph{YTF-400}, probably due to the fact that a larger number of anchors may better reflect the overall structure of the data. 
As the number of anchors cannot exceed the number of original samples, in our experiments, we use the number of anchors $p=\min\{1000,N\}$ on all benchmark datasets.

\subsection{Influence of Number of Nearest Neighbors $K$}

In this section, we test the influence of the number of nearest neighbors  $K$ in the FastMICE method, which corresponds to the number of (nearest) anchors that are linked to each data sample. Specifically, we illustrate the performance of the proposed FastMICE method as the number of nearest neighbors goes from 1 to 10  in Fig.~\ref{fig:comp_nmi_Knn}. As shown in Fig.~\ref{fig:comp_nmi_Knn}, a moderate value of $K$ can often be beneficial on most of the benchmark datasets. Empirically, it is suggested that the number of nearest neighbors be set in the range of $[4,8]$. In our experiments, we use the number of nearest neighbors  $K=5$ on all benchmark datasets.

\subsection{Influence of Random View Groups}

In this section, we test the influence of the random view groups in our framework. As discussed in Section~\ref{sec:view_group_formation}, the previous early fusion methods can be regarded as a special instance of our view group formation with all views in a single group, while the late fusion methods can also be regarded as a special instance of our view group formation with a single view in each group. In this section, we compare our FastMICE method using random view groups against using a single view in each group and using all views in a group. As shown in Fig.~\ref{fig:comp_nmi_sglFull}, the use of random view groups leads to substantial improvements on most of the datasets. Though the use of all views in a group leads to comparable performance to the use of random view groups on the \emph{Yale} dataset, yet on most of the other datasets the FastMICE method using random view groups outperforms or significantly outperforms the variants using a single view in each group or using all views in a group, which verifies the benefits brought in by the random view groups.

\subsection{Execution Time}

In this section, we evaluate the time costs of the FastMICE method and the other MVC methods on the benchmark datasets. Note that for the proposed FastMICE method, no dataset-specific hyperparameter-tuning is needed. For the baseline methods, if the dataset-specific hyperparameter-tuning is required, then the time costs of tuning and running will be respectively reported.

As shown in Table~\ref{table:compare_time}, more than half of the baseline methods cannot go beyond the \emph{ALOI} dataset due to the computational complexity bottleneck. Though the large-scale baseline methods, including BMVC, LMVSC, SMVSC, and FPMVS-CAG, have shown their scalability for some larger datasets, yet they still encounter heavy computational burdens or even the out-of-memory error when they process the \emph{YTF-200} or \emph{YTF-400} datasets. Remarkably, on the  \emph{YTF-200} dataset with 286,006 samples, our FastMICE method only consumes 263.81 seconds of running time, while the LMVSC, SMVSC, and FPMVS-CAG methods respectively consume 3,184.79 seconds, 74,063.30 seconds, and 32,974.03 seconds of running times. {In terms of the \emph{YTF-400} dataset with 398,191 samples, FastMICE is the only method that is computationally feasible on this large-scale dataset, which demonstrates the clear advantage of our FastMICE method in scalability for very large-scale datasets.}

To summarize, as shown in Tables~\ref{table:compare_nmi}, \ref{table:compare_ari}, \ref{table:compare_acc}, \ref{table:compare_pur}, and \ref{table:compare_time}, the proposed FastMICE method is capable of yielding highly competitive clustering performance over the state-of-the-art, while showing advantageous scalability on very large-scale multi-view datasets.

\section{Conclusion and Future Work}
\label{sec:conclusion}
In this paper, we propose a new large-scale MVC approach termed FastMICE, which is featured by its scalability (for extremely large-scale datasets), superiority (in clustering performance), and simplicity (to be applied without dataset-specific tuning). In particular, different from the previous approaches that mostly adopt some types of single-stage fusion strategies, this paper presents a hybrid early-late fusion strategy based on the random view groups. A large number of random view groups are first formed to serve as a flexible view-organizations to investigate the view-wise relationships. Then, three levels of diversity, i.e., the feature-level diversity, the anchor-level diversity, and the neighborhood-level diversity, are jointly leveraged to explore the rich and versatile information in the random view groups and thereby enable the highly efficient construction of the view-sharing bipartite graphs. By fast partitioning of these view-sharing bipartite graphs from different view groups, a set of diversified base clusterings can be generated, which are further formulated into a unified bipartite graph for achieving the final clustering result. 

{It is noteworthy that our FastMICE approach has almost linear time and space complexity, and is able to perform robustly and accurately on various general-scale and large-scale datasets without requiring dataset-specific hyperparameter-tuning. Extensive experimental results on 22 multi-view datasets have demonstrated the superiority of our FastMICE approach over the state-of-the-art. In the future work, the concept of random view groups and the diversification-and-fusion strategy may also be investigated for more MVC tasks, such as the incomplete MVC task \cite{yang22TPAMI} and the deep MVC task \cite{lin21CVPR,lin22TPAMI}, so as to promote their robustness while ensuring scalability for very large datasets.}

\ifCLASSOPTIONcompsoc
  \section*{Acknowledgments}
\else
  \section*{Acknowledgment}
\fi

This project was supported by the NSFC (61976097, 62276277  \& U22A2095), the National Key Research and Development Program of China (2021YFF1201200), the Natural Science Foundation of Guangdong Province (2021A1515012203), and the Science and Technology Program of Guangzhou, China (202201010314).

\ifCLASSOPTIONcaptionsoff
  \newpage
\fi



\bibliographystyle{IEEEtran}
\bibliography{refs}

\begin{IEEEbiography}[{\includegraphics[width=1in,height=1.25in,clip,keepaspectratio]{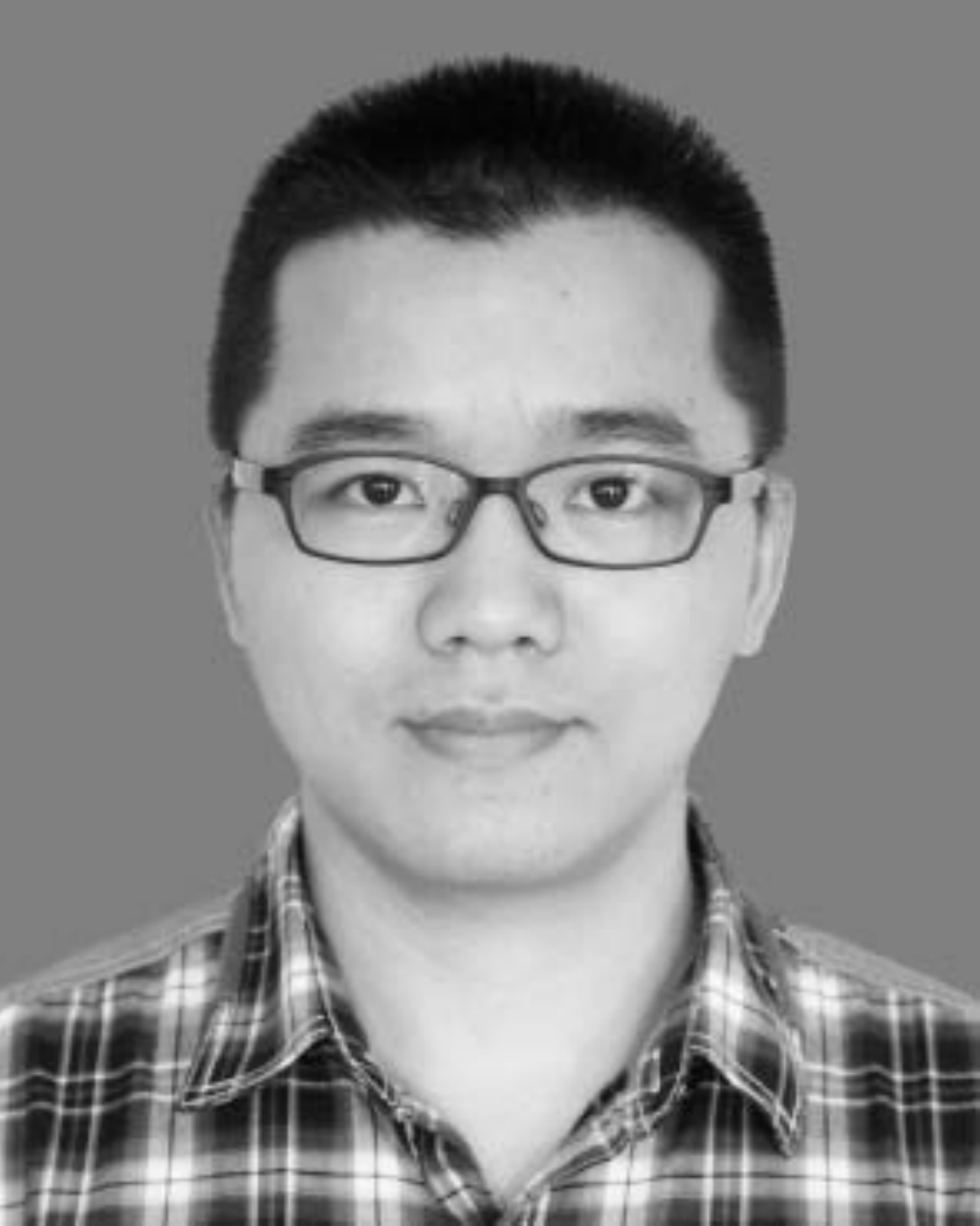}}]{Dong Huang}
received the B.S. degree in computer science in 2009 from South China University of Technology, Guangzhou, China. He received the M.Sc. degree in computer science in 2011 and the Ph.D. degree in computer science in 2015, both from Sun Yat-sen University, Guangzhou, China. He joined South China Agricultural University in 2015, where he is currently an Associate Professor with the College of Mathematics and Informatics. From July 2017 to July 2018, he was a visiting fellow with the School of Computer Science and Engineering, Nanyang Technological University, Singapore. His research interests include data mining and machine learning. He has published more than 60 papers in refereed journals and conferences, such as IEEE TKDE, IEEE TNNLS, IEEE TCYB, IEEE TSMC-S, ACM TKDD, SIGKDD, AAAI, and ICDM. He was the recipient of the 2020 ACM Guangzhou Rising Star Award.
\end{IEEEbiography}

\begin{IEEEbiography}[{\includegraphics[width=1in,height=1.25in,clip,keepaspectratio]{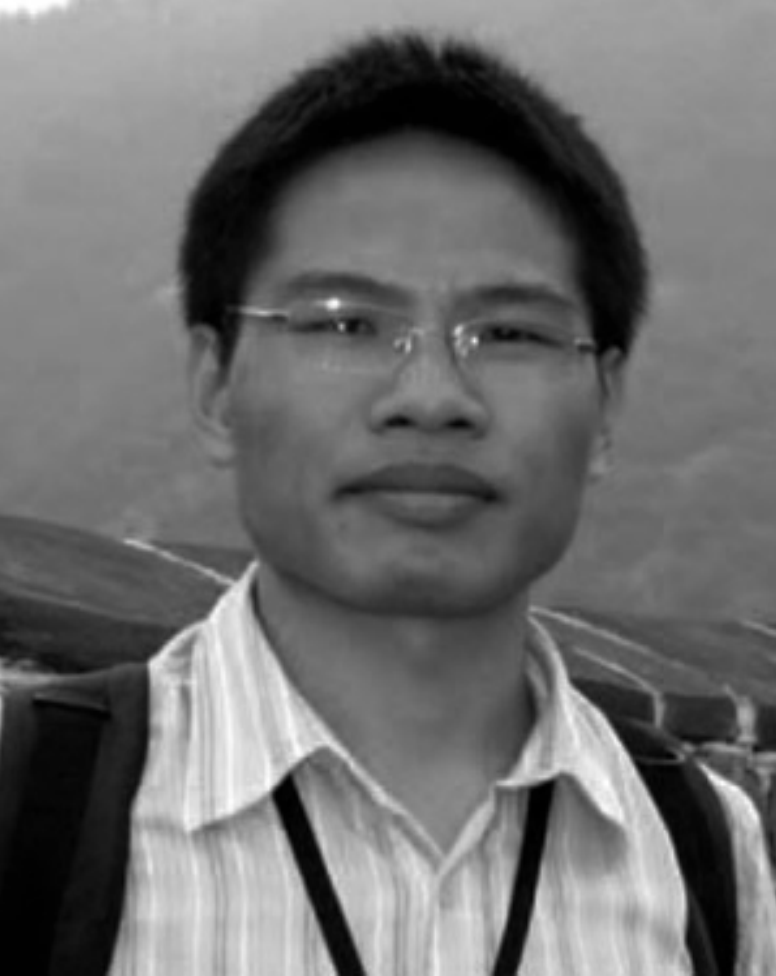}}]{Chang-Dong Wang}
received the B.S. degree in applied mathematics in 2008, the M.Sc. degree in computer science in 2010, and the Ph.D. degree in computer science in 2013, all from Sun Yat-sen University, Guangzhou, China. He was a visiting student at the University of Illinois at Chicago from January 2012 to November 2012. He is currently an Associate Professor with the School of Data and Computer Science, Sun Yat-sen University, Guangzhou, China. His current research interests include machine learning and data mining. He has published more than 100 scientific papers in international journals and conferences such as IEEE TPAMI, IEEE TKDE, IEEE TNNLS, IEEE TSMC-C, ACM TKDD, Pattern Recognition, KDD, ICDM and SDM. His ICDM 2010 paper won the Honorable Mention for Best Research Paper Award. He was awarded 2015 Chinese Association for Artificial Intelligence (CAAI) Outstanding Dissertation.
\end{IEEEbiography}

\begin{IEEEbiography}[{\includegraphics[width=1in,height=1.25in,clip,keepaspectratio]{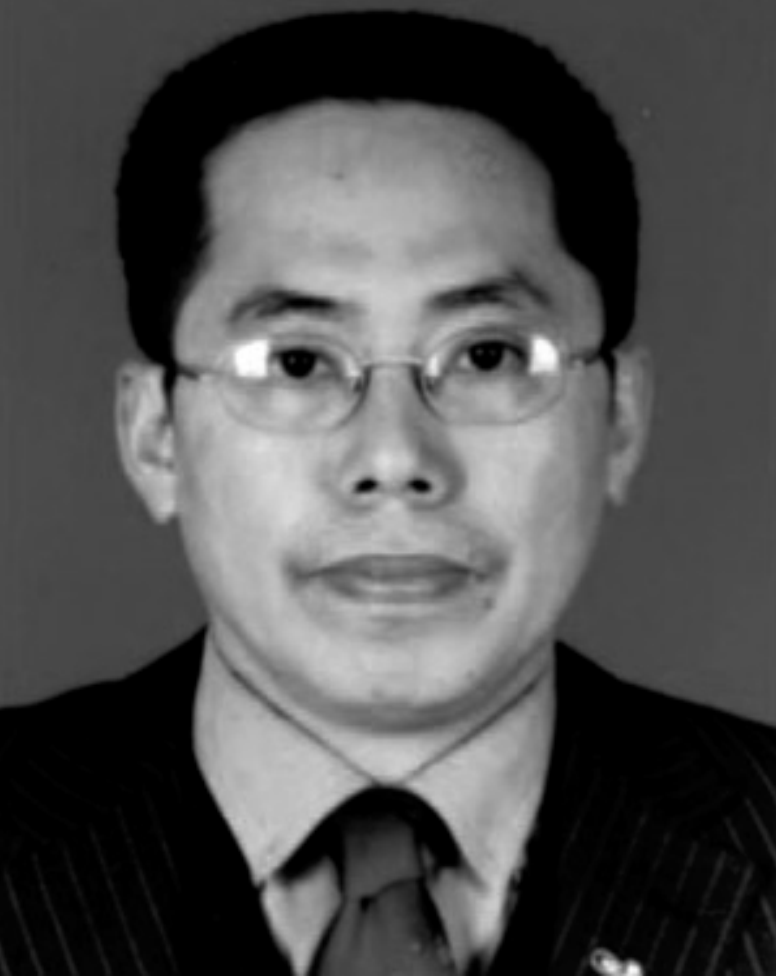}}]{Jian-Huang Lai}
received the M.Sc. degree in applied mathematics in 1989 and the Ph.D. degree in mathematics in 1999 from Sun Yat-sen University, China. He joined Sun Yat-sen University in 1989 as an Assistant Professor, where he is currently a Professor with the School of Data and Computer Science. His current research interests include the areas of digital image processing, pattern recognition, multimedia communication, wavelet and its applications. He has published more than 200 scientific papers in the international journals and conferences on image processing and pattern recognition, such as IEEE TPAMI, IEEE TKDE, IEEE TNN, IEEE TIP, IEEE TSMC-B, Pattern Recognition, ICCV, CVPR, IJCAI, ICDM and SDM. Prof. Lai serves as a Standing Member of the Image and Graphics Association of China, and also serves as a Standing Director of the Image and Graphics Association of Guangdong.
\end{IEEEbiography}

\end{document}